\ifx\XeTeXrevision\undefined\pdfoutput=1\fi
\documentclass[11pt,onecolumn]{cleantechnicalreport}
\usepackage[authoryear,round]{natbib}
\usepackage{hybridfrontmatter}

\usepackage{setspace}    %

\usepackage{threeparttable}
\usepackage{adjustbox}
\let\TPTorig\threeparttable
\let\endTPTorig\endthreeparttable
\renewenvironment{threeparttable}{\begin{adjustbox}{max totalheight=0.97\textheight}\TPTorig}{\endTPTorig\end{adjustbox}}
\usepackage{longtable}
\usepackage{array}
\usepackage{multirow}
\usepackage{makecell}
\usepackage{rotating}   %
\usepackage{tikz}
\usetikzlibrary{arrows.meta,positioning,fit,calc,shapes}

\newcommand{\RepoURL}{\url{https://github.com/ruijiang81/agentic_multiverse_analysis}}

\newcommand{\nPapers}{324}  %
\newcommand{\nRnePapers}{201}  %
\newcommand{\nAddedPapers}{123}  %
\newcommand{\nExperiments}{13}  %
\newcommand{\nListedN}{337}  %
\newcommand{\nAddLinkedN}{4{,}863}  %
\newcommand{\nAddCandN}{296}  %
\newcommand{\nAddRegN}{283}  %
\newcommand{\nAddDownloadN}{90}  %
\newcommand{\nAddInfeasibleN}{55}  %
\newcommand{\nAddNoestN}{15}  %
\newcommand{\nAddVersionsN}{9}  %
\newcommand{\nAddNoobtainN}{4}  %
\newcommand{\nAddedExpN}{10}  %
\newcommand{\nDir}{306}  %
\newcommand{\nReg}{304}  %
\newcommand{\nCells}{59{,}106}  %
\newcommand{\nShareMeanPct}{25\%}  %
\newcommand{\nShareMedianPct}{15\%}  %
\newcommand{\nShareZeroPct}{23\%}  %
\newcommand{\nShareZeroN}{71}  %
\newcommand{\nShareHalfN}{59}  %
\newcommand{\nShareAllN}{12}  %
\newcommand{\nMedzMean}{0.77}  %
\newcommand{\nMedzMedian}{0.70}  %
\newcommand{\nMedzNegPct}{27\%}  %
\newcommand{\nCorrShareMedz}{0.89}  %
\newcommand{\nCorrShareRep}{0.46}  %
\newcommand{\nCorrRepMedz}{0.50}  %
\newcommand{\nSignMeanPct}{68\%}  %
\newcommand{\nSignMedianPct}{77\%}  %
\newcommand{\nCorrShareSign}{0.64}  %
\newcommand{\nCorrSignMedz}{0.84}  %
\newcommand{\nRelShare}{0.93}  %
\newcommand{\nRelShareR}{0.87}  %
\newcommand{\nRelMedz}{0.93}  %
\newcommand{\nRelMedzR}{0.88}  %
\newcommand{\nRelAllShare}{0.93}  %
\newcommand{\nRelAllMedz}{0.94}  %
\newcommand{\nRelAllN}{323}  %
\newcommand{\nRepAllPct}{42\%}  %
\newcommand{\nRepAllK}{128}  %
\newcommand{\nRepAllN}{305}  %
\newcommand{\nRepExactPct}{45\%}  %
\newcommand{\nRepAltPct}{33\%}  %
\newcommand{\nRepFoundPct}{83\%}  %
\newcommand{\nRepFoundK}{15}  %
\newcommand{\nRepFoundN}{18}  %
\newcommand{\nRepRebuiltPct}{39\%}  %
\newcommand{\nRepOldAllPct}{47\%}  %
\newcommand{\nRepOldOnlyN}{16}  %
\newcommand{\nRepMatFoundN}{20}  %
\newcommand{\nRepMatDataN}{8}  %
\newcommand{\nRepMatNoneN}{304}  %
\newcommand{\nCellsAll}{62{,}409}  %
\newcommand{\nCellsAllPapers}{323}  %
\newcommand{\nGridMedian}{192}  %
\newcommand{\nCellsFail}{256}  %
\newcommand{\nCellsFailPapers}{9}  %
\newcommand{\nCompRunsN}{259}  %
\newcommand{\nCompRestoredMedian}{2}  %
\newcommand{\nRepMiss}{177}  %
\newcommand{\nRepFailMag}{115}  %
\newcommand{\nRepFailSign}{60}  %
\newcommand{\nRepFailP}{137}  %
\newcommand{\nRepFailPPct}{77\%}  %
\newcommand{\nRepFailOnlyP}{49}  %
\newcommand{\nRepFailAllThree}{42}  %
\newcommand{\nRepFailMulti}{93}  %
\newcommand{\nRepSignAlsoP}{55}  %
\newcommand{\nRepMissNComp}{163}  %
\newcommand{\nRepMissNDiff}{141}  %
\newcommand{\nRepMissNDiffPct}{87\%}  %
\newcommand{\nRepSpearNgap}{0.28}  %
\newcommand{\nRepSpearNgapP}{\ensuremath{p < 0.001}}  %
\newcommand{\nRepSpearN}{278}  %
\newcommand{\nRepMedDevPct}{46\%}  %
\newcommand{\nRepExpMin}{21\%}  %
\newcommand{\nRepExpMinName}{the mutual fund scandal}  %
\newcommand{\nRepExpMax}{83\%}  %
\newcommand{\nRepExpMaxName}{BlackRock's acquisition of BGI}  %
\newcommand{\nShareExpMinPct}{14\%}  %
\newcommand{\nShareExpMinName}{the universal demand laws}  %
\newcommand{\nShareExpMaxPct}{43\%}  %
\newcommand{\nShareExpMaxName}{the Dodd-Frank Act}  %
\newcommand{\nCellOppPct}{32\%}  %
\newcommand{\nCellPfailPct}{43\%}  %
\newcommand{\nCellSupPct}{25\%}  %
\newcommand{\nCellPfailAmongSignPct}{63\%}  %
\newcommand{\nExpRegfdSharePct}{32\%}  %
\newcommand{\nExpRegfdN}{44}  %
\newcommand{\nExpRegshoSharePct}{21\%}  %
\newcommand{\nExpRegshoN}{39}  %
\newcommand{\nExpRussellN}{37}  %
\newcommand{\nExpK}{12}  %
\newcommand{\nExpRsqShare}{0.07}  %
\newcommand{\nExpFShareP}{\ensuremath{p = 0.023}}  %
\newcommand{\nExpRsqSign}{0.06}  %
\newcommand{\nExpFSignP}{\ensuremath{p = 0.047}}  %
\newcommand{\nExpRsqRep}{0.08}  %
\newcommand{\nExpFRepP}{\ensuremath{p = 0.012}}  %
\newcommand{\nExpRsqMedz}{0.07}  %
\newcommand{\nExpFMedzP}{\ensuremath{p = 0.028}}  %
\newcommand{\nBrodeurDays}{13}  %
\newcommand{\nShareRepMeanPct}{41\%}  %
\newcommand{\nShareNotRepMeanPct}{14\%}  %
\newcommand{\nShareRepN}{128}  %
\newcommand{\nShareNotRepN}{176}  %
\newcommand{\nMedzRepMean}{1.47}  %
\newcommand{\nMedzNotRepMean}{0.26}  %
\newcommand{\nOrdOneN}{8}  %
\newcommand{\nOrdOneMedz}{2.09}  %
\newcommand{\nOrdTwoN}{181}  %
\newcommand{\nOrdTwoRep}{43\%}  %
\newcommand{\nOrdThreeN}{115}  %
\newcommand{\nOrdThreeRep}{39\%}  %
\newcommand{\nOrdOneExp}{5}  %
\newcommand{\nOrdOneRepN}{6}  %
\newcommand{\nOthN}{64}  %
\newcommand{\nOthRegfdN}{39}  %
\newcommand{\nIvN}{35}  %
\newcommand{\nIvMfsRusN}{33}  %
\newcommand{\nRdN}{8}  %
\newcommand{\nOthExpN}{11}  %
\newcommand{\nRsqExpOth}{0.56}  %
\newcommand{\nJourTopN}{76}  %
\newcommand{\nJourUtdN}{14}  %
\newcommand{\nJourFtfiftyN}{30}  %
\newcommand{\nJourOthjN}{142}  %
\newcommand{\nJourWpN}{42}  %
\newcommand{\nMatRegN}{18}  %
\newcommand{\nVersionDiffN}{23}  %
\newcommand{\nVersionYearOnlyN}{2}  %
\newcommand{\cShareNoAloneWpPp}{12}  %

\newcommand{\cShareNoAloneWpP}{\ensuremath{p = 0.019}}
\newcommand{\cShareNoAloneRdPp}{25}  %

\newcommand{\cShareNoAloneRdP}{\ensuremath{p = 0.034}}
\newcommand{\cShareNoAloneOthdesPp}{10}  %

\newcommand{\cShareNoAloneOthdesP}{\ensuremath{p = 0.021}}
\newcommand{\cShareNoAloneLogtPp}{21}  %

\newcommand{\cShareNoAloneLogtP}{\ensuremath{p < 0.001}}
\newcommand{\cShareFeAloneOrdOnePp}{35}  %

\newcommand{\cShareFeAloneOrdOneP}{\ensuremath{p = 0.063}}
\newcommand{\cShareFeAloneOrdThreePp}{7}  %

\newcommand{\cShareFeAloneOrdThreeP}{\ensuremath{p = 0.070}}
\newcommand{\cShareFeAloneFtfiftyPp}{4}  %

\newcommand{\cShareFeAloneFtfiftyP}{\ensuremath{p = 0.58}}
\newcommand{\cShareFeAloneOthjPp}{8}  %

\newcommand{\cShareFeAloneOthjP}{\ensuremath{p = 0.054}}
\newcommand{\cShareFeAloneWpPp}{8}  %

\newcommand{\cShareFeAloneWpP}{\ensuremath{p = 0.102}}
\newcommand{\cShareFeAloneIvPp}{9}  %

\newcommand{\cShareFeAloneIvP}{\ensuremath{p = 0.29}}
\newcommand{\cShareFeAloneRdPp}{19}  %

\newcommand{\cShareFeAloneRdP}{\ensuremath{p = 0.19}}
\newcommand{\cShareFeAloneOthdesPp}{4}  %

\newcommand{\cShareFeAloneOthdesP}{\ensuremath{p = 0.56}}
\newcommand{\cShareFeAloneAltPp}{1}  %

\newcommand{\cShareFeAloneAltP}{\ensuremath{p = 0.80}}
\newcommand{\cShareFeAloneMatPp}{20}  %

\newcommand{\cShareFeAloneMatP}{\ensuremath{p = 0.014}}
\newcommand{\cShareFeAloneLogtPp}{20}  %

\newcommand{\cShareFeAloneLogtP}{\ensuremath{p < 0.001}}
\newcommand{\cShareFeAloneTopinstPp}{7}  %

\newcommand{\cShareFeAloneTopinstP}{\ensuremath{p = 0.11}}
\newcommand{\cShareFeAloneAuthmPpd}{3.8}  %

\newcommand{\cShareFeAloneAuthmP}{\ensuremath{p = 0.013}}
\newcommand{\cShareFeAloneEarlierPpd}{1.6}  %

\newcommand{\cShareFeAloneEarlierP}{\ensuremath{p = 0.14}}
\newcommand{\cShareFeJointOrdOnePp}{32}  %

\newcommand{\cShareFeJointOrdOneP}{\ensuremath{p = 0.028}}
\newcommand{\cShareFeJointMatPp}{17}  %

\newcommand{\cShareFeJointMatP}{\ensuremath{p = 0.050}}
\newcommand{\cShareFeJointLogtPp}{18}  %

\newcommand{\cShareFeJointLogtP}{\ensuremath{p < 0.001}}
\newcommand{\cShareFeJointTopinstPp}{2}  %

\newcommand{\cShareFeJointTopinstP}{\ensuremath{p = 0.72}}
\newcommand{\cShareFeJointAuthmPpd}{3.7}  %

\newcommand{\cShareFeJointAuthmP}{\ensuremath{p = 0.029}}
\newcommand{\cSignNoAloneIvPp}{9}  %

\newcommand{\cSignNoAloneIvP}{\ensuremath{p = 0.073}}
\newcommand{\cSignNoAloneRdPp}{15}  %

\newcommand{\cSignNoAloneRdP}{\ensuremath{p = 0.088}}
\newcommand{\cSignNoAloneAltPp}{10}  %

\newcommand{\cSignNoAloneAltP}{\ensuremath{p = 0.016}}
\newcommand{\cSignNoJointAltPp}{13}  %

\newcommand{\cSignNoJointAltP}{\ensuremath{p = 0.004}}
\newcommand{\cSignFeAloneOthjPp}{9}  %

\newcommand{\cSignFeAloneOthjP}{\ensuremath{p = 0.034}}
\newcommand{\cSignFeAloneMatPp}{26}  %

\newcommand{\cSignFeAloneMatP}{\ensuremath{p < 0.001}}
\newcommand{\cSignFeAloneLogtPp}{9}  %

\newcommand{\cSignFeAloneLogtP}{\ensuremath{p = 0.050}}
\newcommand{\cSignFeAloneTopinstPp}{8}  %

\newcommand{\cSignFeAloneTopinstP}{\ensuremath{p = 0.026}}
\newcommand{\cSignFeAloneAuthmPpd}{1.7}  %

\newcommand{\cSignFeAloneAuthmP}{\ensuremath{p = 0.31}}
\newcommand{\cSignFeAloneEarlierPpd}{2.4}  %

\newcommand{\cSignFeAloneEarlierP}{\ensuremath{p = 0.013}}
\newcommand{\cSignFeJointUtdPp}{16}  %

\newcommand{\cSignFeJointUtdP}{\ensuremath{p = 0.091}}
\newcommand{\cSignFeJointAltPp}{10}  %

\newcommand{\cSignFeJointAltP}{\ensuremath{p = 0.036}}
\newcommand{\cSignFeJointMatPp}{26}  %

\newcommand{\cSignFeJointMatP}{\ensuremath{p < 0.001}}
\newcommand{\cSignFeJointEarlierPpd}{2.9}  %

\newcommand{\cSignFeJointEarlierP}{\ensuremath{p = 0.015}}
\newcommand{\cRepNoAloneIvPp}{18}  %

\newcommand{\cRepNoAloneIvP}{\ensuremath{p = 0.029}}
\newcommand{\cRepNoAloneAltPp}{11}  %

\newcommand{\cRepNoAloneAltP}{\ensuremath{p = 0.078}}
\newcommand{\cRepNoJointAltPp}{16}  %

\newcommand{\cRepNoJointAltP}{\ensuremath{p = 0.010}}
\newcommand{\cRepFeAloneOrdOnePp}{32}  %

\newcommand{\cRepFeAloneOrdOneP}{\ensuremath{p = 0.11}}
\newcommand{\cRepFeAloneOrdThreePp}{2}  %

\newcommand{\cRepFeAloneOrdThreeP}{\ensuremath{p = 0.78}}
\newcommand{\cRepFeAloneOthjPp}{13}  %

\newcommand{\cRepFeAloneOthjP}{\ensuremath{p = 0.068}}
\newcommand{\cRepFeAloneRdPp}{18}  %

\newcommand{\cRepFeAloneRdP}{\ensuremath{p = 0.48}}
\newcommand{\cRepFeAloneAltPp}{5}  %

\newcommand{\cRepFeAloneAltP}{\ensuremath{p = 0.44}}
\newcommand{\cRepFeAloneMatPp}{46}  %

\newcommand{\cRepFeAloneMatP}{\ensuremath{p < 0.001}}
\newcommand{\cRepFeAloneLogtPp}{29}  %

\newcommand{\cRepFeAloneLogtP}{\ensuremath{p < 0.001}}
\newcommand{\cRepFeAloneTopinstPp}{13}  %

\newcommand{\cRepFeAloneTopinstP}{\ensuremath{p = 0.075}}
\newcommand{\cRepFeAloneAuthmPpd}{1.0}  %

\newcommand{\cRepFeAloneAuthmP}{\ensuremath{p = 0.72}}
\newcommand{\cRepFeAloneEarlierPpd}{3.1}  %

\newcommand{\cRepFeAloneEarlierP}{\ensuremath{p = 0.056}}
\newcommand{\cRepFeJointOrdOnePp}{31}  %

\newcommand{\cRepFeJointOrdOneP}{\ensuremath{p = 0.075}}
\newcommand{\cRepFeJointMatPp}{45}  %

\newcommand{\cRepFeJointMatP}{\ensuremath{p = 0.002}}
\newcommand{\cRepFeJointLogtPp}{26}  %

\newcommand{\cRepFeJointLogtP}{\ensuremath{p < 0.001}}
\newcommand{\cRepFeJointEarlierPpd}{3.0}  %

\newcommand{\cRepFeJointEarlierP}{\ensuremath{p = 0.11}}

\newcommand{\cMedzNoAloneRdCoef}{0.70}  %

\newcommand{\cMedzNoAloneRdP}{\ensuremath{p = 0.073}}
\newcommand{\cMedzFeAloneOrdOneCoef}{1.17}  %

\newcommand{\cMedzFeAloneOrdOneP}{\ensuremath{p = 0.19}}
\newcommand{\cMedzFeAloneOrdThreeCoef}{\ensuremath{-}0.13}  %

\newcommand{\cMedzFeAloneOrdThreeP}{\ensuremath{p = 0.40}}
\newcommand{\cMedzFeAloneUtdCoef}{\ensuremath{-}0.10}  %

\newcommand{\cMedzFeAloneUtdP}{\ensuremath{p = 0.77}}
\newcommand{\cMedzFeAloneFtfiftyCoef}{\ensuremath{-}0.27}  %

\newcommand{\cMedzFeAloneFtfiftyP}{\ensuremath{p = 0.33}}
\newcommand{\cMedzFeAloneOthjCoef}{\ensuremath{-}0.43}  %

\newcommand{\cMedzFeAloneOthjP}{\ensuremath{p = 0.017}}

\newcommand{\cMedzFeAloneRdCoef}{0.59}  %

\newcommand{\cMedzFeAloneRdP}{\ensuremath{p = 0.27}}

\newcommand{\cMedzFeAloneAltCoef}{\ensuremath{-}0.14}  %

\newcommand{\cMedzFeAloneAltP}{\ensuremath{p = 0.49}}
\newcommand{\cMedzFeAloneMatCoef}{0.84}  %

\newcommand{\cMedzFeAloneMatP}{\ensuremath{p = 0.001}}
\newcommand{\cMedzFeAloneLogtCoef}{0.73}  %

\newcommand{\cMedzFeAloneLogtP}{\ensuremath{p < 0.001}}
\newcommand{\cMedzFeAloneTopinstCoef}{0.40}  %

\newcommand{\cMedzFeAloneTopinstP}{\ensuremath{p = 0.024}}
\newcommand{\cMedzFeAloneAuthmCoef}{\ensuremath{-}0.11}  %

\newcommand{\cMedzFeAloneAuthmP}{\ensuremath{p = 0.069}}
\newcommand{\cMedzFeAloneEarlierCoef}{\ensuremath{-}0.10}  %

\newcommand{\cMedzFeAloneEarlierP}{\ensuremath{p = 0.012}}
\newcommand{\cMedzFeJointOrdOneCoef}{1.05}  %

\newcommand{\cMedzFeJointOrdOneP}{\ensuremath{p = 0.17}}
\newcommand{\cMedzFeJointOthjCoef}{\ensuremath{-}0.04}  %

\newcommand{\cMedzFeJointOthjP}{\ensuremath{p = 0.87}}
\newcommand{\cMedzFeJointAltCoef}{\ensuremath{-}0.32}  %

\newcommand{\cMedzFeJointAltP}{\ensuremath{p = 0.075}}
\newcommand{\cMedzFeJointMatCoef}{0.75}  %

\newcommand{\cMedzFeJointMatP}{\ensuremath{p = 0.004}}
\newcommand{\cMedzFeJointTopinstCoef}{0.15}  %

\newcommand{\cMedzFeJointTopinstP}{\ensuremath{p = 0.44}}
\newcommand{\cMedzFeJointAuthmCoef}{\ensuremath{-}0.14}  %

\newcommand{\cMedzFeJointAuthmP}{\ensuremath{p = 0.038}}
\newcommand{\cMedzFeJointEarlierCoef}{\ensuremath{-}0.11}  %

\newcommand{\cMedzFeJointEarlierP}{\ensuremath{p = 0.019}}

\newcommand{\nJointK}{15}  %
\newcommand{\nOmRuns}{69}  %
\newcommand{\nOmFindN}{108}  %
\newcommand{\nOmEstFindN}{14}  %
\newcommand{\nOmInfFindN}{33}  %
\newcommand{\nOmDocFindN}{61}  %
\newcommand{\nOmEstFindRunsN}{13}  %
\newcommand{\nOmKappa}{0.94}  %
\newcommand{\nOmDisagreeN}{4}  %
\newcommand{\nDecZeroN}{83}  %
\newcommand{\nDecNonzeroN}{223}  %
\newcommand{\nDecDecidesN}{50}  %
\newcommand{\nDecDecidesPct}{16\%}  %
\newcommand{\nDecNullExceedPct}{68\%}  %
\newcommand{\nDecNullDraws}{1{,}000}  %
\newcommand{\nFamEstN}{104}  %
\newcommand{\nFamRdivN}{46}  %
\newcommand{\nFamOutcN}{213}  %
\newcommand{\nFamOutcTop}{21\%}  %
\newcommand{\nCompFullN}{251}  %
\newcommand{\nCompFullUp}{115}  %
\newcommand{\nCompFullDn}{77}  %
\newcommand{\nCompFullSignP}{\ensuremath{p = 0.007}}  %
\newcommand{\nCompSubN}{248}  %
\newcommand{\nCompSubSignP}{\ensuremath{p = 0.007}}  %
\newcommand{\nCompAllN}{258}  %
\newcommand{\nRestoredInv}{507}  %
\newcommand{\nRestoredAll}{633}  %
\newcommand{\nVerInvonlyP}{\ensuremath{p = 0.13}}  %
\newcommand{\nVerMixedP}{\ensuremath{p = 0.032}}  %
\newcommand{\nPathN}{267}  %
\newcommand{\nPathRepN}{114}  %
\newcommand{\nPathNotRepN}{152}  %
\newcommand{\nPathRepMissWord}{one}  %
\newcommand{\nPathAllZeroPct}{43\%}  %
\newcommand{\nPathAllOnePct}{36\%}  %
\newcommand{\nPathAllThreePct}{28\%}  %
\newcommand{\nPathRepZeroPct}{81\%}  %
\newcommand{\nPathRepOnePct}{64\%}  %
\newcommand{\nPathRepTwoPct}{57\%}  %
\newcommand{\nPathRepThreePct}{47\%}  %
\newcommand{\nPathNotrepZeroPct}{14\%}  %
\newcommand{\nPathNotrepTwoPct}{15\%}  %
\newcommand{\nPathAllAllkPct}{30\%}  %
\newcommand{\nPathKThreePct}{82\%}  %
\newcommand{\nValidCellsPct}{62\%}  %
\newcommand{\nValidPapers}{278}  %
\newcommand{\nValidCorr}{0.97}  %
\newcommand{\nValidSame}{173}  %
\newcommand{\nNullAbsMedz}{0.50}  %
\newcommand{\nNullRepN}{3}  %
\newcommand{\nNullBigPN}{12}  %
\newcommand{\nNull}{16}  %
\newcommand{\nUnclear}{2}  %
\newcommand{\nNullUnclearN}{18}  %
\newcommand{\nValRounds}{535}  %
\newcommand{\nValRoundsPerRun}{1.7}  %
\newcommand{\nValBlockRunsN}{49}  %
\newcommand{\nValMajorRunsN}{186}  %
\newcommand{\nValReproAuditN}{271}  %
\newcommand{\nValReproErrN}{201}  %
\newcommand{\nValSampleFixN}{246}  %
\newcommand{\nValSampleOpenN}{115}  %
\newcommand{\nKappaData}{0.86}  %
\newcommand{\nAlphaOrder}{0.86}  %
\newcommand{\nAgentsMedian}{88}  %
\newcommand{\nHoursMedian}{22}  %
\newcommand{\nTokensMedian}{126}  %
\newcommand{\nComChoicesN}{1{,}855}  %
\newcommand{\nComSilentPct}{42\%}  %
\newcommand{\nComPartialPct}{27\%}  %
\newcommand{\nComUnsupPct}{18\%}  %
\newcommand{\nComDefendedPct}{7\%}  %
\newcommand{\nComDownplayedPct}{5\%}  %
\newcommand{\nForksMedian}{26}  %
\newcommand{\nOptionsMedian}{215}  %
\newcommand{\nInfWildPct}{52\%}  %
\newcommand{\nInfRandPct}{20\%}  %
\newcommand{\nExactDataN}{247}  %
\newcommand{\nAltDataN}{77}  %
\newcommand{\nHeathN}{76}  %
\newcommand{\nFixedOwnPct}{83\%}  %
\newcommand{\nFixedInvalidN}{Forty}  %
\newcommand{\nQuestCrossedPct}{8\%}  %
\newcommand{\nQuestCellsPct}{38\%}  %
\newcommand{\nFixOneSidedN}{3}  %
\newcommand{\nFixJointN}{7}  %
\newcommand{\nHeathRneN}{67}  %
\newcommand{\nKappaDataAdjN}{18}  %
\newcommand{\nConfigBackN}{56}  %
\newcommand{\nReproBackN}{24}  %
\newcommand{\nReproUnresolvedWord}{two}  %
\newcommand{\nOrderSameN}{287}  %
\newcommand{\nReportFailWord}{four}  %
\newcommand{\nStaggerSilentPct}{41\%}  %
\newcommand{\nIntExtraAuditN}{5}  %
\newcommand{\nIntExtraPassN}{3}  %
\newcommand{\nIntUnresolvedN}{2}  %
\newcommand{\nIntAnchorN}{3}  %
\newcommand{\nIntNoteRunsN}{99}  %
\newcommand{\nIntStopN}{91}  %
\newcommand{\nIntStopCauseN}{59}  %
\newcommand{\nIntRetryN}{15}  %
\newcommand{\nIntVoidRunsN}{35}  %
\newcommand{\nIntVoidStatusWord}{two}  %
\newcommand{\nIntVoidIncompleteWord}{two}  %
\newcommand{\nRneTabsN}{51}  %
\newcommand{\nRneTabsMultiN}{19}  %
\newcommand{\nRneTabsOutWord}{five}  %
\newcommand{\nAttStartedN}{216}  %
\newcommand{\nAttStoppedN}{15}  %
\newcommand{\nAttNeverN}{121}  %
\newcommand{\nAttRestrictedN}{64}  %
\newcommand{\nAttNotHeldN}{19}  %
\newcommand{\nAttNoEstN}{51}  %
\newcommand{\nAttOtherN}{2}  %
\newcommand{\nPathAudNoKZeroN}{141}  %
\newcommand{\nPathAudNoKOneN}{53}  %
\newcommand{\nIntVoidCompanionWord}{two}  %
\newcommand{\nIntRerunWord}{one}  %
\newcommand{\nReproBackErrFixedN}{14}  %
\newcommand{\nInfShoRandN}{40}  %
\newcommand{\nInfShoN}{45}  %
\newcommand{\nInfUdlWildN}{24}  %
\newcommand{\nInfUdlN}{25}  %
\newcommand{\nInfRusAnalyticN}{27}  %
\newcommand{\nInfRusN}{38}  %
\newcommand{\nInfResampleN}{251}  %
\newcommand{\nInfAllN}{324}  %
\newcommand{\nForbidMinN}{6}  %
\newcommand{\nForbidMaxN}{49}  %
\newcommand{\nRneRowsN}{822}  %
\newcommand{\nRneTitlesN}{338}  %
\newcommand{\nListedDistinctN}{336}  %
\newcommand{\nAttNeverRestrictedN}{52}  %
\newcommand{\nAttNeverNotHeldN}{16}  %
\newcommand{\nAttNeverNoEstN}{51}  %
\newcommand{\nAttNeverOtherN}{2}  %
\newcommand{\nAttStoppedRestrictedN}{12}  %
\newcommand{\nAttStoppedNotHeldN}{3}  %
\newcommand{\nAttMsrRestrictedN}{12}  %
\newcommand{\nSelNotDoneN}{136}  %
\newcommand{\nSelTierChiP}{\ensuremath{p < 0.001}}  %
\newcommand{\nSelTopP}{\ensuremath{p = 0.39}}  %
\newcommand{\nSelWpP}{\ensuremath{p = 0.03}}  %
\newcommand{\nSelBlankN}{12}  %
\newcommand{\nSelWpNoBlankP}{\ensuremath{p = 0.28}}  %
\newcommand{\nSelTierNoBlankP}{\ensuremath{p < 0.001}}  %
\newcommand{\nSelYearP}{\ensuremath{p = 0.10}}  %
\newcommand{\nSelExpP}{\ensuremath{p < 0.001}}  %
\newcommand{\nSelPermN}{20{,}000}  %
\newcommand{\nTierTopN}{60}  %
\newcommand{\nTierFieldN}{50}  %
\newcommand{\nTierOtherN}{26}  %
\newcommand{\nTierWpN}{65}  %
\newcommand{\nTierPubN}{136}  %
\newcommand{\nJournalsN}{39}  %
\newcommand{\nBlankJournalN}{16}  %
\newcommand{\nBlankJournalFilledN}{4}  %
\newcommand{\nBlankJournalWpN}{12}  %
\newcommand{\nTierFieldBlankWpN}{48}  %
\newcommand{\nTierOtherBlankWpN}{24}  %
\newcommand{\nTierWpBlankWpN}{69}  %
\newcommand{\nRneAuthorErrWord}{four}  %
\newcommand{\nVerDiffN}{33}  %
\newcommand{\nVerSameJournalN}{130}  %
\newcommand{\nVerSameWpN}{32}  %
\newcommand{\nVerBlankJournalN}{3}  %
\newcommand{\nVerBlankWpN}{13}  %
\newcommand{\nVerWpToJournalN}{21}  %
\newcommand{\nVerJournalToEarlierN}{2}  %
\newcommand{\nVerTitleAuthorN}{10}  %
\newcommand{\nVerAuthorDiffN}{4}  %
\newcommand{\nVerTitleDiffN}{6}  %
\newcommand{\nStaggerCheckN}{20}  %
\newcommand{\nStaggerCheckOffN}{13}  %
\newcommand{\nComTwoPapersN}{20}  %
\newcommand{\nComTwoChoicesN}{116}  %
\newcommand{\nComTwoSilentPct}{34\%}  %
\newcommand{\nComTwoFirstSilentPct}{42\%}  %
\newcommand{\nComTwoAgreePct}{85\%}  %
\newcommand{\nComTwoKappa}{0.69}  %
\newcommand{\nComTwoKappaFour}{0.50}  %
\newcommand{\nHeldOffWord}{five}  %
\newcommand{\nIPapers}{201}  %
\newcommand{\nIDir}{188}  %
\newcommand{\nIReg}{187}  %
\newcommand{\nIShareMeanPct}{23\%}  %
\newcommand{\nIShareMedianPct}{10\%}  %
\newcommand{\nIShareZeroPct}{26\%}  %
\newcommand{\nIMedzMedian}{0.73}  %
\newcommand{\nIRepAllPct}{44\%}  %
\newcommand{\nIRepAllK}{82}  %
\newcommand{\nIRepAllN}{188}  %
\newcommand{\nIOrdOneN}{4}  %
\newcommand{\nIOrdOneSharePct}{98\%}  %
\newcommand{\iShareFeAloneOrdOnePp}{78}  %

\newcommand{\iShareFeAloneOrdOneP}{\ensuremath{p < 0.001}}
\newcommand{\iShareFeAloneIvPp}{18}  %

\newcommand{\iShareFeAloneIvP}{\ensuremath{p = 0.052}}
\newcommand{\iShareFeAloneLogtPp}{23}  %

\newcommand{\iShareFeAloneLogtP}{\ensuremath{p < 0.001}}
\newcommand{\iShareFeAloneTopinstPp}{12}  %

\newcommand{\iShareFeAloneTopinstP}{\ensuremath{p = 0.033}}
\newcommand{\iShareFeAloneAuthmPpd}{3.8}  %

\newcommand{\iShareFeAloneAuthmP}{\ensuremath{p = 0.034}}
\newcommand{\iShareFeJointAuthmPpd}{1.8}  %

\newcommand{\iShareFeJointAuthmP}{\ensuremath{p = 0.31}}

\newcommand{\iSignFeAloneLogtP}{\ensuremath{p = 0.21}}
\newcommand{\iRepFeAloneMatPp}{48}  %

\newcommand{\iRepFeAloneMatP}{\ensuremath{p = 0.002}}

\newcommand{\iMedzFeAloneOthjCoef}{\ensuremath{-}0.46}  %

\newcommand{\iMedzFeAloneOthjP}{\ensuremath{p = 0.034}}

\newcommand{\iMedzFeAloneTopinstCoef}{0.57}  %

\newcommand{\iMedzFeAloneTopinstP}{\ensuremath{p = 0.004}}
\newcommand{\iMedzFeAloneAuthmCoef}{\ensuremath{-}0.15}  %

\newcommand{\iMedzFeAloneAuthmP}{\ensuremath{p = 0.037}}
\newcommand{\iMedzFeAloneEarlierCoef}{\ensuremath{-}0.08}  %

\newcommand{\iMedzFeAloneEarlierP}{\ensuremath{p = 0.101}}

\newcommand{\iMedzFeJointEarlierCoef}{\ensuremath{-}0.01}  %

\newcommand{\iMedzFeJointEarlierP}{\ensuremath{p = 0.81}}
\newcommand{\nISignMeanPct}{70\%}  %
\newcommand{\kShareNoAloneExpPp}{0.2}  %
\newcommand{\kShareNoAloneExpP}{\ensuremath{p = 0.90}}
\newcommand{\kShareFeAloneAuthmaxPp}{1.7}  %
\newcommand{\kShareFeAloneAuthmaxP}{\ensuremath{p = 0.043}}
\newcommand{\kShareFeAloneSamemPp}{1.6}  %
\newcommand{\kShareFeAloneSamemP}{\ensuremath{p = 0.54}}
\newcommand{\kRepNoAloneExpPp}{1.1}  %
\newcommand{\kRepNoAloneExpP}{\ensuremath{p = 0.54}}

\newcommand{\kMedzNoAloneExpCoef}{0.01}  %
\newcommand{\kMedzNoAloneExpP}{\ensuremath{p = 0.73}}

\newcommand{\kMedzFeAloneToptenCoef}{0.48}  %
\newcommand{\kMedzFeAloneToptenP}{\ensuremath{p = 0.038}}

\newcommand{\kMedzFeAloneTopfiftyCoef}{0.41}  %
\newcommand{\kMedzFeAloneTopfiftyP}{\ensuremath{p = 0.004}}

\newcommand{\kMedzFeJointToptenCoef}{0.17}  %
\newcommand{\kMedzFeJointToptenP}{\ensuremath{p = 0.45}}

\newcommand{\kMedzFeJointTopfiftyCoef}{0.26}  %
\newcommand{\kMedzFeJointTopfiftyP}{\ensuremath{p = 0.076}}

\newcommand{\nCrowdPapers}{622}  %
\newcommand{\nCrowdEntries}{624}  %
\newcommand{\nCrowdExpN}{14}  %
\newcommand{\nCrowdVersionsN}{9}  %
\newcommand{\nCrowdScreenN}{287}  %
\newcommand{\nListedNotDoneN}{136}  %
\newcommand{\nCrowdExpMin}{18}  %
\newcommand{\nCrowdExpMax}{87}  %
\newcommand{\nCrowdEarlierMedian}{20}  %
\newcommand{\nCrowdAuthAnyPct}{68\%}  %
\newcommand{\nCrowdAuthmMedian}{0.67}  %
\newcommand{\nTopInstN}{60}  %
\newcommand{\nITopInstPct}{25\%}  %
\newcommand{\nAffAddedN}{123}  %
\newcommand{\nExASharePct}{61\%}  %
\newcommand{\nExAImputPct}{100\%}  %
\newcommand{\nExAStackedPct}{81\%}  %
\newcommand{\nExACsPct}{2\%}  %
\newcommand{\nCompFullMainPct}{23\%}  %
\newcommand{\nCompFullCompPct}{26\%}  %
\newcommand{\nCompSubCompPct}{28\%}  %
\newcommand{\nCompSubMainPct}{23\%}  %
\newcommand{\nCompSubMedianMainPct}{10\%}  %
\newcommand{\nCompSubMedianSubPct}{12\%}  %
\newcommand{\nValidMeanPct}{23\%}  %
\newcommand{\nValidMeanAllPct}{24\%}  %
\newcommand{\nThrOnePct}{11\%}  %
\newcommand{\nThrTenPct}{25\%}  %
\newcommand{\nThrFiftyPct}{50\%}  %
\newcommand{\nShareExactMeanPct}{25\%}  %
\newcommand{\nShareAltMeanPct}{24\%}  %
\newcommand{\nNullShareMeanPct}{7\%}  %
\newcommand{\nNullShareMaxPct}{48\%}  %
\newcommand{\nOrdOneSharePct}{65\%}  %
\newcommand{\nOrdTwoSharePct}{26\%}  %
\newcommand{\nOrdThreeSharePct}{22\%}  %
\newcommand{\nDecTopMedianPp}{23}  %
\newcommand{\nDecTopQOnePp}{0}  %
\newcommand{\nDecTopQThreePp}{50}  %
\newcommand{\nDecNullMedianPp}{7}  %
\newcommand{\nFamEstRangePp}{18}  %
\newcommand{\nFamRdivRangePp}{21}  %
\newcommand{\nFamEstNetPp}{14}  %
\newcommand{\nFamRdivNetPp}{15}  %
\newcommand{\nExAEstRangePp}{98}  %
\newcommand{\nExASecondPp}{17}  %
\newcommand{\nVerInvonlyDiffPp}{5}  %
\newcommand{\nVerMixedDiffPp}{7}  %

\title{An AI-Powered Multiverse Analysis of Empirical Finance Research: Evidence from Reused Natural Experiments}
\author[1]{Ruijiang Gao}
\author[1]{Steven Chong Xiao}
\affil[1]{Naveen Jindal School of Management, The University of Texas at Dallas}
\reportauthornote{Xiao thanks the Department of Finance at the Wharton School, University of Pennsylvania, for data
access through WRDS during a visiting scholar appointment.}
\reportwebsite{https://github.com/ruijiang81/agentic_multiverse_analysis}
\reportcontact{ruijiang.gao@utdallas.edu, steven.xiao@utdallas.edu}

\begin{abstract}
Empirical finance research often reuses natural experiments to study other outcomes, raising concerns about
multiple testing and publication bias. We use AI agents to run a multiverse analysis of \nPapers{} papers that
reuse \nExperiments{} natural experiments. The agents rebuild the data, reproduce each paper's main estimate and
re-estimate it across grids totaling \nCellsAll{} estimated defensible specifications. Based on the reproductions and the specification grids, we develop measures of empirical robustness. We reproduce \nRepAllPct{} of the main estimates of claims with a signed effect under the papers' own
specifications. Across all defensible specifications, on average \nShareMeanPct{} keep the published sign and have
$p < 0.10$. Across experiments,
claims based on regression discontinuity are more robust than difference-in-differences claims. Within experiments,
claims about first-order effects, claims with larger published $t$-statistics, claims whose authors' data or code we
find, claims in top journals, and claims with fewer earlier papers on the same experiment are more robust. Claims on
experiments that more papers reuse are not less robust on average. Our findings quantify how much published claims depend on analysis choices, thereby
showing that large-scale multiverse analysis becomes feasible with agentic coding.
\end{abstract}

\begin{document}

\maketitle

\section{Introduction}\label{sec:intro}

In empirical finance, a single natural experiment can generate dozens of published findings. As researchers reuse the
same shock to study outcomes further down the economic chain, from its immediate, first-order effect to more distant
outcomes, the literature accumulates claims whose robustness has rarely been examined together. How much of this growing
body of evidence survives alternative, defensible analysis choices? Prior studies raise two concerns about this
practice. First, researchers who test many outcomes of the same shock face a multiple testing problem
\citep{heath2023reusing}, as do researchers who test many return predictors in asset pricing \citep{harvey2016cross,harvey2020evaluation,chordia2020anomalies}.
Second, journals and authors publish significant results more often than insignificant ones, and this selection biases
the published record \citep{ringgenberg2026publication,chen2020publication}. This line of research uses simulations and models of
reported test statistics to assess how much such biases could distort published findings. However, outside
asset-pricing anomalies, the published findings themselves have rarely been re-estimated and compared at scale. The
obstacle is cost: even with the authors' data and code, the replication teams in \citet{brodeur2026reproducibility}
spent \nBrodeurDays{} active days on average reproducing an article and checking its robustness. AI coding agents, large
language models that read files, write and run code, and save their output, now make such re-analysis feasible
\citep{kohler2026read,holtdirk2026automated}.

In this paper, we use AI agents to run a multiverse analysis of each of \nPapers{} papers (journal articles and working
papers) that reuse \nExperiments{} natural experiments, and we combine the results in a meta-analysis. Between the shock
and the published coefficient, the authors of each paper choose its sample, outcome definition, estimator, controls and
fixed effects. The median paper has \nForksMedian{} such choices with \nOptionsMedian{} options across them. A
multiverse analysis re-estimates the paper's claim under every combination of defensible options
\citep{steegen2016increasing,simonsohn2020specification}. The share of combinations that keep the paper's claim
therefore measures how much the claim depends on these choices. In our multiverse analysis, AI agents carry out each
step, and scripts check the output of each step. The design builds on evidence that AI agents given the same data make
different analysis choices \citep{gao2026nonstandard}. The advantage of our approach is that, instead of replicating a
random draw of published papers, we target papers that reuse the same experiments. Because these papers share the
treatment, the period and often the data, our approach holds the setting fixed and makes the papers comparable. Thus, we
can ask which characteristics of a paper are associated with the variation in robustness.

We study the finance and accounting experiments that the Reusing Natural Experiments (RNE) database of
\citet{heath2023reusing} lists most often, including the Regulation SHO (Reg SHO) pilot of short-sale rules and the
state business combination laws. Our sample includes \nRnePapers{} papers listed in RNE for which the full pipeline
completes and \nAddedPapers{} papers on the same experiments that we find through the citation network of the RNE
papers. For each paper, agents search for the authors' data or code (the authors' materials), rebuild the data from the
original sources where needed, and reproduce the paper's main estimate. They then list the analysis choices and the
defensible options for each choice. An agent, the choice auditor, rates each option and combines the valid and
questionable options of the four to eight most consequential choices into a grid, i.e., every combination of these
options. The median grid has \nGridMedian{} specifications. The auditor also fixes the grid's statistical inference before
any agent estimates an alternative specification. A different agent checks each step's output before we compute any
outcome. Of the \nPapers{} main claims, one per paper, \nDir{} report an effect of a given sign; the other
\nNullUnclearN{} report no effect or are unclear, and we summarize them separately.
We measure the robustness of each directional claim with four outcomes. The \emph{share supported} is the share of
its specifications in which the estimate has the published sign and a two-sided $p$-value below 0.10 under the grid's statistical inference. The \emph{share with the published sign} is a looser version of the \emph{share supported}: the share of
specifications whose estimate has the published sign, whatever its $p$-value. It separates the direction of the
estimates from their precision. \emph{Reproduced} indicates that our estimate of the paper's own specification, under the paper's own statistical inference (the standard errors or other inference method it reports), has the published sign and $p \le 0.10$, and is close to the
published estimate. The \emph{median signed $z$} is, in effect, the median $t$-statistic across specifications, positive
when the estimate has the published sign.

Across the \nReg{} directional claims we compare, the mean \emph{share supported} is
\nShareMeanPct{} and the median \nShareMedianPct{}. For \nShareZeroPct{} of the claims, the \emph{share supported} is zero,
and the median claim has a \emph{median signed $z$} of \nMedzMedian. We reproduce \nRepAllPct{} of the main estimates of the directional claims with a
published test (\nRepAllK{} of \nRepAllN). \emph{Reproduced} claims are more robust, with a mean \emph{share supported} of
\nShareRepMeanPct{}, compared with \nShareNotRepMeanPct{} for the others. The estimates keep the published sign more
often than they keep significance: the mean \emph{share with the published sign} is \nSignMeanPct, compared with
\nShareMeanPct{} for the \emph{share supported}.

Next, we run a meta-analysis across papers, relating the four outcomes to paper characteristics one at a time. Within
experiments, we compare papers that use the same natural experiment (experiment fixed effects); across experiments, we
compare all papers. We find that claims about first-order effects are more robust. Their \emph{share supported} is
higher by \cShareFeAloneOrdOnePp{} percentage points than that of claims about second-order effects
(\cShareFeAloneOrdOneP), a gap based on \nOrdOneN{} papers in \nOrdOneExp{} experiments. Claims with a larger published
$|t|$ are more robust on all four outcomes. A one-unit increase in the natural log of the published $|t|$ is associated
with a \emph{share supported} higher by \cShareFeAloneLogtPp{} percentage points (\cShareFeAloneLogtP). We reproduce the
main estimate more often when we find the authors' materials than when we do not, by \cRepFeAloneMatPp{} percentage
points (\cRepFeAloneMatP). Across natural experiments, claims based on a regression-discontinuity design (RDD) are more
robust than difference-in-differences claims, with a \emph{share supported} higher by \cShareNoAloneRdPp{} percentage
points (\cShareNoAloneRdP). Within experiments, claims in top journals (the top finance, economics and accounting journals) are more robust
than claims in journals outside both the UT Dallas 24 (UTD24) and Financial Times 50 (FT50) journal lists. Relative to top journals, claims in these other journals
have a lower \emph{median signed $z$} (\cMedzFeAloneOthjCoef, \cMedzFeAloneOthjP) and a \emph{share supported} lower by
\cShareFeAloneOthjPp{} percentage points (\cShareFeAloneOthjP).

In contrast, claims on experiments that more papers reuse are not less robust on average: the coefficient on the number
of papers on the experiment is not statistically significant for any of the four outcomes (for the \emph{share
supported}, \kShareNoAloneExpP). Within experiments, however, claims with fewer earlier papers on the same experiment are
more robust: ten more earlier papers are associated with a \emph{share with the published sign} lower by
\cSignFeAloneEarlierPpd{} percentage points (\cSignFeAloneEarlierP) and a coefficient of \cMedzFeAloneEarlierCoef{} on
the \emph{median signed $z$} (\cMedzFeAloneEarlierP). Within experiments, claims that we estimate with \emph{alternative data}, i.e., other data for at least one of the paper's data
sources, are as robust on average as claims that we rebuild from the paper's original sources. They are somewhat less
robust once we hold the other characteristics fixed, with a coefficient of \cMedzFeJointAltCoef{} on the \emph{median
signed $z$} (\cMedzFeJointAltP). Because alternative data are not the data the authors used, they may explain part of
the low robustness we observe. When we include all the characteristics in one multivariate regression, five stand out:
the first-order effect, the published $|t|$, the availability of the \emph{authors' materials}, the
\emph{repeat-author count} and the \emph{number of earlier papers}. Claims whose authors have written fewer other papers
on these experiments are more robust, and so are claims with fewer earlier papers on the same experiment.

We also look inside the grids. In \nDecDecidesN{} of the \nDir{} directional claims (\nDecDecidesPct), a single analysis
choice decides the result: the claim holds in at least half of the specifications under one option and in no more than 10\%
under another option. Among the papers in which any choice moves the \emph{share supported}, the most influential choice
(the one that moves it most) is the definition of the outcome variable in \nFamOutcTop{} of these papers, more often than any
other choice. The grid does not always include the paper's own specification. The choice auditor can rate one of the paper's own
options invalid, for example a two-way fixed effects regression when treatment starts at different dates, or hold a
choice fixed at an option other than the paper's. To measure how much these exclusions matter, an agent also estimates
a companion grid, i.e., the same grid with the paper's own options put back. Among the \nCompFullN{} directional claims with a companion grid, the
mean \emph{share supported} rises from \nCompFullMainPct{} on the main grid to \nCompFullCompPct{} on the companion
grid.

Finally, we find that specifications closer to the paper's own are more likely to keep the directional claim. Among
\emph{reproduced} claims whose own specification we can identify, that specification, re-estimated with the grid's statistical inference,
keeps the directional claim in \nPathRepZeroPct{} of cases. The mean \emph{share supported} falls to \nPathRepOnePct{}
with one choice changed and to \nPathRepThreePct{} with three or more. For each paper, an agent picks the four to six
choices that most threaten the claim (the riskiest choices). In \nComSilentPct{} of the cases, authors are silent on the
specification choices that agents consider most threatening to their claim.

Our findings speak to the concerns about reused natural experiments. \citet{heath2023reusing} and
\citet{ringgenberg2026publication} show how multiple testing and publication bias can produce false or inflated
findings; we re-estimate the published claims across their defensible specifications. Earlier critiques examine one
setting or one method \citep{karpoff2018institutional,appel2024identification,wei2024selection,baker2022much}. We apply
one protocol to \nPapers{} claims in \nExperiments{} settings.

This study contributes to three literatures. First, studies of researcher degrees of freedom, from
\citet{leamer1983lets} to many-analysts studies \citep{silberzahn2018many,breznau2022observing,menkveld2024nonstandard},
show that reasonable analysts reach different answers from the same data, usually for one research question at a time.
We measure how much a whole literature of published claims depends on such choices. Second, replication studies in
finance have largely focused on asset pricing, where standardized data sources and empirical specifications allow
researchers to re-estimate hundreds of return anomalies with common code
\citep{hou2020replicating,chen2022open,jensen2023there}. A reproducibility study also finds that rerunning the code of 168
research teams on the same raw data regenerates their exact results only 52\% of the time
\citep{perignon2024computational}. In contrast, studies of natural experiments are less standardized, since each paper
builds its own sample, outcome and estimator. Third, we add to work on AI agents in empirical research
\citep{brodeur2026ai,bertran2026many} by showing that agents, checked by scripts and audits at each step, can rebuild
the data and run a full multiverse analysis for a large sample of papers. Our measures can still contain errors, as the agents
can make mistakes that the scripts and audits do not catch. However, our findings that paper characteristics such as the published $|t|$ and the availability of the authors'
materials explain a statistically significant part of the variation in our measures suggest that these measures do not
merely capture noise.

Our results do not imply that a credible study must hold in every defensible specification. No study is likely to meet
that bar, and a true effect can lose significance under options that add noise or change what the estimate measures. If
referees and readers expect every result to survive every check, authors may also report only the checks that pass. This
concern is consistent with our finding that authors are silent on \nComSilentPct{} of their riskiest choices.
Transparency about how a result fares across defensible specifications, combined with realistic expectations about how
often a result grounded in a sound economic hypothesis should hold across them, may be a more productive path. AI agents now
make such multiverse analyses inexpensive, and we share our pipeline at \RepoURL{} so that researchers can apply it to
their own papers.

\section{Natural Experiments and the Sample}\label{sec:sample}

\subsection{The RNE Database and the Experiments}\label{sec:sample-rne}

We obtain our sample from the Reusing Natural Experiments (RNE) database that accompanies \citet{heath2023reusing}. The
database has one tab for each natural experiment and one row for each paper and outcome variable that uses it.

We study the experiment tabs that list more than one paper and exclude \nRneTabsOutWord{} such tabs whose settings lie
outside finance and accounting. Of the \nRneTabsN{} experiment tabs in the database version we downloaded,
\nRneTabsMultiN{} list more than one paper, and we keep \nCrowdExpN{}. They include state laws adopted at different dates,
pilots of the Securities and Exchange Commission (SEC) that changed trading rules for a subset of stocks, market-wide
rule changes such as Regulation Fair Disclosure (Regulation FD), and events that affect a subset of firms or funds.
Table~\ref{tab:tab_experiments} in Appendix~\ref{app:experiments} describes what changed
and when.

No paper on the Morningstar ratings completes the pipeline (Section~\ref{sec:sample-attrition}), so our final sample
consists of \nPapers{} papers from \nExperiments{} experiments. Of these, \nRnePapers{} are listed in RNE, and
\nAddedPapers{} are papers on the same experiments that we identify through the citation network of the RNE papers
(Section~\ref{sec:sample-added}; Table~\ref{tab:tab_sample}). The Regulation SHO (Reg SHO) pilot and the business combination
laws, the two settings to which \citet{heath2023reusing} apply their multiple testing corrections, contribute \nHeathN{}
papers.

\begin{table}[!htbp]
\centering
\begin{threeparttable}
\caption{Natural experiments and sample papers}
\label{tab:tab_sample}
\small
\setlength{\tabcolsep}{4pt}
\begin{tabular*}{\textwidth}{@{\extracolsep{\fill}}>{\raggedright\arraybackslash}p{1.55in}lrrrrrr}
\toprule
 &  & \makecell{Listed\\in RNE} & \makecell{RNE\\papers} & \makecell{Added\\papers} & \makecell{Alternative\\data} & \makecell{Directional\\claims} & \makecell{Authors'\\materials} \\
Experiment & Years & (1) & (2) & (3) & (4) & (5) & (6) \\
\midrule
Reg SHO pilot & 2005--2007 & 62 & 45 & 0 & 3 & 39 & 3 \\
Business combination laws & 1985--1991 & 33 & 22 & 9 & 3 & 30 & 1 \\
Mutual fund scandal & 2003 & 33 & 20 & 0 & 13 & 19 & 2 \\
Brokerage closures and mergers & 1984--2008 & 24 & 18 & 18 & 14 & 36 & 1 \\
Universal demand laws & 1989--2005 & 20 & 17 & 8 & 5 & 25 & 3 \\
Russell index reconstitution & Annual & 18 & 16 & 22 & 6 & 37 & 3 \\
Decimalization & 2000--2001 & 26 & 15 & 12 & 9 & 26 & 0 \\
Tick Size Pilot & 2016--2018 & 18 & 13 & 7 & 1 & 20 & 1 \\
Regulation FD & 2000 & 21 & 12 & 36 & 16 & 44 & 1 \\
EDGAR phase-in & 1993--1996 & 13 & 10 & 2 & 2 & 11 & 2 \\
BlackRock-BGI acquisition & 2009 & 34 & 6 & 1 & 1 & 6 & 3 \\
Dodd-Frank Act & 2010 & 11 & 6 & 8 & 3 & 12 & 0 \\
EBA capital exercise & 2011--2012 & 7 & 1 & 0 & 1 & 1 & 0 \\
Morningstar sustainability ratings & 2016 & 17 & 0 & 0 & 0 & 0 & 0 \\
\midrule
Total &  & 337 & 201 & 123 & 77 & 306 & 20 \\
\bottomrule
\end{tabular*}
\begin{tablenotes}[flushleft]
\footnotesize
\item[] Columns (4) to (6) cover both the RNE papers and the added papers. Appendix~\ref{app:experiments} describes the experiments.
\end{tablenotes}
\end{threeparttable}
\end{table}

\subsection{Construction of the Sample}\label{sec:sample-attrition}\label{sec:sample-added}

We start with the \nListedN{} entries that RNE lists on the \nCrowdExpN{} experiments, after merging each paper's rows
into one entry per paper and experiment. We also search for papers on the same experiments that cite or are cited by the
RNE papers. Through this citation network, we find \nAddLinkedN{} candidates. Agents screen each candidate using RNE's
inclusion rule: a paper belongs to an experiment when it uses the experiment as a source of exogenous variation for an
estimate, even in a side test. We also require a version in a finance, economics or accounting journal or in
Management Science.

A paper enters the sample when the pipeline of Section~\ref{sec:method} finishes every stage for it, including the final
check of its report. Of the \nListedN{} RNE entries, \nRnePapers{} complete the pipeline, and most of the others drop out
because the data their main estimates require are not available to us. Of the added papers, \nAddedPapers{} complete
the pipeline.
Table~\ref{tab:tab_flow} tracks both sets through each step to the final sample of \nPapers{} papers.

\begin{table}[!htbp]
\centering
\begin{threeparttable}
\caption{Construction of the sample}
\label{tab:tab_flow}
\small
\begin{tabular*}{\textwidth}{@{\extracolsep{\fill}}lr}
\toprule
 & Papers \\
\midrule
\multicolumn{2}{l}{\emph{Panel A. Papers listed in RNE}} \\
Entries listed in RNE (one per paper and experiment) & \nListedN \\
\quad Restricted, proprietary or hand-collected data & \nAttRestrictedN \\
\quad Data available but not held & \nAttNotHeldN \\
\quad No usable main estimate & \nAttNoEstN \\
\quad Other & \nAttOtherN \\
Runs completed: RNE papers in the sample & \nRnePapers \\
\addlinespace
\multicolumn{2}{l}{\emph{Panel B. Papers added from the citation network}} \\
Papers linked to the RNE papers of an experiment & \nAddLinkedN \\
Pass the inclusion screens & \nAddCandN \\
\quad Versions of RNE papers & \nAddVersionsN \\
\quad Full text not obtained & \nAddNoobtainN \\
Enter the pipeline & \nAddRegN \\
\quad Data available but not held & \nAddDownloadN \\
\quad Restricted, proprietary or hand-collected data & \nAddInfeasibleN \\
\quad No usable estimate of the experiment's effect & \nAddNoestN \\
Runs completed: added papers in the sample & \nAddedPapers \\
\midrule
Sample & \nPapers \\
\bottomrule
\end{tabular*}
\begin{tablenotes}[flushleft]
\footnotesize
\item[] The table follows the papers that RNE lists on the \nCrowdExpN{} experiments (Panel A) and the papers we add from the
citation network of the RNE papers (Panel B) to the sample. A paper enters the sample when its run completes
(Section~\ref{sec:sample-attrition}). In Panel A, we started runs for \nAttStartedN{} entries, of which \nAttStoppedN{} stopped for lack of
data; the other \nAttNeverN{} entries were never started. Indented rows count the papers that leave the sample at each step.
Tables~\ref{tab:tab_attrition} and~\ref{tab:tab_attrition_added} give the counts by experiment.
\end{tablenotes}
\end{threeparttable}
\end{table}

Appendix~\ref{app:sample} compares the RNE entries that complete the pipeline with the
others. Because \nAttRestrictedN{} of the RNE entries that do not complete
the pipeline require restricted, proprietary or hand-collected data, papers based on such data are underrepresented in
our sample. The added papers come from \nAddedExpN{} of the experiments. In Appendix~\ref{app:rne}, we repeat the main analyses for
the RNE papers alone.

\subsection{The Main Claim}\label{sec:sample-claim}

We extract one main claim from each paper. Because a paper can estimate effects on several outcomes, each with several
specifications, we fix the main estimate in two steps before any agent estimates an alternative specification.

The first step produces a list of candidate estimates and their locations in the paper. The list comes from the Internet Appendix of
\citet{heath2023reusing} for the \nHeathRneN{} Reg SHO and business combination law papers in our sample that RNE lists
and from an agent for the other papers. For the added papers, which have no RNE outcomes, the agent selects up to three
outcomes that the abstract presents as the main effects of the experiment. For each outcome, the agent locates the table
that the paper presents as its main result and, within that table, the column of the preferred specification
(Appendix~\ref{app:sample-details}).

In the second step, an agent selects one estimate based only on how the paper presents its results. The agent selects
the estimate for the listed outcome that the paper presents as its main result, by convention the outcome in its first
main-results table.

Two agents independently code each main claim as directional, null or unclear. A claim is directional when the paper
reports an effect of a given sign that it treats as statistically significant. A claim is null when the paper reports
that the shock does not affect the outcome. A claim is unclear when neither coder can classify it as directional or
null. Of the \nPapers{} claims, \nDir{} are directional. The rest are \nNull{} null
claims and \nUnclear{} unclear claims. The unclear claims and one directional claim report no test
(Section~\ref{sec:robust}). From here on, we use paper and claim interchangeably.

We focus on the \nDir{} directional claims and discuss the null and unclear claims in Section~\ref{sec:robust-null}.

\subsection{Data}\label{sec:sample-data}

An agent first searches for the authors' data and code (the authors' materials). Unless these materials include both
data and code, agents rebuild the data from the source databases (Section~\ref{sec:method-reproduce}). The agent finds
such materials for \nRepMatFoundN{} papers, \nRepMatDataN{} of which include
data. For the other \nRepMatNoneN{} of the \nPapers{} papers, the agents rebuild the data using the paper and any online
appendix. They also rebuild the data for the papers whose materials include code but no data. The data come from Wharton
Research Data Services (WRDS) and public sources as of 2026. These data are newer vintages of the databases that the
authors used years earlier.

For \nExactDataN{} papers, we take every input to the main estimate from the databases the paper uses or from the
authors' own files (exact data). For the other \nAltDataN, we use other data for at least one input (\emph{alternative
data}). We analyze the version of each paper that we could obtain, and its journal and the count of earlier
papers in Section~\ref{sec:predict} refer to this version.

\section{Multiverse Analysis with AI Agents}\label{sec:method}

\subsection{Overview}\label{sec:method-overview}

\begin{figure}[!htbp]
\centering
\begin{tikzpicture}[
    >={Stealth[length=1.6mm]}, node distance=15mm and 4.5mm,
    gen/.style={rectangle, rounded corners=1.5pt, draw=black!75, line width=0.5pt, fill=white,
                text width=26mm, minimum height=13mm, align=center, inner sep=1.5pt, font=\footnotesize,
                execute at begin node={\hyphenpenalty=10000\exhyphenpenalty=10000}},
    chk/.style={gen, fill=black!13},
    gate/.style={diamond, draw=black!85, line width=0.6pt, fill=white, aspect=1.3, inner sep=0.6pt,
                 font=\scriptsize, minimum width=10mm},
    arr/.style={->, line width=0.5pt, draw=black!80},
    loop/.style={->, line width=0.5pt, draw=black!60, dashed},
    lab/.style={font=\scriptsize, text=black!75, inner sep=1pt, fill=white},
]
\node[gen] (pb) {Rebuild data\\[-1pt]{\scriptsize\itshape\color{black!70} package finder, package builder}};
\node[chk, right=of pb] (sr) {Sample review\\[-1pt]{\scriptsize\itshape\color{black!70} sample reviewer}};
\node[gate, right=of sr] (gs) {GS};
\node[gen, right=of gs] (cf) {Configuration\\[-1pt]{\scriptsize\itshape\color{black!70} configuration drafter}};
\node[chk, right=of cf] (cr) {Configuration review\\[-1pt]{\scriptsize\itshape\color{black!70} configuration reviewer}};
\node[gate, right=of cr] (gc) {GC};
\node[gen, below=19mm of gc] (rp) {Reproduction\\[-1pt]{\scriptsize\itshape\color{black!70} reproducer}};
\node[gate, left=of rp] (g0) {G0};
\node[chk, left=7mm of g0] (ra) {Reproduction audit\\[-1pt]{\scriptsize\itshape\color{black!70} reproduction auditor}};
\node[gen, left=of ra] (fe) {Choice listing\\[-1pt]{\scriptsize\itshape\color{black!70} choice brainstormer}};
\node[chk, left=of fe] (fa) {Choice audit\\[-1pt]{\scriptsize validity ratings, the grid's inference}\\[-1pt]{\scriptsize\itshape\color{black!70} choice auditor}};
\node[gate, left=of fa] (g2) {G2};
\node[gen, below=of g2] (im) {Implementation\\[-1pt]{\scriptsize\itshape\color{black!70} implementer}};
\node[chk, right=of im] (ia) {Implementation audit\\[-1pt]{\scriptsize\itshape\color{black!70} implementation auditor}};
\node[gate, right=of ia] (g4) {G4};
\node[gen, right=of g4] (cg) {Companion grid\\[-1pt]{\scriptsize paper's choices restored}\\[-1pt]{\scriptsize\itshape\color{black!70} companion implementer, companion auditor}};
\node[gate, right=of cg] (g4c) {G4C};
\node[chk, below=of g4c] (ac) {Author commentary\\[-1pt]{\scriptsize\itshape\color{black!70} author-commentary agent}};
\node[gen, left=of ac] (re) {Report\\[-1pt]{\scriptsize\itshape\color{black!70} reporter}};
\node[gate, left=of re] (g6) {G6};
\draw[arr] (pb) -- (sr);
\draw[arr] (sr) -- (gs);
\draw[arr] (gs) -- (cf);
\draw[arr] (cf) -- (cr);
\draw[arr] (cr) -- (gc);
\draw[arr] (gc) -- (rp);
\draw[arr] (rp) -- (g0);
\draw[arr] (g0) -- node[lab, above=0.4mm] {miss} (ra);
\draw[arr] (g0.south) -- ++(0,-4mm) -| node[lab, below, pos=0.25] {pass} (fe.south);
\draw[arr] (ra) -- (fe);
\draw[arr] (fe) -- (fa);
\draw[arr] (fa) -- (g2);
\draw[arr] (g2) -- (im);
\draw[arr] (im) -- (ia);
\draw[arr] (ia) -- (g4);
\draw[arr] (g4) -- (cg);
\draw[arr] (cg) -- (g4c);
\draw[arr] (g4c) -- (ac);
\draw[arr] (ac) -- (re);
\draw[arr] (re) -- (g6);
\draw[loop] (gs.north) to[out=125, in=55] node[lab, above] {fixable: builder-fix agent, review again} (sr.north);
\draw[loop] (gc.north) to[out=125, in=55] node[lab, above] {fail: redraft} (cf.north);
\draw[loop] (ra.north) to[out=35, in=145, looseness=0.8] node[lab, above] {errors: builder-fix agent, reproduce again} ([xshift=-8mm]rp.north);
\draw[loop] (ia.south) to[out=-130, in=-50] node[lab, below] {critical or serious findings: fix-round agent, re-audit} (im.south);
\end{tikzpicture}
\caption{The agentic multiverse pipeline as run for each paper}
\label{fig:fig_pipeline}
\par\vspace{0.4em}
\begin{minipage}{\textwidth}\footnotesize\singlespacing
White boxes are roles that produce files, and shaded boxes roles that check them. Diamonds are the script checks described in Appendix~\ref{app:pipeline}, and dashed arrows are fix loops.
\end{minipage}
\end{figure}
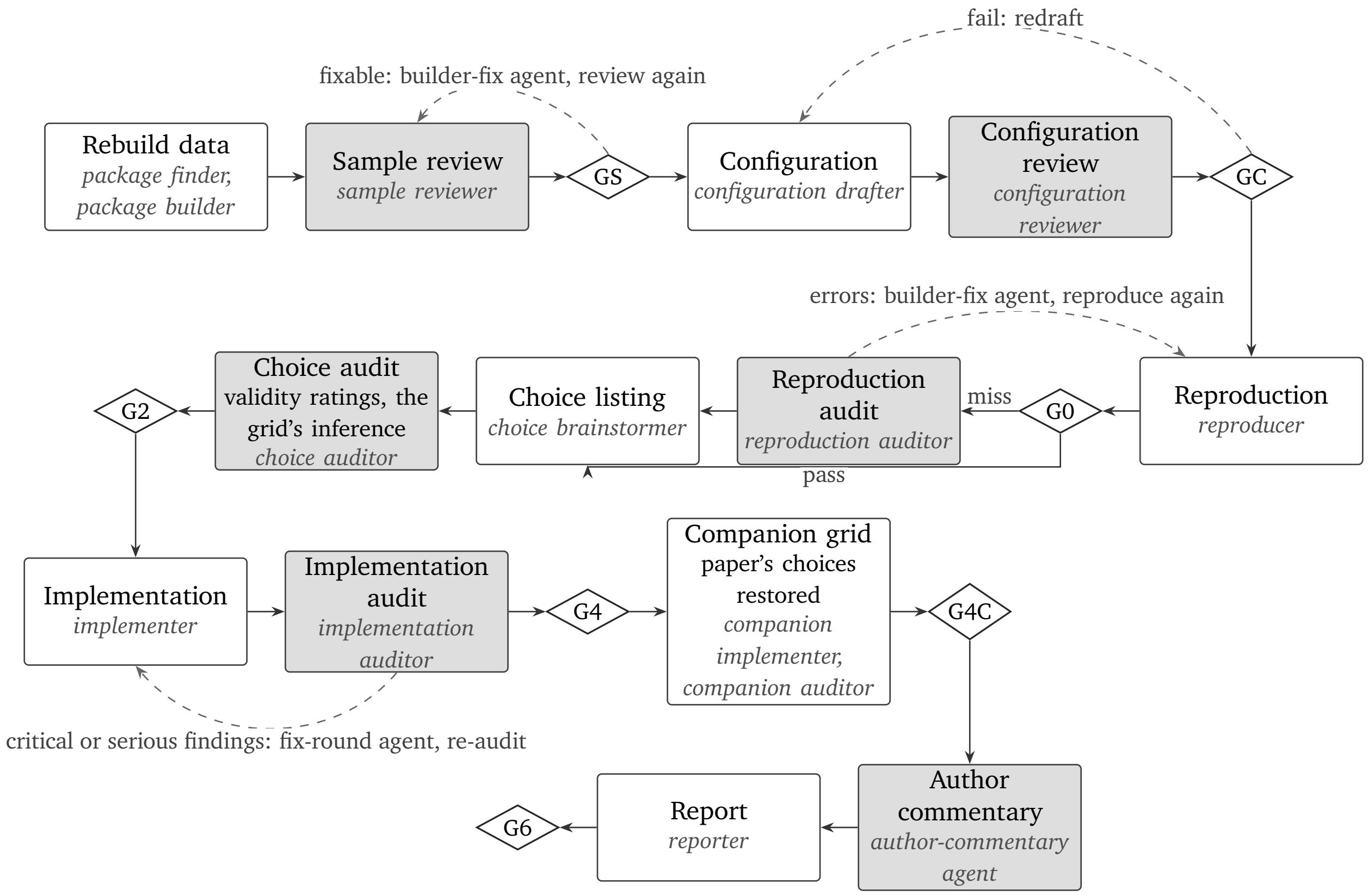

AI agents carry out the multiverse analysis for each paper. They first reproduce the paper's main estimate. They then
build a grid of alternative specifications, estimate each one and report how many keep the paper's directional claim.
Figure~\ref{fig:fig_pipeline} shows the steps, which are the same for all \nPapers{} papers.

No agent checks its own work, and a script checks the files from each step before the next step begins. The agents use
Claude Sonnet 5.5 (Sonnet) and Claude Opus 5.5 (Opus).

We restrict the information available to the agents so that they cannot choose analyses based on the results: the
agents that rebuild and review the data may not regress any outcome on the treatment, and an agent fixes the grid and its statistical inference before any agent estimates an alternative specification. No agent may read a third-party critique,
comment, replication or reanalysis of the paper it analyzes.

The repository at \RepoURL{} documents every step, prompt and script. Appendix~\ref{app:pipeline}
describes the pipeline in detail.

\subsection{Reproducing the Main Estimate}\label{sec:method-reproduce}

When the authors' materials include both data and code, the agents use them to reproduce the main estimate. Otherwise,
an agent rebuilds the data from the sources described in the paper, mostly WRDS and public databases
(Section~\ref{sec:sample-data}), and other agents check it against the paper's sample rules and observation counts and
fix the discrepancies they can resolve.

An agent then estimates the paper's main specification. If this estimate differs from the published estimate by more
than 10\% and by more than half a unit of its last printed decimal, a reproduction auditor, i.e., an agent that checks the data and code for errors, reviews the reproduction.
The auditor may not search for alternative specifications that produce estimates closer to the published one. Our
audited reproduction is the starting point for the paper's grid.

A main estimate is considered \emph{reproduced} when our estimate meets three conditions. First, it has the published
sign. Second, it has a two-sided $p$-value of at most 0.10 under the paper's own statistical inference, i.e., the standard errors or other inference method the paper reports, applied to our data. Third, it is within the closeness bound, i.e., its magnitude differs from the
published estimate by no more than 25\% (or by less than 0.10 standard deviations of the outcome in linear models).

\subsection{Building the Grid}\label{sec:method-forks}

An agent reads the paper, the data and the reproduction and lists every analysis choice that can affect the main
estimate, such as the sample filters, the event window, the estimator and the controls. For each choice, it lists the
options a careful analyst could take, including the paper's own. The median paper has \nForksMedian{} choices and
\nOptionsMedian{} options. For example, when states adopt a law in different years, the options for the estimator
include three estimators designed for such staggered adoption: the group-time estimator of
\citet{callaway2021difference}, the interaction-weighted event-study estimator of \citet{sun2021estimating} and the
imputation estimator of \citet{borusyak2024revisiting}. The agent may also list the two-way fixed effects regression.

A second agent, the choice auditor, rates every option against a written rubric based on the three conditions of
\citet{simonsohn2020specification}: a specification must be consistent with the paper's claim, statistically valid and
non-redundant with the other specifications. The auditor rates each option \emph{valid}, \emph{questionable} (arguable,
kept and flagged) or \emph{invalid} (Table~\ref{tab:tab_validation}). Examples of the usual cases for an \emph{invalid} rating in the rubric include
clustering below the level of treatment assignment, controlling for a variable that the treatment affects, using
already-treated units as controls under staggered adoption, using conventional cluster-robust inference with few
clusters, and ignoring a known randomization. In the staggered-adoption example above, the estimate of the two-way fixed
effects regression is a weighted average of all two-group, two-period comparisons, including comparisons that use
already-treated units as controls \citep{goodmanbacon2021difference}. The rubric therefore gives grounds to rate this
estimator \emph{invalid}. The other three estimators avoid such comparisons. The auditor judges the paper's own options
by the same standard as any other option.

\begin{table}[!htbp]
\centering
\begin{threeparttable}
\caption{Audits inside the pipeline and what they find}
\label{tab:tab_validation}
\footnotesize
\setlength{\tabcolsep}{5pt}
\begin{tabular}{lccc}
\toprule
 & \makecell{All\\(324)} & \makecell{Exact data\\(247)} & \makecell{Alternative data\\(77)} \\
\midrule
\multicolumn{4}{l}{\textit{Panel A. Implementation audits (check G4)}} \\
Audit rounds & 535 & 420 & 115 \\
Critical / serious findings in round 1 & 67 / 371 & 51 / 317 & 16 / 54 \\
Papers with a critical / a serious finding in round 1 & 49 / 186 & 38 / 152 & 11 / 34 \\
Critical / serious findings open at the end & 0 / 108 & 0 / 92 & 0 / 16 \\
Papers with an open critical / serious finding at the end & 0 / 69 & 0 / 58 & 0 / 11 \\
\addlinespace
\multicolumn{4}{l}{\textit{Panel B. Reproduction audits (after a G0 miss)}} \\
Papers audited & 271 & 205 & 66 \\
Papers with an error found or the miss unresolved & 201 & 144 & 57 \\
\addlinespace
\multicolumn{4}{l}{\textit{Panel C. Sample reviews (check GS)}} \\
Papers / items fixable at first review & 246 / 576 & 177 / 426 & 69 / 150 \\
Papers / items still fixable at the end & 115 / 167 & 73 / 110 & 42 / 57 \\
\addlinespace
\multicolumn{4}{l}{\textit{Panel D. Choice audit (check G2)}} \\
Options rated & 69,189 & 52,218 & 16,971 \\
Share rated \emph{valid} & 0.52 & 0.52 & 0.53 \\
Share rated \emph{questionable} & 0.27 & 0.27 & 0.26 \\
Share rated \emph{invalid} & 0.21 & 0.21 & 0.21 \\
\addlinespace
\multicolumn{4}{l}{\textit{Panel E. Reliability of coded variables and of the outcomes}} \\
Data class, Cohen's kappa (two coders, 324 papers) & 0.86 &  &  \\
Claim type, share coded alike (two coders, 324 papers) & 1.00 &  &  \\
Order of the outcome, Krippendorff's alpha (three coders, 324 papers) & 0.86 &  &  \\
Share supported, split-half reliability (\nRelAllN{} papers) & \nRelAllShare &  &  \\
Median signed $z$, split-half reliability (\nRelAllN{} papers) & \nRelAllMedz &  &  \\
\bottomrule
\end{tabular}
\begin{tablenotes}[flushleft]
\footnotesize
\item[] Script checks and audit findings as defined in Appendix~\ref{app:pipeline}, and option ratings as defined in Section~\ref{sec:method}. Items are sample gaps, and the split-half rows cover all papers with a numeric published estimate.
\end{tablenotes}
\end{threeparttable}
\end{table}

Some options change what the estimate measures, so that its interpretation differs from the paper's claim. For example,
a paper may study the short-term response to a shock, while the agent that lists the choices proposes a longer event
window. The auditor marks such options, and we exclude their specifications from the \emph{share supported} measure,
i.e., the share of specifications that keep the paper's directional claim (Section~\ref{sec:method-measure}).

The auditor then builds the grid, i.e., the full set of specifications, in three steps. First, it selects the four to
eight choices that it judges most consequential for the result. Second, for each of these choices, it keeps the options
it rated \emph{valid} or \emph{questionable} and combines them in every possible way. Each combination is one
specification. Third, for each of the other choices, it picks one option, typically the paper's own, and uses that
option in every specification. It writes down why it picked that option. The median grid varies six choices and has \nGridMedian{}
specifications, within the allowed range of 100 to 250.

\label{sec:method-inference}Before any agent estimates the grid, the auditor also selects one statistical inference method for the main estimate. The same statistical inference applies to every specification in the grid. The rubric ties the statistical inference to how the experiment assigns treatment. The auditor generally selects standard errors clustered at the level of
treatment assignment \citep{abadie2023when}. When clusters are few or unbalanced, as with state laws adopted at
different dates, it selects a wild cluster bootstrap because conventional cluster-robust tests reject the null too often
\citep{cameron2008bootstrap,mackinnon2017wild}. For randomized experiments, such as the Reg SHO pilot and the Tick Size
Pilot, the auditor selects randomization inference based on the actual treatment assignment \citep{young2019channeling}.

\subsection{Estimating and Checking the Grid}\label{sec:method-audits}

An agent writes and runs a script that estimates every specification under the grid's statistical inference. A separate agent, the
implementation auditor, checks the script against the choice auditor's ratings, the grid and the data, and another
agent fixes the errors it finds until the audit passes and a script confirms that the grid is complete.

The choice auditor can exclude a paper's own option by rating it \emph{invalid} or by holding the choice fixed at
another option. Another agent therefore estimates a companion grid, i.e., the same grid with the paper's own options
restored wherever the auditor excluded them, under the same statistical inference. We compare the two grids in
Section~\ref{sec:robust-companion}. Finally, an agent picks the riskiest choices, i.e., the choices that most threaten
the paper's claim, and quotes what the paper says about them (Section~\ref{sec:robust-commentary}).

\subsection{Outcomes}\label{sec:method-measure}

We summarize each paper with four outcomes: the share of its specifications that keep its directional claim, the share
of its specifications whose estimates have the published sign, whether we reproduce its main estimate, and the median
signed $z$-statistic across its specifications, computed from the grid $p$-values. We call these outcomes \emph{share
supported}, \emph{share with the published sign}, \emph{reproduced} and \emph{median signed $z$}.

\paragraph{\emph{Share supported}.} A specification $c$ of paper $i$ keeps the directional claim when its estimate
$b_{ic}$ has the sign of the published main estimate $\hat{\beta}_i$ and its two-sided $p$-value under the grid's statistical inference,
$p_{ic}$, is below 0.10. The set $\mathcal{C}_i$ holds the $N_i$ specifications of paper $i$ whose
estimation succeeds and that use no option that changes what the estimate measures. The \emph{share supported} of paper
$i$ is the share of these specifications that keep the directional claim (Appendix~\ref{app:measure} in the Internet
Appendix gives the notation in full):
\begin{equation}
S_i = \frac{1}{N_i} \sum_{c \in \mathcal{C}_i}
\mathbf{1}\{\operatorname{sign}(b_{ic}) = \operatorname{sign}(\hat{\beta}_i)\} \cdot
\mathbf{1}\{p_{ic} < 0.10\}.
\label{eq:share}
\end{equation}
The two conditions amount to a one-sided test at the 5\% level in the published direction. We impose no condition on
the size of the estimate because some grid options change its units, such as a log outcome in place of a level. The
\nCellsAllPapers{} papers with a numeric published estimate have \nCellsAll{} specifications, \nCells{} of them for the
\nDir{} directional claims. We adjust the $p$-values of a few grids whose hypothesis test is one-sided or joint or whose claim
compares the estimate with a nonzero benchmark (Appendix~\ref{app:measure}).

\paragraph{Share with the published sign.} The \emph{share with the published sign} is a looser version of the
\emph{share supported}: it counts the specifications whose estimates have the published sign, regardless of their
$p$-values. It equals equation~(\ref{eq:share}) without the factor $\mathbf{1}\{p_{ic} < 0.10\}$, so it is never below
the \emph{share supported}.

\paragraph{Reproduced.} A main estimate is \emph{reproduced} when our estimate meets the conditions in
Section~\ref{sec:method-reproduce}. This outcome concerns one specification, our implementation of the paper's own. \emph{Reproduction} uses the
paper's own statistical inference, i.e., the standard errors or other inference method the paper reports. The \emph{share supported}
uses the statistical inference that the choice auditor fixes for the grid. A \emph{reproduced} main estimate therefore
need not keep the directional claim under the grid's inference.

\paragraph{Median signed $z$.} The \emph{median signed $z$} is, in effect, the median $t$-statistic across a claim's
specifications, positive when the estimate has the published sign and negative when it has the opposite sign. We compute
it from $p$-values because the wild cluster bootstrap and randomization inference, used for \nInfWildPct{} and
\nInfRandPct{} of the papers, give a $p$-value but no $t$-statistic. For each specification, we take the two-sided grid
$p$-value, find the $z$-statistic that gives the same $p$-value under the standard normal distribution, and give it a
positive sign when the estimate has the published sign and a negative sign otherwise,
\begin{equation}
z_{ic} = \operatorname{sign}(b_{ic})\operatorname{sign}(\hat{\beta}_i)\,
\Phi^{-1}\!\left(1 - \tfrac{1}{2}\max(p_{ic}, 10^{-4})\right),
\label{eq:medz}
\end{equation}
where $\Phi$ is the standard normal distribution function (Appendix~\ref{app:measure} explains the
floor on the $p$-value). The \emph{median signed $z$} of paper $i$ is the median of $z_{ic}$ over its specifications. A
value above 1.645 indicates that the median specification keeps the directional claim, and a negative value indicates
an estimate with the opposite sign.

\paragraph{Roles of the four outcomes.} The \emph{share supported} is our primary outcome because it is easy to
interpret: it measures the share of defensible specifications that keep the directional claim. Its cutoff discards the
information in $p$-values above and below 0.10. The \emph{median signed $z$} retains this information and distinguishes
among the claims with a \emph{share supported} of zero. The \emph{share with the published sign} separates the
direction of the estimates from their precision. \emph{Reproduction} checks the starting point of each grid.

\section{Reproducing the Main Estimate}\label{sec:repro}

Table~\ref{tab:tab_reproduction} reports the reproduction results. We reproduce \nRepAllK{} of the main estimates of the
\nRepAllN{} directional claims with a published test (\nRepAllPct). The most common failing condition is the $p$-value: of the
\nRepMiss{} claims not reproduced, \nRepFailP{} (\nRepFailPPct) have a $p$-value above 0.10 (Panel~B of
Table~\ref{tab:tab_reproduction}; Appendix~\ref{app:results45-details} gives the full breakdown
and a looser $p$-value cutoff). Across the directional claims, the median absolute difference between our
estimate and the published estimate is \nRepMedDevPct{} of the published value.

\begin{table}[!htbp]
\centering
\begin{threeparttable}
\caption{Reproduction of the main estimates}
\label{tab:tab_reproduction}
\small
\begin{tabular*}{\textwidth}{@{\extracolsep{\fill}}lcccc}
\toprule
 & \multicolumn{2}{c}{Reproduced} & \multicolumn{2}{c}{$p < 0.20$ variant} \\
\cmidrule(lr){2-3}\cmidrule(lr){4-5}
 & Papers & Share & Papers & Share \\
\midrule
\multicolumn{5}{l}{\textit{Panel A. Reproduction rate}} \\
All directional claims & 128/305 & 0.420 & 144/305 & 0.472 \\
Exact data & 104/232 & 0.448 & 116/232 & 0.500 \\
Alternative data & 24/73 & 0.329 & 28/73 & 0.384 \\
Authors' materials & 15/18 & 0.833 & 15/18 & 0.833 \\
No authors' materials & 113/287 & 0.394 & 129/287 & 0.449 \\
\addlinespace
Alternative minus exact data, FE & 304 & $-$0.053 & 304 & $-$0.025 \\
 &  & (0.070) &  & (0.071) \\
Authors' materials minus none, FE & 304 & 0.464$^{***}$ & 304 & 0.405$^{***}$ \\
 &  & (0.125) &  & (0.127) \\
\addlinespace
\multicolumn{5}{l}{\textit{Panel B. Conditions not met (share of papers not reproduced)}} \\
Papers not reproduced & 177 &  & 161 &  \\
Fail closeness & 115 & 0.650 & 115 & 0.714 \\
Fail sign & 60 & 0.339 & 60 & 0.373 \\
Fail $p$-value & 137 & 0.774 & 113 & 0.702 \\
Closeness only & 35 & 0.198 & 36 & 0.224 \\
Sign only & 0 & 0.000 & 1 & 0.006 \\
$p$-value only & 49 & 0.277 & 33 & 0.205 \\
Closeness and sign only & 5 & 0.028 & 11 & 0.068 \\
Closeness and $p$-value only & 33 & 0.186 & 32 & 0.199 \\
Sign and $p$-value only & 13 & 0.073 & 12 & 0.075 \\
All three & 42 & 0.237 & 36 & 0.224 \\
\addlinespace
\multicolumn{5}{l}{\textit{Panel C. Sample size}} \\
Not reproduced, sample size off by more than 1\% & 141/163 & 0.865 & 130/150 & 0.867 \\
Spearman correlation, sample-size gap and estimate gap & 278 & 0.282 & 278 & 0.282 \\
\bottomrule
\end{tabular*}
\begin{tablenotes}[flushleft]
\footnotesize
\item[] Rows marked FE estimate the difference by OLS with experiment fixed effects. Heteroskedasticity-robust (HC3) standard errors are in parentheses, and *, ** and *** denote significance at the 10\%, 5\% and 1\% levels.
\end{tablenotes}
\end{threeparttable}
\end{table}

We reproduce \nRepExactPct{} of the claims with exact data and \nRepAltPct{} of the claims with \emph{alternative data}.
We reproduce the main estimate more often when we find the authors' materials: \nRepFoundPct{} (\nRepFoundK{} of \nRepFoundN) of the
directional claims with materials, compared with \nRepRebuiltPct{} without (Section~\ref{sec:predict-mat} reports the within-experiment
estimate).\footnote{For some papers, the authors share their code on the journal's website, but the website
blocks the agents' access. We count the authors' materials as missing for these papers.}

The most common difference between an estimate we do not reproduce and the published analysis lies in the sample,
measured by the number of observations. Among the claims not reproduced for which both observation counts are available,
our number of observations differs from the published number by more than 1\% in \nRepMissNDiff{} of \nRepMissNComp{}
claims (\nRepMissNDiffPct). The observation counts also differ for some \emph{reproduced} main estimates. Across the
\nRepSpearN{} papers with both counts, a larger difference in sample size is associated with a larger difference between
our estimate and the published one (Spearman correlation \nRepSpearNgap, \nRepSpearNgapP). This comparison is
approximate because some papers report the number of firms or pairs.

\section{Robustness across Specifications}\label{sec:robust}

In this section, we compare the outcomes across \nReg{} of the \nDir{} directional claims. We exclude one claim with no
published test. We also exclude the one paper whose experiment no other paper in our sample uses, because the
within-experiment regressions in Section~\ref{sec:predict} cannot use it and we keep one sample for all analyses. The remaining \nReg{} claims come from \nExpK{} experiments, each with at least three directional claims, and form
the regression sample in Section~\ref{sec:predict}. Counts of
specifications use all \nDir{} claims.

\subsection{The Distribution of the Outcomes}\label{sec:robust-dist}

The mean \emph{share supported} is \nShareMeanPct{} and the median is \nShareMedianPct{} (Panel~C of
Table~\ref{tab:tab_outcomes}). For \nShareZeroN{} of the \nReg{} claims (\nShareZeroPct), it is zero. Among all \nDir{} directional claims, at least half of the specifications keep the
directional claim for \nShareHalfN, and all of them do for \nShareAllN.

\begin{table}[!htbp]
\centering
\begin{threeparttable}
\caption{The outcomes: correlations and distribution}
\label{tab:tab_outcomes}
\small
\begin{tabular*}{\textwidth}{@{\extracolsep{\fill}}lcccc}
\toprule
 & Share supported & \makecell{Share with the\\published sign} & Reproduced & Median signed $z$ \\
 & (1) & (2) & (3) & (4) \\
\midrule
\textit{Panel A. Pearson correlation} & & & & \\
Share supported & 1 &  &  &  \\
Share with the published sign & 0.638 & 1 &  &  \\
Reproduced & 0.464 & 0.493 & 1 &  \\
Median signed $z$ & 0.888 & 0.842 & 0.500 & 1 \\
\addlinespace
\textit{Panel B. Spearman correlation} & & & & \\
Share supported & 1 &  &  &  \\
Share with the published sign & 0.747 & 1 &  &  \\
Reproduced & 0.495 & 0.517 & 1 &  \\
Median signed $z$ & 0.873 & 0.919 & 0.544 & 1 \\
\addlinespace
\textit{Panel C. Distribution} & & & & \\
Mean & 0.252 & 0.685 & 0.421 & 0.768 \\
Standard deviation & 0.292 & 0.292 & 0.495 & 1.192 \\
Median & 0.145 & 0.771 & 0.000 & 0.704 \\
Share of claims at zero & 0.234 & 0.013 & 0.579 &  \\
Papers & 304 & 304 & 304 & 304 \\
\bottomrule
\end{tabular*}
\begin{tablenotes}[flushleft]
\footnotesize
\item[] Directional claims in the regression sample (Section~\ref{sec:robust}). Outcomes as defined in Section~\ref{sec:method-measure}.
\end{tablenotes}
\end{threeparttable}
\end{table}

The \emph{median signed $z$} has a mean of \nMedzMean{} and a median of \nMedzMedian, both below 1.645. For
\nMedzNegPct{} of the \nReg{} claims, the median specification has the opposite sign from the published estimate.

The two grid outcomes move together closely, with a correlation of \nCorrShareMedz{} between the \emph{share supported}
and the \emph{median signed $z$}, and \emph{reproduction} moves with them less (Table~\ref{tab:tab_outcomes};
Appendix~\ref{app:results45-details}).

We also find that estimates keep the published sign more often than they keep statistical significance. The mean
\emph{share with the published sign} is \nSignMeanPct{} and the median is \nSignMedianPct, compared with
\nShareMeanPct{} and \nShareMedianPct{} for the \emph{share supported}.

Claims whose main estimate we reproduce are more robust than the other claims. Among the \nReg{} claims, the
\nShareRepN{} \emph{reproduced} claims have a mean \emph{share supported} of \nShareRepMeanPct{} and a mean \emph{median
signed $z$} of \nMedzRepMean. The other \nShareNotRepN{} claims have corresponding means of \nShareNotRepMeanPct{} and
\nMedzNotRepMean. For claims we do not reproduce, however, the outcomes also reflect the gap between our estimate of the
paper's own specification and the published estimate.

\subsection{Sign and Significance across Specifications}\label{sec:robust-cells}

To examine why specifications fail to keep the directional claim, we classify each of the \nCells{} specifications of
the directional claims first by sign and then by $p$-value. Table~\ref{tab:tab_cells} presents the results. The estimate
has the opposite sign in \nCellOppPct{} of specifications. Another \nCellPfailPct{} have the published sign but a
$p$-value of 0.10 or more, and \nCellSupPct{} keep the directional claim. Among specifications with the published sign,
\nCellPfailAmongSignPct{} have a $p$-value of 0.10 or more. Weighting each paper equally gives nearly the same shares
(second column of Table~\ref{tab:tab_cells}).

\begin{table}[!htbp]
\centering
\begin{threeparttable}
\caption{Sign and $p$-value of the specifications of directional claims}
\label{tab:tab_cells}
\small
\begin{tabular*}{\textwidth}{@{\extracolsep{\fill}}lcc}
\toprule
 & Pooled over specifications & Mean over papers \\
\midrule
Opposite sign & 0.32 & 0.31 \\
Published sign, $p \geq 0.10$ & 0.43 & 0.43 \\
Published sign, $p < 0.10$ (keeps the directional claim) & 0.25 & 0.25 \\
\addlinespace
Specifications & 59,106 & 59,106 \\
Papers & 306 & 306 \\
\bottomrule
\end{tabular*}
\begin{tablenotes}[flushleft]
\footnotesize
\item[] Specifications of the main grids, classified by sign and then by the $p$-value under the grid's statistical inference (Section~\ref{sec:method-measure}). The first column pools all specifications, so papers with larger grids carry more weight. The second column computes the shares within each paper and averages them across papers, so each paper counts equally.
\end{tablenotes}
\end{threeparttable}
\end{table}

\subsection{Experiments}\label{sec:robust-exp}

\begin{sidewaysfigure}
\centering
\includegraphics[width=\linewidth]{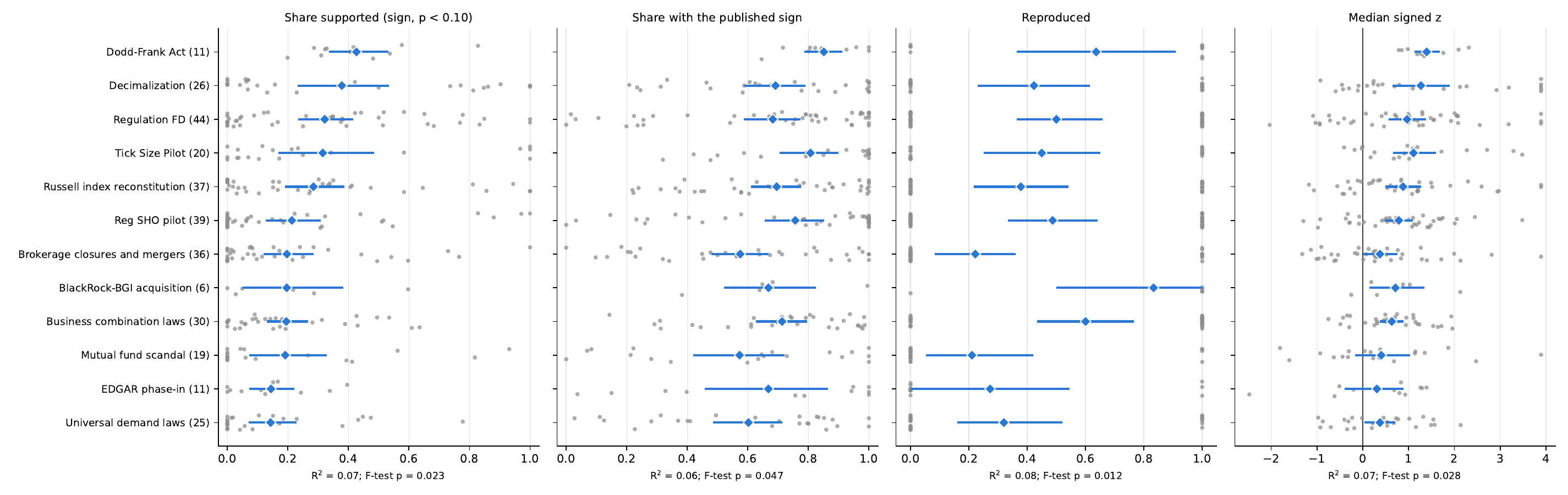}
\caption{Share supported, share with the published sign, reproduction and median signed $z$ by natural experiment}
\label{fig:fig_experiments}
\par\vspace{0.4em}
\begin{minipage}{\linewidth}\footnotesize\singlespacing
The \nReg{} directional claims of the regression sample (Section~\ref{sec:robust}) in its \nExpK{} experiments, with the count for each experiment in parentheses. Dots are claims (jittered), diamonds experiment means, and bars their 95\% bootstrap confidence intervals.
\end{minipage}
\end{sidewaysfigure}

The experiments differ in the \emph{share supported}, the \emph{share with the published sign}, \emph{reproduction} and the \emph{median signed $z$} (Appendix~\ref{app:results45-details} reports
the tests). Figure~\ref{fig:fig_experiments} plots the \nReg{} claims in our regressions by experiment. Across the
\nExpK{} experiments, the mean \emph{share supported} ranges from \nShareExpMinPct{} for \nShareExpMinName{} to
\nShareExpMaxPct{} for \nShareExpMaxName{} (Appendix Table~\ref{tab:tab_experiments} describes each experiment). The Reg SHO pilot, the finance and accounting experiment that RNE lists most often, has a mean \emph{share supported} of
\nExpRegshoSharePct{} across its \nExpRegshoN{} directional claims, compared with the sample mean of \nShareMeanPct. The
Reg SHO pilot is also the subject of a reanalysis project on specification choice
\citep{black2019preanalysis,black2024sec}.

\subsection{Null and Unclear Claims}\label{sec:robust-null}

The unclear claims report no test, so their \emph{reproduction} is not defined. For a null claim, a low \emph{share
supported} agrees with the paper's finding of no effect. The null claims have a mean \emph{share supported} of
\nNullShareMeanPct, and none exceeds \nNullShareMaxPct. The mean absolute \emph{median signed $z$} is \nNullAbsMedz. For
a null claim, we consider the main estimate reproduced when our estimate of the paper's own specification has a
$p$-value of 0.10 or more, as the paper reports. By this standard, we reproduce \nNullBigPN{} of the \nNull{} null
claims (Appendix~\ref{app:results45-details} classifies the others).

\section{Paper Characteristics and the Outcomes}\label{sec:predict}

\begin{figure}[!htbp]
\centering
\includegraphics[width=\textwidth]{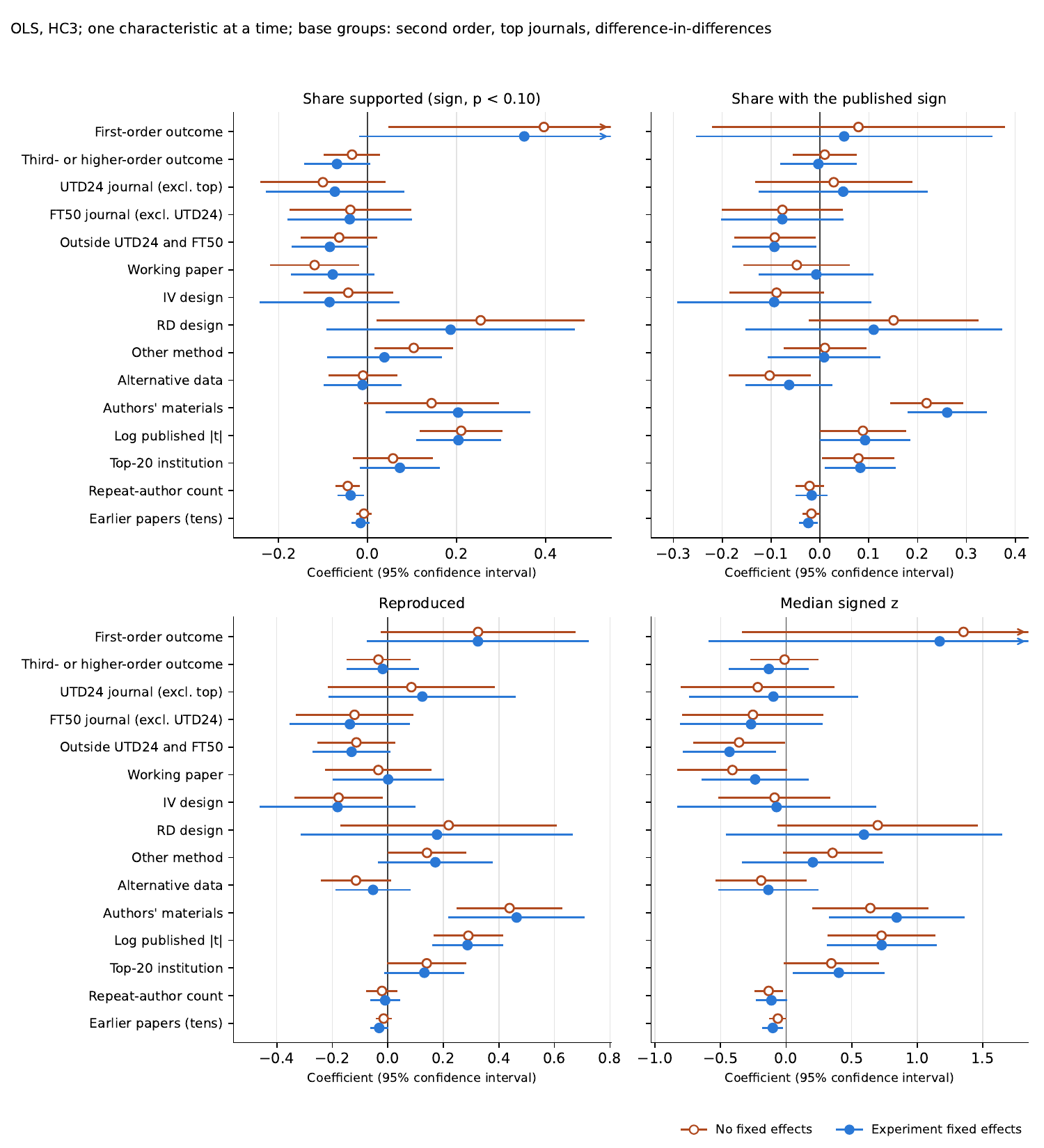}
\caption{Paper characteristics and the outcomes, one characteristic at a time}
\label{fig:fig_coefs}
\par\vspace{0.4em}
\begin{minipage}{\textwidth}\footnotesize\singlespacing
Directional claims (\nReg{} papers), with the characteristics of Section~\ref{sec:predict-chars}. Arrows mark confidence intervals that extend beyond the axis.
\end{minipage}
\end{figure}

In this section, we examine how the four outcomes relate to paper
characteristics, including the order of the outcome studied, journal, research design and authors. The unit of
observation is a directional claim, and the sample consists of the \nReg{} claims of Section~\ref{sec:robust}. We
estimate the following model by ordinary least squares (OLS):
\begin{equation}
Y_i = \alpha_{e(i)} + X_i'\beta + \varepsilon_i,
\label{eq:predict}
\end{equation}
where $Y_i$ is an outcome of claim $i$ (one of the
four outcomes), $X_i$ contains the characteristics defined in
Section~\ref{sec:predict-chars}, $\beta$ contains their coefficients, $\alpha_{e(i)}$ is a fixed effect for the
experiment $e(i)$ used by claim $i$, and $\varepsilon_i$ is the error term. We estimate Equation~(\ref{eq:predict}) with
each characteristic separately (Figure~\ref{fig:fig_coefs}) and with all characteristics together
(Table~\ref{tab:tab_joint}). We estimate both versions without experiment fixed effects (across experiments) and with
them (within experiments). With experiment fixed effects, we compare papers that use the same natural experiment. We use
heteroskedasticity-robust standard errors and report coefficients on the two shares and on \emph{reproduction} in
percentage points.

\begin{table}[!htbp]
\centering
\begin{threeparttable}
\caption{Paper characteristics and the outcomes, jointly}
\label{tab:tab_joint}
\footnotesize
\setlength{\tabcolsep}{4pt}
\begin{tabular*}{\textwidth}{@{\extracolsep{\fill}}lcccccccc}
\toprule
 & \multicolumn{2}{c}{\makecell{Share\\supported}} & \multicolumn{2}{c}{\makecell{Share with the\\published sign}} & \multicolumn{2}{c}{Reproduced} & \multicolumn{2}{c}{\makecell{Median\\signed $z$}} \\
\cmidrule(lr){2-3}\cmidrule(lr){4-5}\cmidrule(lr){6-7}\cmidrule(lr){8-9}
 & (1) & (2) & (3) & (4) & (5) & (6) & (7) & (8) \\
\midrule
First-order outcome & 0.378$^{**}$ & 0.324$^{**}$ & 0.075 & 0.020 & 0.319$^{*}$ & 0.312$^{*}$ & 1.297$^{*}$ & 1.055 \\
 & (0.150) & (0.148) & (0.140) & (0.139) & (0.182) & (0.175) & (0.765) & (0.761) \\
Third or higher order & $-$0.007 & $-$0.040 & 0.034 & 0.017 & 0.041 & 0.031 & 0.088 & $-$0.042 \\
 & (0.038) & (0.040) & (0.037) & (0.040) & (0.063) & (0.067) & (0.155) & (0.158) \\
UTD24 (excl. top) & 0.009 & 0.039 & 0.107 & 0.156$^{*}$ & 0.226 & 0.297 & 0.205 & 0.428 \\
 & (0.085) & (0.088) & (0.090) & (0.092) & (0.174) & (0.189) & (0.351) & (0.355) \\
FT50 (excl. UTD24) & 0.019 & 0.019 & $-$0.013 & 0.006 & $-$0.009 & $-$0.010 & 0.028 & 0.059 \\
 & (0.067) & (0.068) & (0.072) & (0.074) & (0.111) & (0.110) & (0.290) & (0.294) \\
Outside UTD24 and FT50 & $-$0.005 & $-$0.003 & $-$0.022 & 0.001 & 0.002 & 0.033 & $-$0.083 & $-$0.038 \\
 & (0.048) & (0.049) & (0.053) & (0.057) & (0.082) & (0.082) & (0.215) & (0.226) \\
Working paper & $-$0.064 & $-$0.024 & $-$0.001 & 0.049 & 0.079 & 0.115 & $-$0.189 & 0.019 \\
 & (0.049) & (0.049) & (0.059) & (0.067) & (0.096) & (0.104) & (0.216) & (0.234) \\
IV design & $-$0.004 & $-$0.002 & $-$0.079 & $-$0.021 & $-$0.138 & $-$0.046 & 0.015 & 0.260 \\
 & (0.060) & (0.089) & (0.056) & (0.110) & (0.087) & (0.130) & (0.257) & (0.423) \\
RD design & 0.191$^{*}$ & 0.162 & 0.093 & 0.111 & 0.164 & 0.201 & 0.392 & 0.473 \\
 & (0.105) & (0.120) & (0.100) & (0.141) & (0.175) & (0.217) & (0.368) & (0.494) \\
Other method & 0.080$^{*}$ & 0.029 & 0.032 & $-$0.019 & 0.163$^{**}$ & 0.150 & 0.329$^{*}$ & 0.098 \\
 & (0.045) & (0.068) & (0.045) & (0.064) & (0.078) & (0.119) & (0.198) & (0.280) \\
Alternative data & $-$0.050 & $-$0.053 & $-$0.125$^{***}$ & $-$0.096$^{**}$ & $-$0.164$^{***}$ & $-$0.108 & $-$0.352$^{**}$ & $-$0.319$^{*}$ \\
 & (0.036) & (0.039) & (0.043) & (0.046) & (0.063) & (0.068) & (0.165) & (0.179) \\
Authors' materials & 0.132$^{*}$ & 0.165$^{*}$ & 0.208$^{***}$ & 0.260$^{***}$ & 0.421$^{***}$ & 0.449$^{***}$ & 0.573$^{***}$ & 0.747$^{***}$ \\
 & (0.076) & (0.084) & (0.043) & (0.051) & (0.122) & (0.143) & (0.218) & (0.261) \\
Log published $|t|$ & 0.173$^{***}$ & 0.179$^{***}$ & 0.077 & 0.087$^{*}$ & 0.247$^{***}$ & 0.265$^{***}$ & 0.635$^{***}$ & 0.681$^{***}$ \\
 & (0.047) & (0.048) & (0.048) & (0.048) & (0.068) & (0.068) & (0.218) & (0.220) \\
Top-20 institution & 0.005 & 0.017 & 0.015 & 0.024 & 0.058 & 0.048 & 0.098 & 0.147 \\
 & (0.045) & (0.046) & (0.044) & (0.046) & (0.080) & (0.080) & (0.188) & (0.188) \\
Repeat-author count & $-$0.038$^{**}$ & $-$0.037$^{**}$ & $-$0.029$^{*}$ & $-$0.027 & $-$0.020 & $-$0.019 & $-$0.133$^{**}$ & $-$0.141$^{**}$ \\
 & (0.016) & (0.017) & (0.017) & (0.018) & (0.027) & (0.027) & (0.063) & (0.068) \\
Earlier papers (tens) & $-$0.013 & $-$0.018 & $-$0.020$^{*}$ & $-$0.029$^{**}$ & $-$0.016 & $-$0.030 & $-$0.073$^{*}$ & $-$0.110$^{**}$ \\
 & (0.010) & (0.012) & (0.011) & (0.012) & (0.017) & (0.019) & (0.041) & (0.047) \\
\addlinespace
\midrule
Experiment fixed effects & No & Yes & No & Yes & No & Yes & No & Yes \\
Papers & 304 & 304 & 304 & 304 & 304 & 304 & 304 & 304 \\
$R^2$ & 0.18 & 0.23 & 0.12 & 0.18 & 0.16 & 0.22 & 0.15 & 0.22 \\
\bottomrule
\end{tabular*}
\begin{tablenotes}[flushleft]
\footnotesize
\item[] Characteristics as defined in Section~\ref{sec:predict-chars}, with second-order outcomes, top journals and difference-in-differences as the base groups. Heteroskedasticity-robust (HC3) standard errors are in parentheses, and *, ** and *** denote significance at the 10\%, 5\% and 1\% levels.
\end{tablenotes}
\end{threeparttable}
\end{table}

With \nReg{} claims and \nJointK{} coefficients, the joint model produces imprecise estimates. We therefore treat the
estimates with each characteristic alone as the main evidence and use the joint model to see which characteristics stand
out when all enter together (Section~\ref{sec:predict-joint}). We say that a characteristic stands out when, in the
joint model with experiment fixed effects, its coefficient on the \emph{share supported}, \emph{reproduction} or the
\emph{median signed $z$} has $p < 0.05$. Only \nOrdOneN{} claims study first-order effects,
and only \nRdN{} claims use a regression-discontinuity design. With so few claims in a group, robust standard errors
can understate the uncertainty of a coefficient, so the $p$-values of these two comparisons are less reliable than the
others.

\subsection{The Characteristics}\label{sec:predict-chars}

\paragraph{Order of the outcome.} The order of an outcome measures how far it lies down the chain of responses to the shock. An outcome is first order if
the shock affects it directly (the first-order effect), second order if it responds directly to a first-order change,
and third or higher order if it lies further down the chain. Each additional step requires the paper to establish
another link, with its own assumptions and choices of measure, sample and timing. We write a codebook that identifies
the first-order outcome of each experiment (Table~\ref{tab:tab_experiments} in Appendix~\ref{app:experiments}) and gives
worked examples from the Reg SHO pilot. For example, short selling and the threat of it are the first-order
outcome. Lower stock prices and the financing constraints they create are a second-order outcome. Corporate investment,
which responds to these financing constraints, is a third-order outcome. Three agent coders assign the \emph{order of the outcome} of each paper's main estimate. Each
coder reads the codebook, the paper and a description of its main claim, without seeing any reproduction, robustness or
audit results. The final code is the median of the three codes (Appendix~\ref{app:predict-details} reports the agreement among the coders). Our
regressions include \nOrdOneN{} first-order, \nOrdTwoN{} second-order and \nOrdThreeN{} third- or higher-order claims.
Second order is the base.

\paragraph{Journal.} We classify papers by journal (Section~\ref{sec:sample-data}) into five groups. The top journals (\nJourTopN{} claims)
are the Journal of Finance, the Journal of Financial Economics and the Review of Financial Studies in finance; the
American Economic Review, Econometrica, the Journal of Political Economy, the Quarterly Journal of Economics and the
Review of Economic Studies in economics; and the Journal of Accounting Research, the Journal of Accounting and Economics
and The Accounting Review in accounting. The UTD24 list contains the 24 journals in the UT Dallas research rankings, and
the FT50 list contains the 50 journals in the Financial Times list. The next two groups are UTD24 journals outside the
top journals (\nJourUtdN{} claims) and FT50 journals outside the UTD24 list (\nJourFtfiftyN{} claims). The remaining
groups are journals outside both lists (\nJourOthjN) and working papers (\nJourWpN). The top journals are the base.

\paragraph{Design and data.} We use indicator variables for instrumental variables (\nIvN{} claims), regression discontinuity (RDD, \nRdN) and other
methods (\nOthN), mostly time series and event studies. Difference-in-differences is the base. Another indicator
identifies claims estimated with \emph{alternative data} (Section~\ref{sec:sample-data}), and one identifies papers for
which we find the authors' materials. We measure the strength of the published result using the log of the published
absolute $t$-statistic of the main estimate.

\paragraph{Authors and earlier papers.} An indicator identifies claims with at least one author at a top-20 institution, defined as one of the 20 highest-ranked
institutions in the UT Dallas Top 100 Business School Research Rankings (Appendix~\ref{app:sample} in the Internet
Appendix). We use each author's main affiliation as listed on the paper. In our sample, \nTopInstN{} claims have such an
author. We construct the \emph{repeat-author count} and the \emph{number of earlier papers} from a list of papers on the
\nCrowdExpN{} experiments we take from RNE, including the Morningstar ratings (Appendix~\ref{app:predict-details} describes the list). For each author of a claim, we count the other papers on
these experiments that the same author has written. The \emph{repeat-author count} is the average of these counts
across the claim's authors. Its median is \nCrowdAuthmMedian, and at least one
author has another paper on these experiments for \nCrowdAuthAnyPct{} of the \nReg{} claims. The \emph{number of earlier
papers} counts papers on the same experiment published before the claim's year. Its median is \nCrowdEarlierMedian. We
divide the count by ten, so its coefficient measures the difference per ten earlier papers.

\subsection{Order of the Outcome}\label{sec:predict-order}

\begin{sidewaysfigure}
\centering
\includegraphics[width=\linewidth]{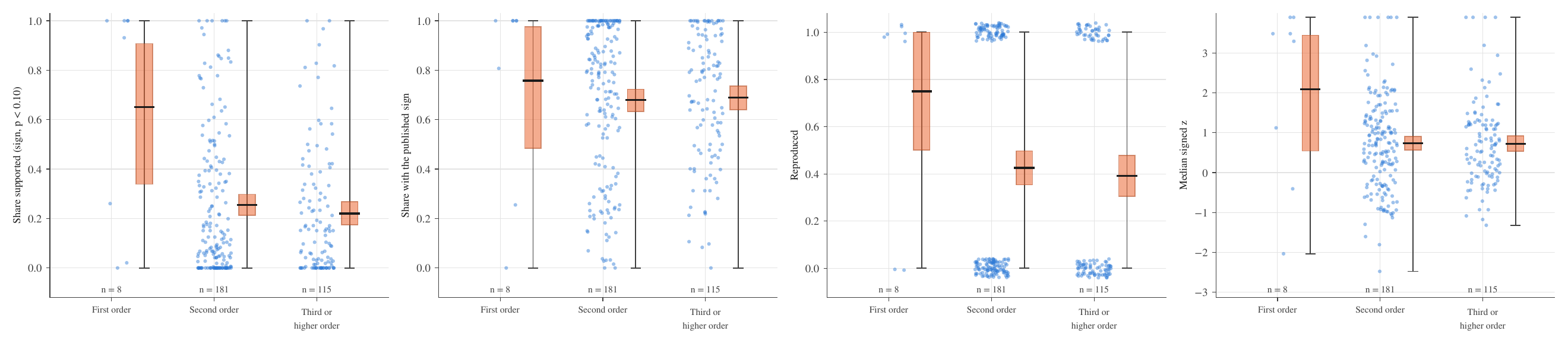}
\caption{Share supported, share with the published sign, reproduction and median signed $z$ by the order of the outcome}
\label{fig:fig_order}
\par\vspace{0.4em}
\begin{minipage}{\linewidth}\footnotesize\singlespacing
Directional claims (\nReg{} papers), shown as jittered dots. Black lines are group means, boxes their 95\% bootstrap confidence intervals, and spikes run from minimum to maximum.
\end{minipage}
\end{sidewaysfigure}

Because each step down the chain adds assumptions and analysis choices, we expect claims about first-order effects to be
more robust. Figure~\ref{fig:fig_order} presents the results. Claims about first-order effects are more robust,
consistent with our expectation. Their mean \emph{share supported} is \nOrdOneSharePct, compared with \nOrdTwoSharePct{}
for second-order claims and \nOrdThreeSharePct{} for third- or higher-order claims. We reproduce \nOrdOneRepN{} of the
\nOrdOneN{} first-order claims, and their mean \emph{median signed $z$} is \nOrdOneMedz. When we enter the \emph{order}
alone with experiment fixed effects, first-order claims have a \emph{share supported} higher by \cShareFeAloneOrdOnePp{}
percentage points than second-order claims (\cShareFeAloneOrdOneP). Their probability of \emph{reproduction} is higher
by \cRepFeAloneOrdOnePp{} percentage points (\cRepFeAloneOrdOneP), and their \emph{median signed $z$} is higher by
\cMedzFeAloneOrdOneCoef{} (\cMedzFeAloneOrdOneP). These last two differences are not statistically significant. All
three gaps rest on \nOrdOneN{} papers from \nOrdOneExp{} experiments.

Beyond the first step, second-order claims have a higher \emph{share supported} than third- or higher-order claims. When
we enter the \emph{order} alone with experiment fixed effects, third- or higher-order claims have a \emph{share
supported} lower by \cShareFeAloneOrdThreePp{} percentage points than second-order claims (\cShareFeAloneOrdThreeP).
Their coefficient on the \emph{median signed $z$} is \cMedzFeAloneOrdThreeCoef{} (\cMedzFeAloneOrdThreeP), which is not
statistically significant. We also reproduce third- or higher-order claims less often than second-order claims,
\nOrdThreeRep{} against \nOrdTwoRep{}. However, with experiment fixed effects, their
probability of \emph{reproduction} is lower by \cRepFeAloneOrdThreePp{} percentage points than that of
second-order claims, a difference that is not statistically significant (\cRepFeAloneOrdThreeP).

\subsection{The Strength of the Published Result}\label{sec:predict-logt}

\begin{sidewaysfigure}
\centering
\includegraphics[width=\linewidth]{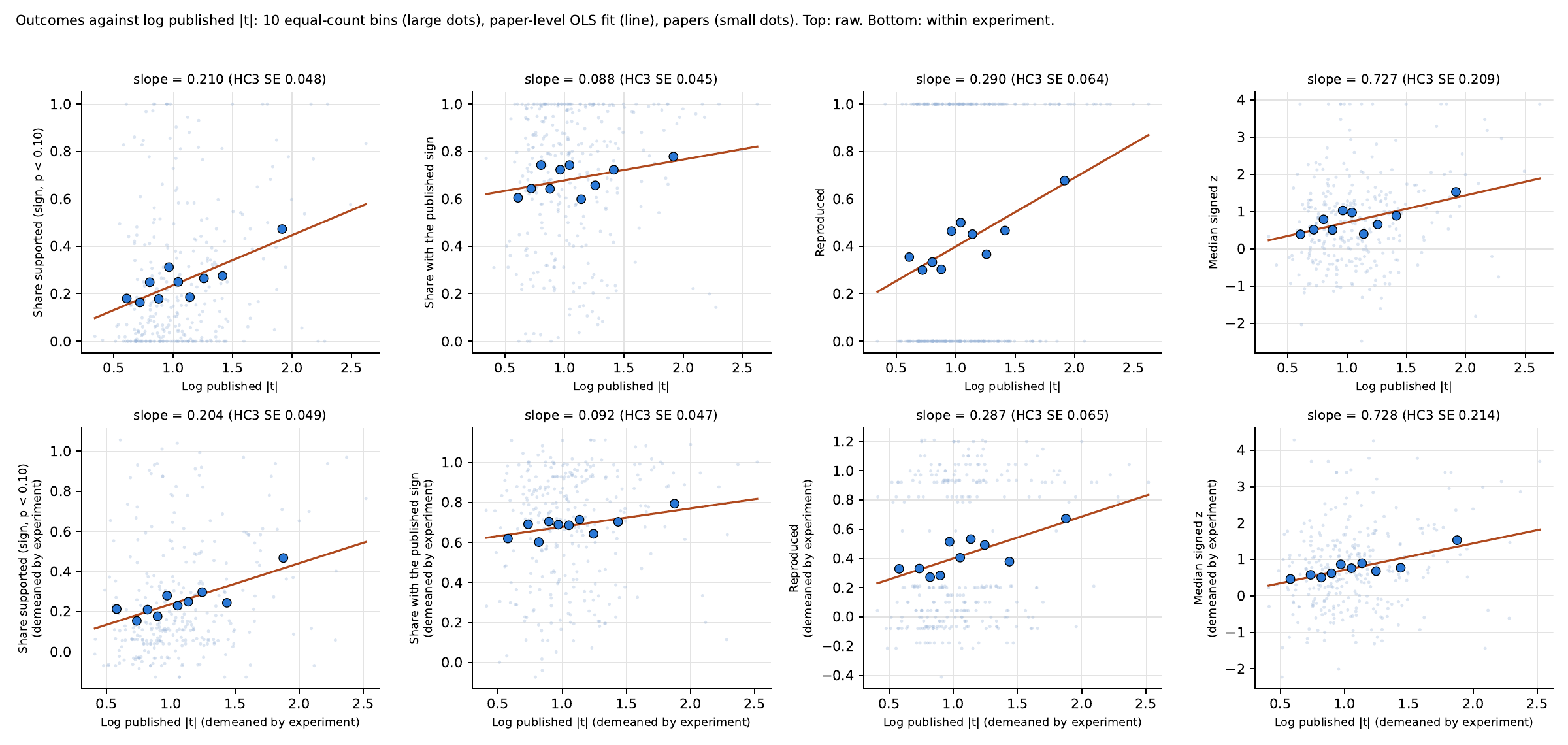}
\caption{The strength of the published result: share supported, share with the published sign, reproduction and median signed $z$}
\label{fig:fig_logt}
\par\vspace{0.4em}
\begin{minipage}{\linewidth}\footnotesize\singlespacing
Directional claims (\nReg{} papers). In the bottom row, both variables are demeaned by experiment, with the sample mean added back.
\end{minipage}
\end{sidewaysfigure}

We next examine whether claims with stronger published results are more robust. A claim that exceeds the significance
threshold by a wider margin has more room to remain significant when analysis choices change. Consistent with this
expectation, claims with stronger published results are more robust on all four outcomes. With experiment fixed
effects, a one-unit increase in the \emph{log published $|t|$} is associated with a \emph{share supported} higher by
\cShareFeAloneLogtPp{} percentage points (\cShareFeAloneLogtP), a \emph{share with the published sign} higher by
\cSignFeAloneLogtPp{} percentage points (\cSignFeAloneLogtP), a probability of \emph{reproduction} higher by
\cRepFeAloneLogtPp{} percentage points (\cRepFeAloneLogtP) and a \emph{median signed $z$} higher by
\cMedzFeAloneLogtCoef{} (\cMedzFeAloneLogtP). Without fixed effects, the \emph{share supported} is higher by
\cShareNoAloneLogtPp{} percentage points (\cShareNoAloneLogtP). Figure~\ref{fig:fig_logt} plots the outcomes against
bins of the \emph{log published $|t|$}, both before and after removing experiment means. The bin with the strongest
published results has the highest mean in every panel.

\subsection{Authors' Materials}\label{sec:predict-mat}

Of the \nReg{} claims in our regressions, we find the authors' materials for \nMatRegN{}. These materials contain data
or code that help us follow the paper's original analysis, so we expect claims with these materials to be easier to
reproduce. Consistent with this expectation, claims for which we find the authors' materials are more robust on all
four outcomes. With experiment fixed effects, their probability of \emph{reproduction} is higher by \cRepFeAloneMatPp{}
percentage points (\cRepFeAloneMatP), their \emph{median signed $z$} is higher by \cMedzFeAloneMatCoef{}
(\cMedzFeAloneMatP), their \emph{share supported} is higher by \cShareFeAloneMatPp{} percentage points
(\cShareFeAloneMatP), and their \emph{share with the published sign} is higher by \cSignFeAloneMatPp{} percentage points
(\cSignFeAloneMatP).

\subsection{Design}\label{sec:predict-design}

We enter the design indicators together, so each coefficient compares a design with difference-in-differences.

\paragraph{Instrumental variables.} Across experiments, instrumental-variable claims have a probability of \emph{reproduction} lower by \cRepNoAloneIvPp{}
percentage points than difference-in-differences claims (\cRepNoAloneIvP) and a \emph{share with the published sign} lower by \cSignNoAloneIvPp{} percentage points
(\cSignNoAloneIvP). For the \emph{share supported} and the \emph{median
signed $z$}, the instrumental-variable coefficients are not statistically significant, either across or within
experiments. Within experiments, the instrumental-variable coefficients for \emph{reproduction} and the \emph{share with the
published sign} are not statistically significant either (Figure~\ref{fig:fig_coefs}).

\paragraph{Regression discontinuity.} All \nRdN{} RDD claims use the Russell reconstitution, which has \nExpRussellN{} claims in total. The remaining Russell
claims use other designs, mainly instrumental variables. Across experiments, RDD claims are more robust than
difference-in-differences claims. Their \emph{share supported} is higher by \cShareNoAloneRdPp{} percentage points
(\cShareNoAloneRdP), their \emph{median signed $z$} is higher by \cMedzNoAloneRdCoef{} (\cMedzNoAloneRdP), and their \emph{share with the
published sign} is higher by \cSignNoAloneRdPp{} percentage points (\cSignNoAloneRdP). Because
all RDD claims use the Russell reconstitution, the regression with experiment fixed effects compares RDD claims with
difference-in-differences claims within that experiment. In this comparison, RDD claims have a \emph{share supported}
higher by \cShareFeAloneRdPp{} percentage points (\cShareFeAloneRdP), a probability of \emph{reproduction} higher by
\cRepFeAloneRdPp{} percentage points (\cRepFeAloneRdP) and a \emph{median signed $z$} higher by \cMedzFeAloneRdCoef{}
(\cMedzFeAloneRdP). These differences rest on \nRdN{} RDD claims and are not statistically significant.

\paragraph{Other methods.} Claims estimated by other methods, mostly time series and event studies, have a \emph{share supported} higher by
\cShareNoAloneOthdesPp{} percentage points than difference-in-differences claims across experiments
(\cShareNoAloneOthdesP). Within experiments, the difference is \cShareFeAloneOthdesPp{} percentage points and is not
statistically significant (\cShareFeAloneOthdesP). Appendix~\ref{app:predict-details} reports how the design claims spread across
experiments.

\subsection{Journal}\label{sec:predict-jyd}

To compare journals, we enter the journal group indicators together, without other paper characteristics, and include
experiment fixed effects (Figure~\ref{fig:fig_coefs}). Claims in top journals are more robust than claims in journals
outside both the UTD24 and FT50 lists. Relative to top journals, claims in these other journals have a coefficient of
\cMedzFeAloneOthjCoef{} on the \emph{median signed $z$} (\cMedzFeAloneOthjP), a \emph{share supported} lower by
\cShareFeAloneOthjPp{} percentage points (\cShareFeAloneOthjP), a \emph{share with the published sign} lower by
\cSignFeAloneOthjPp{} percentage points (\cSignFeAloneOthjP) and a probability of \emph{reproduction} lower by
\cRepFeAloneOthjPp{} percentage points (\cRepFeAloneOthjP). Claims in working papers have a \emph{share supported} lower
than claims in top journals by \cShareNoAloneWpPp{} percentage points across experiments (\cShareNoAloneWpP) and by a
smaller \cShareFeAloneWpPp{} percentage points within experiments, a difference that is not statistically significant
(\cShareFeAloneWpP). Claims in the other two journal groups, UTD24
journals outside the top journals and FT50 journals outside the UTD24 list, do not differ significantly from claims in
top journals on any of the four outcomes, either across or within experiments (Appendix~\ref{app:predict-details}).

\subsection{Alternative Data}\label{sec:predict-alt}

Claims that we estimate with \emph{alternative data}, i.e., other data for at least one of the paper's data sources, are
as robust as claims that we rebuild from the paper's original sources when we enter the indicator alone and compare
claims within experiments. They are
somewhat less robust once we hold the other characteristics fixed. With experiment fixed effects and the indicator
alone, claims with \emph{alternative data} have a \emph{share supported} lower by \cShareFeAloneAltPp{} percentage
points (\cShareFeAloneAltP), a probability of \emph{reproduction} lower by \cRepFeAloneAltPp{} percentage points
(\cRepFeAloneAltP) and a coefficient of \cMedzFeAloneAltCoef{} on the \emph{median signed $z$} (\cMedzFeAloneAltP). With
all characteristics in the model (Table~\ref{tab:tab_joint}), claims with \emph{alternative data} have a coefficient of
\cMedzFeJointAltCoef{} on the \emph{median signed $z$} (\cMedzFeJointAltP) and a \emph{share with the published sign}
lower by \cSignFeJointAltPp{} percentage points (\cSignFeJointAltP). Without experiment fixed effects, their probability
of \emph{reproduction} is lower by \cRepNoJointAltPp{} percentage points (\cRepNoJointAltP) and their \emph{share with
the published sign} by \cSignNoJointAltPp{} percentage points (\cSignNoJointAltP). Across experiments and with the
indicator alone, they also have a \emph{share with the published sign} lower by \cSignNoAloneAltPp{} percentage points
(\cSignNoAloneAltP) and a probability of \emph{reproduction} lower by \cRepNoAloneAltPp{} percentage points
(\cRepNoAloneAltP). Because alternative data
differ from the data the authors used, we expect some loss of robustness. These data may explain part of the low
robustness we observe.

\subsection{Authors' Institutions}\label{sec:predict-affil}

Claims with an author at a \emph{top-20 institution} are more robust when we enter the indicator alone. However, the
differences from claims without such an author shrink when we include the other characteristics. With experiment fixed
effects and the indicator alone, claims with an author at a \emph{top-20 institution} have a \emph{median signed $z$}
higher by \cMedzFeAloneTopinstCoef{} (\cMedzFeAloneTopinstP), a \emph{share with the published sign} higher by
\cSignFeAloneTopinstPp{} percentage points (\cSignFeAloneTopinstP) and a probability of \emph{reproduction} higher by
\cRepFeAloneTopinstPp{} percentage points (\cRepFeAloneTopinstP). Their \emph{share supported} is higher by
\cShareFeAloneTopinstPp{} percentage points, a difference that is not statistically significant
(\cShareFeAloneTopinstP). With all characteristics in the model (Table~\ref{tab:tab_joint}), their \emph{median signed
$z$} is higher by \cMedzFeJointTopinstCoef{} (\cMedzFeJointTopinstP) and their \emph{share supported} by
\cShareFeJointTopinstPp{} percentage points (\cShareFeJointTopinstP). Neither difference is statistically significant.
Appendix~\ref{app:predict-details} reports results with the top 10 and the top 50 institutions.

\subsection{Papers per Experiment and Repeat Authors}\label{sec:predict-crowd}

When many papers reuse an experiment, each new paper adds more tests on the same experiment. \citet{heath2023reusing} show
that the $t$-statistic threshold for a new finding rises with the number of outcomes already tested on the same
experiment. We therefore examine whether claims on experiments
that more papers reuse, and claims whose authors return to these experiments, are less robust.
Section~\ref{sec:predict-chars} defines the \emph{repeat-author count} and the \emph{number of earlier papers}. We also
count all papers on each claim's experiment. Among the experiments in our regressions, the count ranges from
\nCrowdExpMin{} papers on the EDGAR phase-in to \nCrowdExpMax{} on the Reg SHO pilot (Table~\ref{tab:tab_experiments}
describes both experiments). Because the count varies only across experiments, experiment fixed effects absorb all its
variation. We therefore estimate its coefficient without experiment fixed effects (Appendix~\ref{app:predict-details} describes the
inference). We exclude this count from
Table~\ref{tab:tab_joint} and Figure~\ref{fig:fig_coefs}.

Claims on experiments that more papers reuse are not less robust on average. Ten more papers on the experiment are
associated with a \emph{share supported} higher by \kShareNoAloneExpPp{} percentage points (\kShareNoAloneExpP), a
probability of \emph{reproduction} higher by \kRepNoAloneExpPp{} percentage points (\kRepNoAloneExpP) and a \emph{median
signed $z$} higher by \kMedzNoAloneExpCoef{} (\kMedzNoAloneExpP). None of these differences is statistically
significant, and neither is the difference in the \emph{share with the published sign}.

Within experiments, claims with fewer earlier papers are more robust. With experiment fixed effects, ten more earlier
papers are associated with a \emph{share with the published sign} lower by \cSignFeAloneEarlierPpd{} percentage points
(\cSignFeAloneEarlierP) and a coefficient of \cMedzFeAloneEarlierCoef{} on the \emph{median signed $z$}
(\cMedzFeAloneEarlierP). The probability of \emph{reproduction} is lower by \cRepFeAloneEarlierPpd{} percentage points
(\cRepFeAloneEarlierP) and the \emph{share supported} by \cShareFeAloneEarlierPpd{} percentage points
(\cShareFeAloneEarlierP). The difference in the \emph{share supported} is not statistically significant. With all characteristics in the model (Table~\ref{tab:tab_joint}), ten more earlier papers
are associated with a \emph{share with the published sign} lower by \cSignFeJointEarlierPpd{} percentage points
(\cSignFeJointEarlierP) and a coefficient of \cMedzFeJointEarlierCoef{} on the \emph{median signed $z$}
(\cMedzFeJointEarlierP). The probability of \emph{reproduction} is lower by \cRepFeJointEarlierPpd{} percentage points
(\cRepFeJointEarlierP), a difference that is not statistically significant.

Claims whose authors have written fewer papers on these experiments have a higher \emph{share supported}. With
experiment fixed effects, a one-unit increase in the \emph{repeat-author count} is associated with a \emph{share
supported} lower by \cShareFeAloneAuthmPpd{} percentage points (\cShareFeAloneAuthmP) and a coefficient of
\cMedzFeAloneAuthmCoef{} on the \emph{median signed $z$} (\cMedzFeAloneAuthmP). The \emph{share with the published sign}
is lower by \cSignFeAloneAuthmPpd{} percentage points (\cSignFeAloneAuthmP), and the probability of \emph{reproduction}
is lower by \cRepFeAloneAuthmPpd{} percentage points (\cRepFeAloneAuthmP). These last two differences are not
statistically significant. With all characteristics in the model, a one-unit increase in the \emph{repeat-author count}
is associated with a \emph{share supported} lower by \cShareFeJointAuthmPpd{} percentage points (\cShareFeJointAuthmP).
Appendix~\ref{app:predict-details} reports two alternative measures.

\subsection{All Characteristics Together}\label{sec:predict-joint}

Table~\ref{tab:tab_joint} presents the estimates with all characteristics in one model, without and with experiment fixed
effects. We discuss the estimates with experiment fixed effects.
Five characteristics stand out. First-order claims have a \emph{share supported} higher by \cShareFeJointOrdOnePp{}
percentage points than second-order claims (\cShareFeJointOrdOneP), a probability of \emph{reproduction} higher by
\cRepFeJointOrdOnePp{} percentage points (\cRepFeJointOrdOneP) and a \emph{median signed $z$} higher by
\cMedzFeJointOrdOneCoef{}, a difference that is not statistically significant (\cMedzFeJointOrdOneP). A one-unit
increase in the \emph{log published $|t|$} is associated with a \emph{share supported} higher by \cShareFeJointLogtPp{}
percentage points (\cShareFeJointLogtP) and a probability of \emph{reproduction} higher by \cRepFeJointLogtPp{}
percentage points (\cRepFeJointLogtP). Claims for which we find the \emph{authors' materials} have a \emph{share supported}
higher by \cShareFeJointMatPp{} percentage points (\cShareFeJointMatP), a probability of \emph{reproduction} higher by
\cRepFeJointMatPp{} percentage points (\cRepFeJointMatP), a \emph{median signed $z$}
higher by \cMedzFeJointMatCoef{} (\cMedzFeJointMatP) and a \emph{share with the published sign} higher by
\cSignFeJointMatPp{} percentage points (\cSignFeJointMatP). A one-unit increase in the \emph{repeat-author count} is
associated with a \emph{share supported} lower by \cShareFeJointAuthmPpd{} percentage points (\cShareFeJointAuthmP) and
a coefficient of \cMedzFeJointAuthmCoef{} on the \emph{median signed $z$} (\cMedzFeJointAuthmP). Ten more earlier
papers on the same experiment are associated with a coefficient of \cMedzFeJointEarlierCoef{} on the \emph{median
signed $z$} (\cMedzFeJointEarlierP) and a \emph{share with the published sign} lower by \cSignFeJointEarlierPpd{}
percentage points (\cSignFeJointEarlierP). For journals outside both lists, the coefficients shrink
toward zero when we include all characteristics. The coefficient on the \emph{median signed $z$} changes
from \cMedzFeAloneOthjCoef{} with the journal indicators alone to \cMedzFeJointOthjCoef{} with all characteristics
(\cMedzFeJointOthjP). In contrast, claims in UTD24 journals outside the top journals have a \emph{share with
the published sign} higher by \cSignFeJointUtdPp{} percentage points than claims in top journals (\cSignFeJointUtdP).
Most design coefficients retain their signs from the regressions with design alone, and no design
or journal coefficient is significant at the 5\% level with experiment fixed effects.

\section{Inside the Grids}\label{sec:grids}

In this section, we cover the \nDir{} directional claims and use the \emph{share supported}, i.e., the share of a paper's
specifications that keep its directional claim. Our coding of what the papers say about their riskiest choices
(Section~\ref{sec:robust-commentary}) also includes the null and unclear claims.

\subsection{Which Choices Matter}\label{sec:robust-choices}

For each choice varied in a grid, we compute the \emph{share supported} separately for the specifications that use each
option. The range of the choice is the difference between its highest and lowest shares, and the choice with the largest
range is the paper's most influential choice. In the median paper, the range of this choice is \nDecTopMedianPp{}
percentage points, with an interquartile range of \nDecTopQOnePp{} to \nDecTopQThreePp{} percentage points across
papers. We say that a choice decides the result when the claim holds in at least half of the specifications under one
option and in no more than 10\% under another option. Such a choice exists in \nDecDecidesN{} of the \nDir{} directional claims
(\nDecDecidesPct).

These ranges are larger than those produced by randomly permuting the option labels.\footnote{For each paper, we take one varied choice at a time and randomly reassign its option labels among the
specifications that share the same options in all other varied choices. Each specification keeps its estimate and
$p$-value, and each option keeps its number of specifications. We then recompute the range of every choice and record
the largest range. Repeating this \nDecNullDraws{} times gives, for each paper, the distribution of the largest range
when the result does not depend on the options of any choice. The permutations ignore sampling differences between
options estimated on different subsamples or windows.} Under these permutations, the
median range of the most influential choice is \nDecNullMedianPp{} percentage points, and a choice that decides the
result almost never occurs. The largest range exceeds the 95th percentile of the paper's own permutation distribution in
\nDecNullExceedPct{} of the \nDir{} directional claims (Appendix~\ref{app:grids-details}).

Table~\ref{tab:tab_families} groups the varied choices into families by keywords in their names. Estimator and design
choices produce the largest changes in the \emph{share supported} when they are varied, while the outcome definition is
most often a paper's most influential choice. Regression-discontinuity and instrumental-variable design choices
(bandwidths, polynomials, instruments) vary in \nFamRdivN{} papers and have the largest mean range, \nFamRdivRangePp{}
percentage points. Estimator choices have the next largest mean range, \nFamEstRangePp{} percentage points, and vary in
\nFamEstN{} papers. The outcome definition varies in \nFamOutcN{} papers and is the most influential choice in
\nFamOutcTop{} of the \nDecNonzeroN{} papers with a nonzero range.

\begin{table}[!htbp]
\centering
\begin{threeparttable}
\caption{Which choices move the share supported, by choice family}
\label{tab:tab_families}
\small
\setlength{\tabcolsep}{4pt}
\begin{tabular*}{\textwidth}{@{\extracolsep{\fill}}lrrrrrrr}
\toprule
 &  & \multicolumn{3}{c}{Mean range} &  &  &  \\
\cmidrule(lr){3-5}
 &  &  & \multicolumn{2}{c}{By data} &  &  &  \\
\cmidrule(lr){4-5}
Choice family & Papers & All & Exact & Alternative & \makecell{Mean range\\under\\permutation} & \makecell{Median\\range} & \makecell{Most\\influential} \\
\midrule
RD and IV design & 46 & 0.21 & 0.22 & 0.19 & 0.06 & 0.13 & 0.07 \\
Estimator & 104 & 0.18 & 0.17 & 0.23 & 0.04 & 0.08 & 0.12 \\
Outcome definition & 213 & 0.17 & 0.17 & 0.17 & 0.04 & 0.08 & 0.21 \\
Sample period and windows & 191 & 0.13 & 0.14 & 0.10 & 0.04 & 0.06 & 0.13 \\
Other & 58 & 0.12 & 0.14 & 0.05 & 0.03 & 0.07 & 0.05 \\
Treatment definition and timing & 163 & 0.12 & 0.11 & 0.14 & 0.03 & 0.05 & 0.08 \\
Weighting & 18 & 0.12 & 0.16 & 0.01 & 0.02 & 0.00 & 0.01 \\
Control variables & 199 & 0.11 & 0.11 & 0.10 & 0.03 & 0.04 & 0.10 \\
Fixed effects and trends & 176 & 0.11 & 0.12 & 0.07 & 0.03 & 0.06 & 0.08 \\
Sample construction & 157 & 0.10 & 0.11 & 0.10 & 0.03 & 0.05 & 0.06 \\
Functional form & 46 & 0.09 & 0.10 & 0.08 & 0.02 & 0.02 & 0.02 \\
Comparison group & 81 & 0.08 & 0.08 & 0.08 & 0.03 & 0.04 & 0.04 \\
Outliers & 100 & 0.07 & 0.08 & 0.04 & 0.02 & 0.03 & 0.03 \\
\midrule
Papers & 306 &  &  &  &  &  & 223 \\
\multicolumn{8}{@{}l}{Share of papers in which a single choice decides the result: 0.16 (50 of 306)} \\
\bottomrule
\end{tabular*}
\begin{tablenotes}[flushleft]
\footnotesize
\item[] Directional claims. A family's range in a paper is the largest range among the paper's varied choices in that family (Section~\ref{sec:robust-choices}).
\end{tablenotes}
\end{threeparttable}
\end{table}

To illustrate, in Paper A the estimator decides the result (Appendix~\ref{app:grids-details}
explains how we select this example). Paper A studies a state law adopted at different dates and estimates its effect by
OLS with fixed effects. Under staggered adoption, this estimator uses already-treated units as controls and can assign
negative weights to some group-period effects \citep{goodmanbacon2021difference,dechaisemartin2020two,baker2022much}.
The choice auditor rates this estimator \emph{invalid}. The grid includes three other estimators: a stacked
difference-in-differences by adoption cohort \citep{cengiz2019effect}, the imputation estimator of
\citet{borusyak2024revisiting} and the group-time estimator of \citet{callaway2021difference}.

\begin{figure}[!htbp]
\centering
\includegraphics[width=\textwidth]{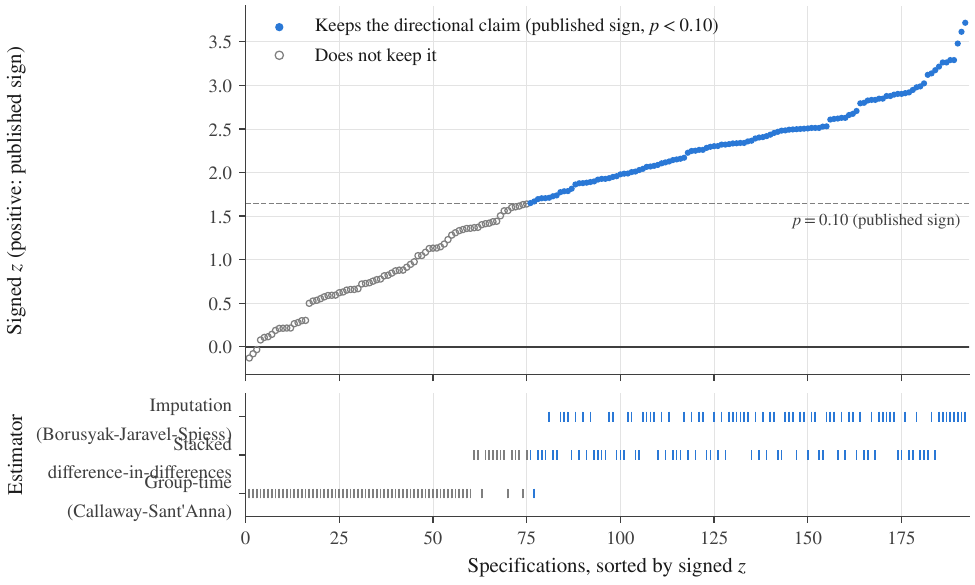}
\caption{Specification curve of one paper (Paper A)}
\label{fig:fig_example}
\par\vspace{0.4em}
\begin{minipage}{\textwidth}\footnotesize\singlespacing
The lower panel marks the estimator of each specification, colored as in the upper panel. Signed $z$ as in equation~(\ref{eq:medz}).
\end{minipage}
\end{figure}

Paper A has a \emph{share supported} of \nExASharePct{} (Figure~\ref{fig:fig_example}). The estimator has a range of
\nExAEstRangePp{} percentage points: the \emph{share supported} is \nExAImputPct{} with the imputation estimator,
\nExAStackedPct{} with the stacked estimator and \nExACsPct{} with the group-time estimator. The next most influential
choice has a range of \nExASecondPp{} percentage points.

\subsection{The Authors' Own Choices}\label{sec:robust-companion}

In \nCompRunsN{} of the \nPapers{} papers, the choice auditor excludes at least one of the paper's own options, either by
rating it \emph{invalid} or by holding its choice fixed at another option. The companion grid reruns the grid under the
same statistical inference with these options restored (Section~\ref{sec:method-audits}). Table~\ref{tab:tab_companion} covers the
\nCompAllN{} of these papers with a numeric published estimate.

\begin{table}[!htbp]
\centering
\begin{threeparttable}
\caption{Main grid and companion grid: share supported}
\label{tab:tab_companion}
\footnotesize
\begin{tabular}{lcccccc}
\toprule
 & \multicolumn{2}{c}{\makecell{Companion grid,\\all papers}} & \multicolumn{2}{c}{\makecell{Companion grid,\\directional claims}} & \multicolumn{2}{c}{Sub-grid} \\
\cmidrule(lr){2-3}\cmidrule(lr){4-5}\cmidrule(lr){6-7}
 & \makecell{All\\specs.} & \makecell{New\\specs.} & \makecell{All\\specs.} & \makecell{New\\specs.} & \makecell{All\\papers} & \makecell{Directional\\claims} \\
\midrule
Papers & 258 & 255 & 251 & 248 & 255 & 248 \\
Mean share supported, main grid & 0.22 & 0.22 & 0.23 & 0.22 & 0.22 & 0.23 \\
\makecell[tl]{Mean share supported,\\\quad companion grid} & 0.26 & 0.26 & 0.26 & 0.27 & 0.28 & 0.28 \\
Mean difference & 0.04 & 0.04 & 0.04 & 0.04 & 0.06 & 0.06 \\
Spearman correlation & 0.88 & 0.85 & 0.87 & 0.84 & 0.75 & 0.75 \\
Papers higher in the companion grid & 118 & 116 & 115 & 114 & 114 & 112 \\
Papers lower in the companion grid & 78 & 80 & 77 & 78 & 76 & 74 \\
Papers equal & 62 & 59 & 59 & 56 & 65 & 62 \\
Sign test $p$ (two-sided) & 0.005 & 0.012 & 0.007 & 0.011 & 0.007 & 0.007 \\
\bottomrule
\end{tabular}
\begin{tablenotes}[flushleft]
\footnotesize
\item[] The companion grid reruns a paper's main grid under the same statistical inference with the paper's own options restored wherever the choice auditor excluded them, either by rating them \emph{invalid} or by holding their choice fixed at another option. The sample consists of the papers with at least one excluded own option and a numeric published estimate. Each column compares a paper's \emph{share supported} on its main grid with its \emph{share supported} on a set of companion-grid specifications: all specifications of the companion grid (All specs.), only the specifications that are not in the main grid (New specs.), or only the specifications that use the paper's own option for every restored choice (Sub-grid). The columns for all papers add the null claims to the directional claims. The mean difference is the companion share minus the main-grid share. The Spearman correlation is between the two shares across papers. The sign test uses the papers whose share differs between the two grids.
\end{tablenotes}
\end{threeparttable}
\end{table}

Among the \nCompFullN{} directional claims with a companion grid, the mean \emph{share supported} rises from
\nCompFullMainPct{} on the main grid to \nCompFullCompPct{} on the companion grid. The share rises in more papers than it
falls, \nCompFullUp{} against \nCompFullDn{} (two-sided sign test on the papers whose share changes, \nCompFullSignP).

We also examine the companion specifications that use the paper's own option for every choice restored in the companion
grid (last columns of Table~\ref{tab:tab_companion}). Among the \nCompSubN{} directional claims with such
specifications, the mean \emph{share supported} is \nCompSubCompPct{} on these specifications, compared with
\nCompSubMainPct{} on the main grid of the same papers. This gain is larger than the gain on the full companion grid.
The share rises in more papers than it falls (sign test, \nCompSubSignP).

For these papers, the choice auditor rates \nRestoredInv{} of the \nRestoredAll{} restored options \emph{invalid}
(Table~\ref{tab:tab_companion_verdict}). Among papers whose companion grid restores only options rated \emph{invalid},
the \emph{share supported} on these specifications is \nVerInvonlyDiffPp{} percentage points higher than on the main
grid, a difference that is not statistically significant (sign test, \nVerInvonlyP).

\begin{table}[!htbp]
\centering
\begin{threeparttable}
\caption{Share supported by rating of the restored options}
\label{tab:tab_companion_verdict}
\small
\setlength{\tabcolsep}{5pt}
\begin{tabular*}{\textwidth}{@{\extracolsep{\fill}}lcccccc}
\toprule
 & Papers & \makecell{Main\\grid} & Sub-grid & \makecell{Mean\\difference} & \makecell{Median\\difference} & \makecell{Sign test\\$p$} \\
\midrule
Rated \emph{invalid} & 238 & 0.23 & 0.28 & 0.06 & 0.00 & 0.009 \\
\quad Only \emph{invalid} & 150 & 0.22 & 0.27 & 0.05 & 0.00 & 0.128 \\
\quad \emph{invalid} and another rating & 88 & 0.23 & 0.31 & 0.07 & 0.02 & 0.032 \\
Only grid-design exclusions & 10 & 0.23 & 0.29 & 0.07 & 0.00 & 0.625 \\
\bottomrule
\end{tabular*}
\begin{tablenotes}[flushleft]
\footnotesize
\item[] Directional claims with a sub-grid. The choice audit excludes a paper's own option either by rating it \emph{invalid} or by grid design (Section~\ref{sec:robust-companion}).
\end{tablenotes}
\end{threeparttable}
\end{table}

\subsection{Choices Changed from the Paper's Own Specification}\label{sec:robust-paths}

We next examine how the \emph{share supported} changes as specifications differ from the paper's own. We count the
number of varied choices for which a specification uses an option different from the paper's own, denoted by $k$. We use
the companion grid where available and the main grid otherwise (Table~\ref{tab:tab_paths}).
Across the \nPathN{} directional claims whose own specification we can identify, the mean \emph{share supported} falls
from \nPathAllZeroPct{} at the paper's own specification ($k = 0$) to \nPathAllThreePct{} with three or more choices
changed. Of the companion specifications, \nPathKThreePct{} have three or more choices changed.

Among the \nPathRepN{} of these claims that we reproduce, the paper's own specification keeps the directional claim in
\nPathRepZeroPct{} of cases. This share is below 100\% mainly because the two
outcomes judge the same specification with different statistical inference. \emph{Reproduction} is defined based on the paper's
own statistical inference. The \emph{share supported} is defined based on the statistical inference that the choice
auditor selects for the grid under our own standard, i.e., our rubric, which ties the statistical inference to how the experiment assigns
treatment. The auditor may select, for example, a wild cluster bootstrap where the paper uses standard errors clustered by state. The grid's
statistical inference can give the same estimate a $p$-value above 0.10. In addition, in \nHeldOffWord{} papers a
choice held fixed in the grid uses an option other than the paper's own (Appendix~\ref{app:additional-paths}). The mean
\emph{share supported} falls to \nPathRepOnePct{} with one choice changed, \nPathRepTwoPct{} with two and
\nPathRepThreePct{} with three or more. Among the other \nPathNotRepN{} claims, the share stays between
\nPathNotrepZeroPct{} and \nPathNotrepTwoPct{} at every value of $k$. Across all \nPathN{} claims, the mean \emph{share
supported} is \nPathAllOnePct{} for specifications with one choice changed, compared with \nPathAllAllkPct{} across all
their specifications. Thus, in these grids, specifications that change one choice at a time, as conventional robustness
tables do, keep the directional claim more often than the full set of specifications.

\subsection{What the Papers Say about Their Riskiest Choices}\label{sec:robust-commentary}

For each paper, an agent picks the riskiest choices, i.e., the four to six choices that most threaten the claim given
the grid results (Section~\ref{sec:method-audits}). The agent searches for what the authors say about each choice,
reruns on our rebuilt data any testable claim the authors make about it, and codes the choice into one of five classes.
Table~\ref{tab:tab_commentary} defines these classes and reports their shares across \nComChoicesN{} choices.

\begin{table}[!htbp]
\centering
\begin{threeparttable}
\caption{What the papers say about their riskiest choices}
\label{tab:tab_commentary}
\small
\setlength{\tabcolsep}{4pt}
\begin{tabular*}{\textwidth}{@{\extracolsep{\fill}}>{\raggedright\arraybackslash}p{2.7cm}>{\raggedright\arraybackslash}p{5.2cm}rrrrr}
\toprule
 &  &  & \multicolumn{3}{c}{Share of choices} &  \\
\cmidrule(lr){4-6}
 &  &  &  & \multicolumn{2}{c}{By data} &  \\
\cmidrule(lr){5-6}
Class & Definition & Choices & All & Exact & Alternative & Papers \\
\midrule
Silent & The agent finds no statement by the authors on the choice. & 776 & 0.42 & 0.43 & 0.40 & 305 \\
Partially anticipated & The authors discuss the issue but not the option that drives the result. & 499 & 0.27 & 0.26 & 0.29 & 271 \\
Claim not supported in our rerun & The authors make a testable claim about the choice that our rerun on the rebuilt data does not bear out. & 341 & 0.18 & 0.18 & 0.19 & 215 \\
Anticipated and defended & The authors discuss the choice and give a reason or a check for their option. & 137 & 0.07 & 0.07 & 0.08 & 102 \\
Disclosed but downplayed & The authors report the choice or its effect but call it minor. & 102 & 0.05 & 0.06 & 0.04 & 92 \\
\midrule
All classes &  & 1,855 & 1.00 & 1.00 & 1.00 & 320 \\
\bottomrule
\end{tabular*}
\begin{tablenotes}[flushleft]
\footnotesize
\item[] Directional, null and unclear claims. The last column counts the papers with at least one choice in the class.
\end{tablenotes}
\end{threeparttable}
\end{table}

In this coding, the papers say nothing about \nComSilentPct{} of their riskiest choices (\emph{silent}) and give a
reason or a check for \nComDefendedPct{} (\emph{anticipated and defended}). For another \nComUnsupPct, the authors make
a testable claim that the agent's rerun on our rebuilt data does not support (\emph{claim not supported in our rerun}).
For \nComPartialPct, the papers discuss the issue but omit the option that drives the result (\emph{partially
anticipated}). For the remaining \nComDownplayedPct{}, the papers report the choice but describe it as minor
(\emph{disclosed but downplayed}). Some riskiest choices concern estimators for staggered adoption that appeared after
the papers from 2018 or earlier were written. These choices do not drive the silent share: without them, the silent
share is \nStaggerSilentPct{} (Appendix~\ref{app:grids-details}).

A second agent coder runs on a different language model and works from its own written definitions of four classes.
Because it does not rerun any claim, it merges \emph{anticipated and defended} and \emph{claim not supported in our
rerun} into one class (Appendix~\ref{app:grids-details}). It
codes the riskiest choices of \nComTwoPapersN{} randomly drawn RNE papers, \nComTwoChoicesN{} choices in all, from the
paper text and a short description of each choice's grid results. The second coder finds the papers silent on
\nComTwoSilentPct{} of these choices, compared with \nComTwoFirstSilentPct{} in the first coding of the same choices.
Cohen's kappa for agreement on whether the paper is silent is \nComTwoKappa{}.

\section{Reliability of the Pipeline}\label{sec:reliab}

We take several steps to make the pipeline reliable. Separate agents audit each step, and a script checks each output
before a paper proceeds to the next step. In the first audit round, the implementation auditor finds errors that require
fixes in \nValBlockRunsN{} papers. No grid passes the final check until these errors are resolved. After the final audit
round, \nOmFindN{} serious findings remain unresolved across \nOmRuns{} papers, of which \nOmEstFindN{} can change
estimates. Two agent coders classify the data source of each main estimate, with a Cohen's kappa of \nKappaData.\footnote{Without seeing any results, each coder records whether the data come from the sources named in the paper,
from the authors' own files or from other sources. We use this classification to define \emph{alternative data}.
Cohen's kappa measures how often the two coders agree, after removing the agreement expected by chance. It equals one
when the coders always agree and zero when they agree only as often as chance predicts. A third agent resolves the
disagreements, which occur for \nKappaDataAdjN{} papers.} Within
a grid, the split-half reliabilities of the \emph{share supported} and the \emph{median signed $z$} are \nRelShare{} and
\nRelMedz, respectively.\footnote{For each paper, we randomly select one varied choice, divide its options into two halves and compute each
outcome separately on the specifications in each half. We then correlate the two half-grid values across the \nDir{}
directional claims and average the correlation over 200 random splits. The Spearman--Brown correction converts this
correlation into the reliability of the full grid. A value of one means that the outcomes computed on the
two halves move perfectly together across papers.} When we restrict the grids to specifications whose options are all rated \emph{valid}, the mean
\emph{share supported} is \nValidMeanPct, compared with \nValidMeanAllPct{} across all specifications of the same
papers. The two versions correlate at \nValidCorr{} across papers.

Two limitations remain. First, we run each paper once
with a single choice auditor, so we do not know how much the choices, ratings and outcomes would differ if other agents
repeated the analysis. Repeated analyses by large language models and AI agents can produce different results
\citep{cui2026same,gao2026nonstandard}. Second, the models' training data likely include most sample papers, and we
cannot determine whether an agent recalls a paper or analyzes it independently. Appendix~\ref{app:reliab} provides a more detailed discussion of the audits
and their unresolved findings, the agreement among the coders, the stability of the outcomes and these limitations.

\section{Conclusion}\label{sec:concl}

In this paper, we use AI agents to run a multiverse analysis of \nPapers{} papers that reuse \nExperiments{} natural
experiments and combine the results in a meta-analysis. Across the \nReg{} directional claims we compare, the mean
\emph{share supported}, i.e., the share of specifications that keep the directional claim, is \nShareMeanPct{}. This
share is zero for \nShareZeroPct{} of the claims, and the median claim has a \emph{median signed $z$} of \nMedzMedian.
We reproduce \nRepAllPct{} of the main estimates of directional claims with a published test. The \emph{reproduced}
claims have a mean \emph{share supported} of \nShareRepMeanPct{}, compared with \nShareNotRepMeanPct{} for the other
claims.

Within the grids, a single analysis choice decides the result in \nDecDecidesN{} of the \nDir{} directional claims. The
definition of the outcome variable is the most influential choice more often than any other choice. Among
\emph{reproduced} claims whose own specification we can identify, the mean \emph{share supported} falls from
\nPathRepZeroPct{} at the paper's own specification to \nPathRepOnePct{} with one choice changed and \nPathRepThreePct{}
with three or more. In \nComSilentPct{} of the cases, authors are silent on the specification choices that agents
consider most threatening to their claim.

Robustness varies with the characteristics of a paper. Within experiments, claims about first-order effects have a \emph{share
supported} higher by \cShareFeAloneOrdOnePp{} percentage points than claims about second-order effects.
Claims with a larger published $t$-statistic are more robust on all four outcomes, consistent with the concern that
results only just significant in print are the most likely to lose significance. We reproduce the main estimate more
often when we find the authors' materials. Across experiments, regression-discontinuity claims are on average more robust than
difference-in-differences claims, and instrumental-variable claims are less robust: we reproduce them less often, and
fewer of their specifications have the published sign. Claims in top journals have a higher \emph{median signed
$z$} than claims in journals outside both the UTD24 and FT50 lists. Claims on experiments that more papers reuse are
not less robust on average, but within experiments, claims with fewer earlier papers on the same experiment are more
robust. Claims for which we use
\emph{alternative data} are somewhat less robust once we hold the other characteristics fixed, so alternative data may explain
part of the low robustness we observe. With all characteristics in one model, the first-order effect, the published
$|t|$, the availability of the \emph{authors' materials}, the \emph{repeat-author count} and the \emph{number of
earlier papers} stand out.

We note several caveats in interpreting these results. First, the comparisons across papers have limited power.
Among the \nReg{} claims in \nExpK{} experiments, the first-order and
RDD contrasts rest on only \nOrdOneN{} and \nRdN{} claims, respectively. Second, a low \emph{share supported} means that a claim
depends on its analysis choices. However, a low share does not establish that the claim is false. A true effect can also
lose significance under defensible options that add noise or change what the estimate measures.

No study is likely to hold in every defensible specification. Transparency about how a result fares across these
specifications, combined with realistic expectations about how often a result grounded in a sound economic hypothesis
should hold across them, may be more productive than expecting every result to survive every check.

Our pipeline makes such an analysis inexpensive. We share the pipeline publicly at \RepoURL{} so that other
researchers can apply it in their research. If authors ran the pipeline on their own
papers, the agents would start from the authors' data and code. The reproduced estimates would then be closer to the
published estimates, and the grids would measure dependence on analysis choices more cleanly.

\bibliographystyle{chicago}
\bibliography{references}

\clearpage
\appendix
\counterwithin{table}{section}
\counterwithin{figure}{section}
\renewcommand{\thetable}{\thesection\arabic{table}}
\renewcommand{\thefigure}{\thesection\arabic{figure}}
\section{The Natural Experiments}\label{app:experiments}

Table~\ref{tab:tab_experiments} describes the 14 experiments we take from RNE: what changed and when, and the direct
target of each shock, which is what the rule or event changes at once for the firms, funds or investors it binds. An
outcome that the shock changes directly is first order (Section~\ref{sec:predict-chars}). The Morningstar sustainability
ratings have no target because the pipeline finished for none of their papers.

\begin{table}[!htbp]
\centering
\begin{threeparttable}
\caption{The natural experiments and their first-order outcomes}
\label{tab:tab_experiments}
\footnotesize
\begin{tabular}{@{}>{\raggedright\arraybackslash}p{1.15in}>{\raggedright\arraybackslash}p{2.5in}>{\raggedright\arraybackslash}p{2.5in}@{}}
\toprule
Experiment & What changed and when & First-order outcome \\
\midrule
Reg SHO pilot & Securities and Exchange Commission (SEC) Regulation SHO pilot: short-sale price tests lifted for every third Russell 3000 stock from May 2, 2005 (list announced July 28, 2004) until July 6, 2007 & Short selling of pilot stocks (short sales, short interest, lending fees and shorting costs) and the threat of it \\ \addlinespace
Business combination laws & State business combination laws, staggered adoption 1985 to 1991 & Takeover threat to covered firms and takeover activity \\ \addlinespace
Mutual fund scandal & Mutual fund late-trading and market-timing scandal, announced September 3, 2003 & Investor flows and assets of the named fund families, and the forced sales that follow \\ \addlinespace
Brokerage closures and mergers & Brokerage closures and mergers that cut analyst coverage, 56 events from 1984 to 2008 & Analyst coverage of affected firms \\ \addlinespace
Universal demand laws & State universal demand laws, staggered adoption 1989 to 2005 & Shareholder derivative litigation against covered firms and the threat of it \\ \addlinespace
Russell index reconstitution & Annual reconstitution of the Russell 1000 and 2000 indexes & Index membership and weight, and holdings of index funds and other benchmarked institutions \\ \addlinespace
Decimalization & Decimal stock quotes: New York Stock Exchange (NYSE) pilot August to December 2000; NYSE and American Stock Exchange January 29, 2001; Nasdaq April 9, 2001 & The smaller tick: quoted spreads, quoted depth and quote setting \\ \addlinespace
Tick Size Pilot & SEC Tick Size Pilot, implemented October 2016, ended 2018 & The wider tick: quoted spreads and depth, the quoting and trading of liquidity providers, and the cost of trading at restricted increments or venues \\ \addlinespace
Regulation FD & Regulation Fair Disclosure, effective October 23, 2000 & Selective disclosure to analysts and favored investors, and the disclosure practices that replace it \\ \addlinespace
EDGAR phase-in & Staggered phase-in of mandatory filing on EDGAR, the SEC's electronic filing system, 1993 to 1996 & Online availability of the firm's filings and their use \\ \addlinespace
BlackRock-BGI acquisition & BlackRock's acquisition of Barclays Global Investors, announced June 11, 2009, completed December 1, 2009 & Combined ownership of firms held by both, and common ownership of rivals held by both \\ \addlinespace
Dodd-Frank Act & Dodd-Frank Act, signed July 21, 2010 & The requirement of the provision the paper uses, and covered firms' compliance with it \\ \addlinespace
EBA capital exercise & European Banking Authority capital exercise, announced October 26, 2011, 9\% core tier 1 by June 2012 & Covered banks' capital ratios \\ \addlinespace
Morningstar sustainability ratings & Launch of the Morningstar sustainability ratings, March 2016 & n/a \\
\bottomrule
\end{tabular}
\begin{tablenotes}[flushleft]
\footnotesize
\item[] First-order outcomes as defined in Section~\ref{sec:predict-chars}. n/a: not coded, because no paper on the experiment is in the sample.
\end{tablenotes}
\end{threeparttable}
\end{table}

\section{Sample Construction}\label{app:sample}

The \nCrowdExpN{} RNE tabs we study contain \nRneRowsN{} rows, one for each paper and outcome variable. Within each tab, these rows contain
\nRneTitlesN{} paper titles. We merge the rows for each title into one entry per paper and experiment. One paper appears as both a
draft and a published article, which we count once, leaving \nListedN{} entries. One other paper appears under two experiments,
so the \nListedN{} entries represent \nListedDistinctN{} distinct papers.

Table~\ref{tab:tab_attrition} follows the RNE entries through sample construction. An entry is started when the pipeline
begins processing it and completed when the pipeline finishes processing it (Section~\ref{sec:sample-attrition}). Of the
\nAttNeverN{} entries for which the pipeline never started, \nAttNeverRestrictedN{} require restricted or hand-collected data, and another \nAttNeverNotHeldN{} require
data we did not hold. Of the remaining entries, \nAttNeverNoEstN{} have no usable main estimate and \nAttNeverOtherN{} fall into other cases. All \nAttStoppedN{}
entries for which the pipeline stopped after starting lacked data: \nAttStoppedRestrictedN{} required restricted data, and \nAttStoppedNotHeldN{} required data we
did not hold. Restricted data include institutional trading records, confidential government microdata, proprietary fund
ratings and data collected by the authors. Data we did not hold are mostly WRDS tables absent from our local copy, such
as early intraday trade data, and data that the authors release on request. All papers on the Morningstar sustainability
ratings drop out, including \nAttMsrRestrictedN{} that require proprietary or hand-collected data.

\begin{table}[!htbp]
\centering
\begin{threeparttable}
\caption{From listed papers to the sample}
\label{tab:tab_attrition}
\small
\begin{tabular*}{\textwidth}{@{\extracolsep{\fill}}>{\raggedright\arraybackslash}p{1.2in}rrrrrrr}
\toprule
 &  &  &  & \multicolumn{4}{c}{Not completed, by reason} \\
\cmidrule(lr){5-8}
Experiment & Listed & Started & Completed & \makecell{Restricted or\\hand-collected\\data} & \makecell{Data we\\do not hold} & \makecell{No usable\\main\\estimate} & Other \\
\midrule
Reg SHO pilot & 62 & 48 & 45 & 6 & 2 & 9 & 0 \\
Business combination laws & 33 & 32 & 22 & 9 & 1 & 1 & 0 \\
Mutual fund scandal & 33 & 20 & 20 & 9 & 1 & 3 & 0 \\
Brokerage closures and mergers & 24 & 19 & 18 & 3 & 3 & 0 & 0 \\
Universal demand laws & 20 & 17 & 17 & 1 & 0 & 1 & 1 \\
Russell index reconstitution & 18 & 16 & 16 & 2 & 0 & 0 & 0 \\
Decimalization & 26 & 15 & 15 & 5 & 3 & 3 & 0 \\
Tick Size Pilot & 18 & 13 & 13 & 2 & 3 & 0 & 0 \\
Regulation FD & 21 & 12 & 12 & 5 & 4 & 0 & 0 \\
EDGAR phase-in & 13 & 10 & 10 & 0 & 1 & 1 & 1 \\
BlackRock-BGI acquisition & 34 & 6 & 6 & 1 & 0 & 27 & 0 \\
Dodd-Frank Act & 11 & 6 & 6 & 4 & 1 & 0 & 0 \\
EBA capital exercise & 7 & 1 & 1 & 5 & 0 & 1 & 0 \\
Morningstar sustainability ratings & 17 & 1 & 0 & 12 & 0 & 5 & 0 \\
\midrule
Total & 337 & 216 & 201 & 64 & 19 & 51 & 2 \\
\bottomrule
\end{tabular*}
\begin{tablenotes}[flushleft]
\footnotesize
\item[] Counts of RNE entries, one per paper and experiment. Columns as defined in Appendix~\ref{app:sample}.
\end{tablenotes}
\end{threeparttable}
\end{table}

Completed entries differ from the others in their distribution across journal tiers and experiments.
Table~\ref{tab:tab_selection} compares the \nRnePapers{} completed entries with the \nSelNotDoneN{} others using two-sided tests. The
distribution across journal tiers differs between the two groups (chi-square test, \nSelTierChiP), mainly because field
journals are more common among completed entries. The share of top-journal entries does not differ (Fisher exact test,
\nSelTopP). Working papers are more common among entries that did not complete (\nSelWpP), but \nSelBlankN{} such entries with
no journal listed in RNE count as working papers. Excluding these \nSelBlankN{} entries, the working-paper difference is not
significant (\nSelWpNoBlankP), while the distribution across tiers still differs (\nSelTierNoBlankP). Publication years do not
differ significantly (Mann--Whitney test, \nSelYearP). The distribution across experiments also differs (\nSelExpP,
based on \nSelPermN{} random permutations of completion), mainly because entries on the BlackRock-BGI acquisition and
Morningstar ratings rarely completed.

\begin{table}[!htbp]
\centering
\begin{threeparttable}
\caption{Completed and not completed entries}
\label{tab:tab_selection}
\small
\begin{tabular*}{\textwidth}{@{\extracolsep{\fill}}lrrr}
\toprule
 & Completed & Not completed & All entries \\
\midrule
Entries & 201 & 136 & 337 \\
\addlinespace
\multicolumn{4}{l}{\textit{Panel A. Journal (share of entries)}} \\
Top journal & 0.30 & 0.25 & 0.28 \\
Field journal & 0.25 & 0.07 & 0.18 \\
Other journal & 0.13 & 0.24 & 0.18 \\
Working paper & 0.32 & 0.44 & 0.37 \\
\addlinespace
\multicolumn{4}{l}{\textit{Panel B. Publication year}} \\
Median & 2018 & 2019 & 2019 \\
Interquartile range & 2016--2020 & 2017--2021 & 2016--2021 \\
\addlinespace
\multicolumn{4}{l}{\textit{Panel C. Experiment (share of entries)}} \\
Reg SHO pilot & 0.22 & 0.13 & 0.18 \\
Business combination laws & 0.11 & 0.08 & 0.10 \\
Mutual fund scandal & 0.10 & 0.10 & 0.10 \\
Brokerage closures and mergers & 0.09 & 0.04 & 0.07 \\
Universal demand laws & 0.08 & 0.02 & 0.06 \\
Russell index reconstitution & 0.08 & 0.01 & 0.05 \\
Decimalization & 0.07 & 0.08 & 0.08 \\
Tick Size Pilot & 0.06 & 0.04 & 0.05 \\
Regulation FD & 0.06 & 0.07 & 0.06 \\
EDGAR phase-in & 0.05 & 0.02 & 0.04 \\
BlackRock-BGI acquisition & 0.03 & 0.21 & 0.10 \\
Dodd-Frank Act & 0.03 & 0.04 & 0.03 \\
EBA capital exercise & 0.00 & 0.04 & 0.02 \\
Morningstar sustainability ratings & 0.00 & 0.13 & 0.05 \\
\bottomrule
\end{tabular*}
\begin{tablenotes}[flushleft]
\footnotesize
\item[] Journal and year as listed in RNE, with blank fields filled and journals grouped into tiers as in Appendix~\ref{app:sample}.
\end{tablenotes}
\end{threeparttable}
\end{table}

We group journals into four tiers. The top journals are the Journal of Finance, the Journal of Financial Economics, the
Review of Financial Studies, the American Economic Review, Econometrica, the Journal of Political Economy, the Quarterly
Journal of Economics, the Review of Economic Studies, the Journal of Accounting Research, the Journal of Accounting and
Economics and The Accounting Review. The twelve field journals are the Journal of Financial and Quantitative Analysis,
the Review of Finance, Management Science, the RAND Journal of Economics, the Journal of Corporate Finance, the Journal
of Banking and Finance, Financial Management, Contemporary Accounting Research, the Review of Accounting Studies, the
Strategic Management Journal, the Journal of Law and Economics and the Journal of Environmental Economics and
Management. We classify journals by matching their names as text, so one RNE paper in a journal whose name contains that
of a field journal counts as a field-journal paper. All remaining journals are other journals, and working papers form
the fourth group. Of the \nRnePapers{} RNE papers in the sample, \nTierTopN{} are in top journals, \nTierFieldN{} in field journals, \nTierOtherN{} in other
journals and \nTierWpN{} are working papers. The \nTierPubN{} published papers appeared in \nJournalsN{} journals.

We fill or correct three RNE fields. RNE leaves the journal blank for \nBlankJournalN{} sample papers, all on the mutual fund scandal.
We classify \nBlankJournalFilledN{} as journal papers and \nBlankJournalWpN{} as working papers. For three of the four journal papers, we enter the journal of
the version we analyze. For the fourth, we enter the journal of its published version but analyze an earlier working
paper. In Section~\ref{sec:predict}, we classify this fourth paper as a working paper because we follow the version we
analyze. If we left the blank journal fields as working papers, the sample would have \nTierFieldBlankWpN{} field-journal papers, \nTierOtherBlankWpN{}
other-journal papers and the same number of top-journal papers. The number of working papers would rise to \nTierWpBlankWpN. Where RNE
gives no year, we use the year of the version we analyze. RNE lists \nRneAuthorErrWord{} author lists incorrectly, and
Table~\ref{tab:tab_papers} reports the authors as they appear on the papers.

We analyze \nVerDiffN{} RNE papers in a version different from the one RNE lists. To identify these differences, we compare the
RNE journal field with the version we analyze for every sample paper. We also compare the first pages of the version we
analyze with the authors and title listed in RNE. For \nVerSameJournalN{} papers, RNE lists a journal and we analyze a journal version
(published, typeset or accepted). For another \nVerSameWpN{} papers, both RNE and the version we analyze identify a working paper.
Of the \nBlankJournalN{} papers with a blank RNE journal field, we analyze \nVerBlankJournalN{} in a journal version. The other \nVerBlankWpN{} are working papers or
theses. In \nVerWpToJournalN{} cases, RNE lists a working paper and we analyze the journal version; in \nVerJournalToEarlierN{} cases, RNE lists a journal and
we analyze an earlier version. Another \nVerTitleAuthorN{} papers have the same publication status as the RNE entry but a different
author list or working-paper title. The author lists differ in \nVerAuthorDiffN{} cases and the titles in \nVerTitleDiffN. In
Section~\ref{sec:predict}, the journal and the year used to count earlier papers follow the version we analyze, so the \nVerWpToJournalN{} papers analyzed in their
journal version count as journal papers. Among the RNE papers whose claims enter our regressions, the journal or year of
the version we analyze differs from the RNE entry for \nVersionDiffN{} claims. These include \nVersionYearOnlyN{} papers
whose analyzed version appeared in the journal RNE lists but in a different year.

For each experiment, a script collects papers in Semantic Scholar that cite or are cited by one of its RNE papers and
that either match the experiment's keywords or link to at least two of its RNE papers. The script excludes RNE papers
and their other versions, reanalyses, replications and comments. The screens for papers added from the citation network
(Section~\ref{sec:sample-added}) follow RNE's inclusion rule with a few additional rules. We write these rules into the
screening codebook before the screens that use them. A paper that measures exposure with an index of several state
takeover laws does not count as a business combination law paper. A paper on a similar rule in another country does not
use the experiment, so we also exclude papers on Chinese brokerage mergers and closures. We exclude papers that estimate
only the joint effect of a bundle of regulations, such as Regulation FD together with the Sarbanes-Oxley Act. We also
exclude companion pieces that reuse the sample and authors of an RNE paper. For decimalization, the Reg SHO pilot and
the Tick Size Pilot, an estimate of the shock's effect on market quality counts as a use of the experiment, as in RNE.
We retain retracted papers. For each paper that passes the screens, we use the journal's typeset version where
available. We record its authors, title, journal and year from that version and use the same information in
Section~\ref{sec:predict}.

The four tiers above describe selection into the sample (Table~\ref{tab:tab_selection}).
In Section~\ref{sec:predict}, we use five journal groups instead (Section~\ref{sec:predict-chars}); their top group
is the list of top journals above.

The paper counts of Section~\ref{sec:predict} come from a list of all papers on the \nCrowdExpN{} experiments, whether
or not we ran them. The list includes all \nListedN{} RNE entries: the \nRnePapers{} in the sample and the
\nListedNotDoneN{} that did not complete the pipeline (Section~\ref{sec:sample-attrition}). It also includes
the \nCrowdScreenN{} papers that remain of the \nAddCandN{} papers passing the citation-network screens
(column (2) of Table~\ref{tab:tab_attrition_added}) after we exclude \nCrowdVersionsN{} versions of RNE papers. Of these \nCrowdScreenN{} papers, \nAddRegN{} enter the pipeline,
including \nAddedPapers{} in the sample, and \nAddNoobtainN{} have no obtainable full text
(Section~\ref{sec:sample-added}). Together, the two sources give \nCrowdEntries{} entries, one per paper and experiment,
covering \nCrowdPapers{} distinct papers. A paper on two experiments has an entry for each experiment. One RNE paper is
listed under two experiments, and one screened paper passes the screens for two experiments. We date each sample paper
by the version we analyze, each other RNE entry by RNE's year, and each other screened paper by its year in Semantic
Scholar. Author names come from the Semantic Scholar records of the citation-network search or, for RNE entries without
such records, from the paper or RNE's author field. We treat two names as the same author when Semantic Scholar assigns
them the same identifier or when they share the surname and first given name. We match a name with only an initial or
surname to an author only when exactly one author fits it.

\begin{table}[!htbp]
\centering
\begin{threeparttable}
\caption{Papers added from the citation network}
\label{tab:tab_attrition_added}
\small
\setlength{\tabcolsep}{4pt}
\begin{tabular*}{\textwidth}{@{\extracolsep{\fill}}>{\raggedright\arraybackslash}p{1.45in}rrrrrrrrr}
\toprule
 & \makecell{Linked\\papers} & \makecell{Pass\\screens} & \makecell{Regis-\\tered} & \makecell{Data\\held} & \makecell{Data not\\down-\\loaded} & \makecell{Data not\\obtain-\\able} & \makecell{No\\usable\\estimate} & \makecell{Runs\\com-\\pleted} & \makecell{In\\sample} \\
Experiment & (1) & (2) & (3) & (4) & (5) & (6) & (7) & (8) & (9) \\
\midrule
Reg SHO pilot & 311 & 30 & 25 & 0 & 22 & 2 & 1 & 0 & 0 \\
Business combination laws & 981 & 20 & 20 & 9 & 7 & 2 & 2 & 9 & 9 \\
Mutual fund scandal & 9 & 0 & 0 & 0 & 0 & 0 & 0 & 0 & 0 \\
Brokerage closures and mergers & 657 & 45 & 45 & 18 & 6 & 14 & 7 & 18 & 18 \\
Universal demand laws & 102 & 23 & 20 & 8 & 11 & 1 & 0 & 8 & 8 \\
Russell index reconstitution & 608 & 43 & 42 & 22 & 15 & 5 & 0 & 22 & 22 \\
Decimalization & 460 & 22 & 21 & 12 & 4 & 5 & 0 & 12 & 12 \\
Tick Size Pilot & 73 & 20 & 19 & 7 & 9 & 2 & 1 & 7 & 7 \\
Regulation FD & 556 & 55 & 55 & 36 & 6 & 12 & 1 & 36 & 36 \\
EDGAR phase-in & 48 & 5 & 4 & 2 & 1 & 1 & 0 & 2 & 2 \\
BlackRock-BGI acquisition & 523 & 7 & 6 & 1 & 2 & 0 & 3 & 1 & 1 \\
Dodd-Frank Act & 120 & 22 & 22 & 8 & 6 & 8 & 0 & 8 & 8 \\
EBA capital exercise & 48 & 4 & 4 & 0 & 1 & 3 & 0 & 0 & 0 \\
Morningstar sustainability ratings & 367 & 0 & 0 & 0 & 0 & 0 & 0 & 0 & 0 \\
\midrule
Total & 4,863 & 296 & 283 & 123 & 90 & 55 & 15 & 123 & 123 \\
\bottomrule
\end{tabular*}
\begin{tablenotes}[flushleft]
\footnotesize
\item[] Columns follow the steps of Appendix~\ref{app:sample}. Columns (4) to (7) split column (3) by the outcome of the data check.
\end{tablenotes}
\end{threeparttable}
\end{table}

The top-institution indicator of Section~\ref{sec:predict-chars} uses each author's main affiliation, i.e., the
affiliation printed with the author's name on the version we analyze. Network affiliations such as the NBER or the CEPR
never count. Two coders read each paper's title page and, where needed, its full text. A third coder resolves their
disagreements. The RNE papers were coded this way in an earlier round. The \nAffAddedN{} added papers were coded later
under the same rules, and the two coders identify the same ranked institutions for every added paper. Two papers report
no affiliation. For one, we use the affiliations from an online record of the paper. The other has no recorded
affiliation, so we code it as having no author at a ranked institution. An affiliation matches a ranked institution when
it names the university or its business school. We use the worldwide UT Dallas Top 100 Business School Research Rankings
by journal, based on publications from 2000 to 2020 in the Journal of Finance, the Journal of Financial Economics, the
Review of Financial Studies, The Accounting Review, the Journal of Accounting Research and the Journal of Accounting and
Economics.

\subsection{Further Details on the Sample and the Main Claim}\label{app:sample-details}

\paragraph{Excluded tabs.} The \nRneTabsOutWord{} experiment tabs that list more than one paper but whose settings lie
outside finance and accounting are the Vietnam Draft Lottery, Rainfall, the Tennessee STAR Project, the Oregon Health
Insurance Experiment and Hurricane Katrina.

\paragraph{Entries that do not complete the pipeline.} Before the pipeline starts, an agent checks whether the data required for the paper's main estimates are available. Of
the \nAttNoEstN{} RNE entries with no usable main estimate, 27 are commentaries, reviews and legal essays listed under
BlackRock's acquisition of Barclays Global Investors. RNE entries that complete the pipeline appear more often than the
others in the twelve field journals. They appear less often in working papers and in journals outside both the top tier
and these twelve journals (Table~\ref{tab:tab_selection}). Completion rates range from zero for the Morningstar ratings
to 89 percent for the Russell reconstitution.

\paragraph{Candidate estimates.} For the RNE papers whose candidate list an agent builds, the agent reads only the paper and the outcomes RNE lists for
it. The preferred specification is the one quoted in the abstract or introduction. If neither quotes a specification,
the agent selects the first column with the full sample and full controls.

\section{Pipeline Details}\label{app:pipeline}

This section describes the pipeline of Section~\ref{sec:method} step by step. The repository at \RepoURL{} holds every
prompt and script. Table~\ref{tab:tab_roles} and Figure~\ref{fig:fig_pipeline} use short names for the agent roles and
short codes for the script checks. We give these names and codes in parentheses so readers can match them to the text.

\begin{table}[!htbp]
\centering
\begin{threeparttable}
\caption{Roles, models and script checks of the pipeline}
\label{tab:tab_roles}
\footnotesize
\begin{tabular}{>{\raggedright\arraybackslash}p{2.7cm}l>{\raggedright\arraybackslash}p{5.6cm}>{\raggedright\arraybackslash}p{5.4cm}}
\toprule
Role & Model & What it does & Script check and pass rule \\
\midrule
Package finder & Sonnet & Searches for the authors' data and code and saves them unchanged & GP (script): paper, data, build log and code present \\
\addlinespace[0.5pt]
Package builder & Opus & Rebuilds the analysis data from the source databases; may not estimate the effect & GP, as above \\
\addlinespace[0.5pt]
Sample reviewer & Opus & Checks every sample rule and reported numbers of observations against the rebuilt data & GS: no fixable gap, or gaps recorded after two reviews \\
\addlinespace[0.5pt]
Builder-fix agent & Opus & Fixes the data errors and sample gaps a reviewer or auditor names & Reviewed again by the role that named them \\
\addlinespace[0.5pt]
Configuration drafter & Sonnet & Records the main estimate and the table that reports it & GC: exactly one main estimate, drawn from the paper's candidate estimates; reviewer approves \\
\addlinespace[0.5pt]
Configuration reviewer & Opus & Checks the main-estimate pick, transcription and design fields & GC, as above; one redraft \\
\addlinespace[0.5pt]
Reproducer & Sonnet & Reimplements the main estimate from the paper's description & G0: within 10\% of the published value or half a unit of its last printed decimal \\
\addlinespace[0.5pt]
Reproduction auditor & Opus & After a G0 miss, names errors before estimating their effect & Clean or minor: continue, gap recorded; errors: fix, reproduce again; unresolved: stop \\
\addlinespace[0.5pt]
Choice brainstormer & Opus & Lists every analysis choice and its options; judges none & G2, below \\
\addlinespace[0.5pt]
Choice auditor & Opus & Rates every option; picks the grid; fixes the test before any estimate & G2: every option rated; 4 to 8 varied choices; 100 to 250 specifications; no \emph{invalid} option varied; test fixed \\
\addlinespace[0.5pt]
Implementer & Sonnet & Writes and runs the grid script under the grid's statistical inference & G4, below \\
\addlinespace[0.5pt]
Implementation auditor & Opus & Reviews the grid code, recomputes 3 random specifications, rates findings & G4: no critical finding in the last audit; round-1 findings fixed and re-audited; grid complete \\
\addlinespace[0.5pt]
Fix-round agent & Sonnet & Fixes each critical finding, and fixes or rebuts each serious finding, in a fresh session & Re-audited under G4; at most 4 audit rounds \\
\addlinespace[0.5pt]
Companion implementer & Opus & Reruns the grid with the paper's own choices restored & G4C, below \\
\addlinespace[0.5pt]
Companion auditor & Opus & Audits the companion grid and recomputes 2 random specifications & G4C: no critical finding in the last audit; at most 3 rounds \\
\addlinespace[0.5pt]
Companion-fix agent & Opus & Fixes the companion audit's findings & Re-audited under G4C \\
\addlinespace[0.5pt]
Author-commentary agent & Opus & Picks the riskiest choices (Table~\ref{tab:tab_commentary}), quotes the authors on them, reruns their checks & None \\
\addlinespace[0.5pt]
Reporter & Sonnet & Writes the report from a script that computes every number & G6: the script regenerates the report exactly; one retry \\
\addlinespace[0.5pt]
Executor & Opus & Runs the check scripts and other shell commands; decides nothing & None \\
\bottomrule
\end{tabular}
\begin{tablenotes}[flushleft]
\footnotesize
\item[] The models are Claude Opus 5.5 and Claude Sonnet 5.5. Script checks as defined in Appendix~\ref{app:pipeline}.
\end{tablenotes}
\end{threeparttable}
\end{table}

\subsection{Roles, Script Checks and Retry Limits}

The pipeline has 18 roles plus a separate agent (the executor) that only runs the check scripts.
Table~\ref{tab:tab_roles} lists the roles in the order they run, the model used for each role and the script check that
reads its output. Figure~\ref{fig:fig_pipeline} shows the sequence of roles and checks.

The pipeline separates duties. A fresh agent performs each step. The agent receives a written prompt and only the files
its role requires, writes its output to files, and ends. No agent checks its own work. Different agents review the
rebuilt data, the choice of main estimate and each grid. The choice auditor rates the options on the list of choices.

A control script drives the pipeline for each paper. It launches each role as a fresh agent that reads its prompt from a
file written for the paper. The control script checks that the role's output file exists and is not empty, and retries
the role once if the file is missing. The executor runs each check script and reports its exit code. The checks are GP,
GS, GC, G0, G2, G4, G4C and G6, plus the reproduction audit after a G0 miss. Figure~\ref{fig:fig_pipeline} omits GP.
Each check reads only files. GP checks that the paper, the data, the build log and the build code are present. GS and GC
read the reviewers' verdicts. G0 compares the reproduced estimate with the published value in the record of the main
estimate. G2 reads the list of choices and the audit file. G4 and G4C read the auditors' counts of findings, and G4 also
checks that the grid is complete. G6 reruns the report script. Each check records every decision in the paper's log,
along with hashes of the files it covers.

Each loop has a retry limit within a session of the control script. The sample review, the configuration draft and the
reproduction audit may each run twice. The implementation audit may run four rounds and the companion audit three. When
the report check (G6) fails, the control script runs the report step once more. Sample gaps that remain after the second
review stay in the data. When a paper exhausts its allowed drafts of the main-estimate record, audit rounds or report
retry, we restart its pipeline at that stage in a new session. Table~\ref{tab:tab_interventions} counts these restarts.

\begin{table}[!htbp]
\centering
\begin{threeparttable}
\caption{Human decisions, restarts and voided process failures}
\label{tab:tab_interventions}
\small
\begin{tabular}{>{\raggedright\arraybackslash}p{7.6cm}ccc}
\toprule
 & \makecell{All\\(324)} & \makecell{Exact data\\(247)} & \makecell{Alternative data\\(77)} \\
\midrule
\multicolumn{4}{l}{\textit{Panel A. Decisions taken outside the agents' rules}} \\
Extra reproduction-audit round granted & 5 & 2 & 3 \\
\quad then passed G0 & 3 & 1 & 2 \\
Unresolved reproduction audit decided by us & 2 & 2 & 0 \\
Main-estimate location resolved by us & 3 & 3 & 0 \\
Runs with a note by us in the paper's log & 99 & 69 & 30 \\
Our notes in the papers' logs & 199 & 145 & 54 \\
\addlinespace
\multicolumn{4}{l}{\textit{Panel B. Restarts}} \\
Stage stopped and restarted, any cause & 91 & 62 & 29 \\
\quad stopped by a usage limit, session end or crash & 59 & 41 & 18 \\
\quad paused by us to pace usage & 14 & 8 & 6 \\
\quad other cause (bug in a check script, retry limit, request by us, cluster job) & 18 & 13 & 5 \\
Restarted after reaching a retry limit & 15 & 12 & 3 \\
\addlinespace
\multicolumn{4}{l}{\textit{Panel C. Process failures caught and voided}} \\
Runs with voided artifacts kept apart & 35 & 29 & 6 \\
Grid stage voided: agents replied to our message instead of running it & 2 & 2 & 0 \\
Grid stage voided: G4 passed an incomplete grid & 2 & 2 & 0 \\
Companion grid voided: earlier definition restored held-fixed choices only & 2 & 2 & 0 \\
Finished run restarted by mistake, stopped and voided & 1 & 1 & 0 \\
\bottomrule
\end{tabular}
\begin{tablenotes}[flushleft]
\footnotesize
\item[] Counts of runs, except the count of our notes. A restarted run counts once, under the first listed cause that applies.
\end{tablenotes}
\end{threeparttable}
\end{table}

\subsection{Rebuilding the Data and Reproducing the Main Estimate}

An agent (the package finder, Section~\ref{sec:sample-data}) searches for the authors' data and code and saves the
materials it finds unchanged. Unless these materials contain both data and code, another agent (the package builder)
rebuilds the analysis data. The package builder may check its work only against sample sizes and descriptive statistics.
A second agent (the sample reviewer) then lists every sample rule and reported number of observations in the paper,
compares them with the rebuilt data, and classifies each gap according to whether it can be fixed. The sample check (GS)
sends fixable gaps to an agent that fixes the data (the builder-fix agent), followed by a second review. Gaps that
remain after two reviews stay in the data, and the paper's report records them.

An agent (the configuration drafter) selects the main estimate and writes the file that records it (the configuration;
Section~\ref{sec:sample-claim}). A second agent (the configuration reviewer) checks the selection and transcription. The
configuration check (GC) requires the reviewer's approval and exactly one main estimate drawn from the paper's list of
candidate estimates (Section~\ref{sec:sample-claim}). Later checks use this estimate as the comparison base.

An agent (the reproducer) then estimates the main specification. With rebuilt data, the reproducer implements the design
once, as the paper describes it. The reproducer may not choose between interpretations of the paper based on how closely
their estimates match the published value. The reproduction check (G0) passes when the reproduced estimate is within
10\% of the published estimate or within half a unit of its last printed decimal. When the reproduced estimate meets
neither bound, G0 sends the paper to the reproduction auditor, i.e., an agent that checks the data and code for errors.
Before estimating anything with a corrected error, the auditor records each candidate error and the evidence for it. The
auditor may not search for a specification that produces an estimate closer to the published value.

The auditor's verdict determines the next step. If the auditor finds errors beyond minor ones, an agent fixes them and a
new agent reruns the reproduction, for at most two rounds. We granted an extra reproduction-audit round for \nIntExtraAuditN{} papers
(Table~\ref{tab:tab_interventions}). If the auditor finds no error, the pipeline continues, and the report records the
remaining gap. If the auditor finds only minor errors that do not explain the miss, an agent fixes them and a new agent
reruns the reproduction once more. The pipeline then continues without another audit, and the report records the
remaining gap. In either case, when our main estimate differs from the published estimate by more than 10\%, the paper's
grid is built around our audited reproduction. When the auditor cannot resolve a case, the pipeline pauses until we make
a decision. In the \nReproUnresolvedWord{} such cases, we resolved the open question based on what the paper documents.
We then sent each paper back for fixes and a new reproduction. If its estimate still met neither bound, the paper went
through a second audit.

\subsection{Building the Grid}

The agent that lists the choices (the choice brainstormer) links each option to a variable or file in the data and code.
Before building the grid, the choice auditor reads the rubric, the list of choices, the record of the main estimate and
the reproduction record, i.e., the file containing the published and reproduced main estimates and the G0 verdict. The
auditor can run small checks on the data to settle a rating. No grid estimate exists yet.

For each choice that the grid does not vary, the auditor uses one option in every specification. For example, if the
auditor does not select the sample period as a varied choice, every specification in the grid uses the same sample
period. The grid check (G2) confirms that the grid follows the rules of Section~\ref{sec:method-forks}, contains 100 to
250 specifications, gives every option a rating and a reason, excludes options rated \emph{invalid} from the varied
choices, and specifies a test.

In \nFixedOwnPct{} of the choices it holds fixed, the auditor keeps the paper's own option. \nFixedInvalidN{} of the
\nPapers{} grids hold a choice fixed at an option rated \emph{invalid}, in most cases the paper's own.
Options rated \emph{questionable} account for \nQuestCrossedPct{} of the options in the grids, and \nQuestCellsPct{} of
the specifications use at least one of them. In Appendix~\ref{sec:reliab-coders}, we recompute the \emph{share supported}
on the specifications whose varied options are all rated \emph{valid}.

\subsection{The Statistical Inference of the Grid}

The choice auditor fixes the test in writing, including its parameters (cluster variable, replications, seed) and a
reason for the choice, before any grid estimate exists. The agent that later estimates the grid (the implementer) must
use this test. When the test fails for a specification, the implementer's grid script records an error and leaves the
specification without an estimate.

The wild cluster bootstrap multiplies the residuals within each cluster by a common random weight, such as a random
sign, and re-estimates the model many times. These re-estimates give the distribution of the test statistic under the
null hypothesis and are more reliable than conventional cluster-robust tests when the number of clusters is small.
However, the wild cluster bootstrap can also mislead when only a few clusters are treated \citep{mackinnon2017wild}.
Randomization inference repeatedly draws treatment assignments according to the experiment's original randomization rule
and compares the estimate with those obtained under these placebo assignments.

The choice auditor selects the test based on the treatment assignment mechanism (Table~\ref{tab:tab_inference} reports
the test of the main estimate by experiment). In the pilots in which the SEC assigned stocks to treatment, the auditor
selects randomization inference for \nInfShoRandN{} of the \nInfShoN{} Reg SHO papers and for every Tick Size Pilot paper. The auditor selects
the wild cluster bootstrap for every paper on business combination laws and for \nInfUdlWildN{} of the \nInfUdlN{} papers on universal demand
laws, where states adopted the laws at different times. For the Russell reconstitution, the auditor selects analytic
methods for \nInfRusAnalyticN{} of the \nInfRusN{} papers. These methods include cluster-robust standard errors and, for regression discontinuity
designs, the robust bias-corrected confidence intervals of \citet{calonico2014robust}. These intervals remove the
leading bias of the local-polynomial estimate and widen the confidence interval to account for the added variance from
the correction. In all, \nInfResampleN{} of the \nInfAllN{} tests use resampling, permutation or a simulated or placebo null. For one paper,
the test combines a wild cluster bootstrap and a randomization test that must both reject.

\begin{table}[!htbp]
\centering
\begin{threeparttable}
\caption{Test of the main estimate by experiment}
\label{tab:tab_inference}
\footnotesize
\setlength{\tabcolsep}{4pt}
\begin{tabular}{lccccccc}
\toprule
Experiment & \makecell{Wild\\cluster\\bootstrap} & \makecell{Random-\\ization\\inference} & \makecell{Analytic} & \makecell{Placebo or\\simulated\\null} & \makecell{Other\\resampling} & \makecell{Combined} & Total \\
\midrule
Reg SHO pilot & 1 & 40 & 4 & 0 & 0 & 0 & 45 \\
Business combination laws & 31 & 0 & 0 & 0 & 0 & 0 & 31 \\
Mutual fund scandal & 8 & 1 & 10 & 0 & 0 & 1 & 20 \\
Brokerage closures and mergers & 28 & 0 & 6 & 0 & 2 & 0 & 36 \\
Universal demand laws & 24 & 0 & 1 & 0 & 0 & 0 & 25 \\
Russell index reconstitution & 10 & 0 & 27 & 0 & 1 & 0 & 38 \\
Decimalization & 12 & 1 & 10 & 1 & 3 & 0 & 27 \\
Tick Size Pilot & 0 & 20 & 0 & 0 & 0 & 0 & 20 \\
Regulation FD & 33 & 0 & 8 & 4 & 3 & 0 & 48 \\
EDGAR phase-in & 11 & 0 & 1 & 0 & 0 & 0 & 12 \\
BlackRock-BGI acquisition & 4 & 0 & 2 & 0 & 1 & 0 & 7 \\
Dodd-Frank Act & 5 & 3 & 4 & 2 & 0 & 0 & 14 \\
EBA capital exercise & 1 & 0 & 0 & 0 & 0 & 0 & 1 \\
\midrule
All papers & 168 & 65 & 73 & 7 & 10 & 1 & 324 \\
\bottomrule
\end{tabular}
\begin{tablenotes}[flushleft]
\footnotesize
\item[] Methods as described in Sections~\ref{sec:method-inference} and~\ref{app:pipeline}. Other resampling is the nonparametric bootstrap.
\end{tablenotes}
\end{threeparttable}
\end{table}

\subsection{Grid Audits, Companion Grid and Report}

The implementer writes and runs the script that estimates every specification under the grid's statistical inference. The script must
also recover the reproduced estimate from the paper's own specification. The implementation auditor checks the script
against the choice auditor's file (ratings, grid and test) and the data, and independently recomputes three randomly
drawn specifications. The auditor classifies each finding as critical, serious or minor. A
critical finding makes the estimates unreliable. A serious finding affects some specifications or their interpretation.
If the first audit finds a critical or serious finding, an agent (the fix-round agent) fixes every critical finding and
either fixes each serious finding or rebuts it in writing with evidence. The auditor then checks the implementation
again. The implementation check (G4) passes when the final audit reports no critical finding and the grid is complete,
within at most four rounds. From the second round on, the control script sends a serious finding to a fix round only if
the same audit also finds a critical one. Serious findings raised after the first round can therefore remain unresolved
when G4 passes. Appendix~\ref{sec:reliab-audits} reports the audit findings, including those that remain unresolved
(Table~\ref{tab:tab_validation}).

An agent (the companion implementer) estimates the companion grid, i.e., the grid with the paper's own option restored
for every choice where the choice auditor excluded it, whether the grid varies that choice or holds it fixed. Another
agent (the companion auditor) checks the companion grid in at most three rounds under the companion check (G4C).

An agent (the author-commentary agent) selects the riskiest choices, i.e., the four to six choices that most threaten
the claim given the grid results. Another agent (the reporter) writes the paper's report using a script that computes
every number from the grid files. The report check (G6) reruns the script and passes only if the script reproduces the
report byte for byte.

\subsection{What the Agents May Read}

Every prompt ends with the same access rules. An agent may not read other papers' folders or discarded outputs. It also
may not read, search for or rely on any third-party critique, comment, reply, replication or reanalysis of its paper.
For each paper, the pipeline identifies between \nForbidMinN{} and \nForbidMaxN{} prohibited sources, including the RNE paper itself, our study
files, and published reanalyses and critiques of papers in the same experiment. Reanalysis materials in the paper's folder,
such as the pre-analysis plan of the Reg SHO reanalysis project \citep{black2019preanalysis}, are
closed to every agent that builds, audits or reports the multiverse. If such materials also code the dates of a law,
agents that build and review the data may read them only for dates and coding. The agent that lists the choices may not
read the choice auditor's ratings. Agents that rebuild and review the data may not estimate the treatment effect or run
any regression of an outcome on the treatment. Licensed replication packages that a journal restricts to verification
are available only to the reproducer and the reproduction auditor. The estimate that G0 checks always comes from the
rebuilt data.

\subsection{Models and Compute}\label{sec:method-compute}

The package finder, the configuration drafter, the reproducer, the implementer, the fix-round agent and the reporter run
on Claude Sonnet 5.5 (Sonnet). Every other role, including the executor agent that runs the check scripts, runs on
Claude Opus 5.5 (Opus). Table~\ref{tab:tab_compute} reports the models, agent counts and computing resources per paper.
Among papers whose entire pipeline ran in one session of the control script on our server, the median paper uses about
\nAgentsMedian{} agents. In a sample of these runs, the median time from the first to the last agent is \nHoursMedian{}
hours, and the median token use is \nTokensMedian{} million, including cache reads. The rest of this subsection reports
the exact figures and explains how we measured them.

\begin{table}[!htbp]
\centering
\begin{threeparttable}
\caption{Models, agents and compute per paper}
\label{tab:tab_compute}
\small
\begin{tabular}{>{\raggedright\arraybackslash}p{6.3cm}rr>{\raggedright\arraybackslash}p{3.6cm}}
\toprule
 & Median & Range & Measured on \\
\midrule
\multicolumn{4}{l}{\textit{Panel A. Models observed}} \\
Package finder, configuration drafter, reproducer, implementer, fix round, reporter & \multicolumn{2}{l}{Claude Sonnet 5.5} & 10 sampled server runs \\
All other roles and the executor & \multicolumn{2}{l}{Claude Opus 5.5} & 10 sampled server runs \\
\addlinespace
\multicolumn{4}{l}{\textit{Panel B. Agents per paper}} \\
All agents & 87.5 & 59--121 & 112 server runs \\
Role agents & 20 & 14--34 & 112 server runs \\
Executor agents & 66.5 & 45--89 & 112 server runs \\
\addlinespace
\multicolumn{4}{l}{\textit{Panel C. Time and tokens per paper}} \\
Wall-clock hours, first to last agent & 22.4 & 8.8--30.2 & 10 sampled server runs \\
Role-agent hours, summed & 7.05 & 2.26--11.41 & 10 sampled server runs \\
Output tokens (millions) & 1.29 & 0.73--1.94 & 10 sampled server runs \\
All tokens, cache reads included (millions) & 126 & 41--178 & 10 sampled server runs \\
Share of output tokens from Opus & 0.92 &  & 10 sampled server runs \\
\bottomrule
\end{tabular}
\begin{tablenotes}[flushleft]
\footnotesize
\item[] Server runs and sampled runs as defined in Appendix~\ref{app:pipeline}. The Opus share is pooled over the sampled runs.
\end{tablenotes}
\end{threeparttable}
\end{table}

We confirmed the model identifiers in the transcripts of 10 sampled runs, in which Opus produced 92\% of the output
tokens. Agent counts come from the 112 RNE runs whose entire pipeline ran in one session on our server (the server
runs). In these runs, a paper uses a median of 87.5 agents. The separate medians are 20 role agents and 66.5 executors.
We drew the 10 sampled runs from these server runs, one for each experiment except the Reg SHO pilot, business
combination laws and the Tick Size Pilot. Five of the sampled runs use exact data. The sampled runs take a median of
22.4 hours from the first to the last agent, including waits between agents. Their role agents work a median of 7.05
hours. A run produces a median of 1.29 million output tokens and uses a median of 126 million tokens in total, including
cache reads. Parallel single-threaded workers estimate the grids. For the server runs, these workers operate on a
computing cluster.

\subsection{Human Decisions, Restarts and Discarded Outputs}

We ran the control script and made a few decisions outside the agents' rules. Table~\ref{tab:tab_interventions} lists
these decisions by data source. We granted an extra reproduction-audit round in \nIntExtraAuditN{} runs, of which
\nIntExtraPassN{} then passed G0. In \nIntUnresolvedN{} runs, the reproduction auditor could not resolve a miss. We
decided each case based on what the paper documents and sent the run back for fixes. In \nIntAnchorN{} runs, we resolved
the location of the main estimate in the paper. The logs of \nIntNoteRunsN{} runs contain our notes on these decisions
and routine events such as restarts.

Most restarts followed an interruption. In \nIntStopN{} runs, a stage stopped and we restarted it. In \nIntStopCauseN{}
of them, a usage limit, the end of a session or a crash interrupted the stage. In \nIntRetryN{} runs, we restarted a
stage after it reached a retry limit.

We identified process failures and discarded the affected outputs, which Table~\ref{tab:tab_interventions} calls voided.
In \nIntVoidRunsN{} runs, we moved the discarded outputs to folders that no agent may read. We discarded
\nIntVoidStatusWord{} grid stages because their agents replied to a message from us instead of running the stage, and
\nIntVoidIncompleteWord{} because G4 had passed an incomplete grid. We also discarded \nIntVoidCompanionWord{} companion grids
because they followed an earlier definition that restored the paper's own option only for choices held fixed.
In these papers, the choice auditor had also excluded the paper's own option from a varied choice. We restarted \nIntRerunWord{}
finished run by mistake, then stopped it and discarded its outputs.

\section{Reliability of the Pipeline: Details}\label{app:reliab}

\subsection{Audits and Unresolved Findings}\label{sec:reliab-audits}

In the first round, the implementation auditor (Section~\ref{sec:method-audits}) finds errors that make the estimates
unreliable and must be fixed for \nValBlockRunsN{} papers. The auditor also finds serious issues that affect some
specifications or their interpretation for \nValMajorRunsN{} papers (Table~\ref{tab:tab_validation}). The papers go
through \nValRounds{} audit rounds in total, \nValRoundsPerRun{} per paper on average. No grid passes the final script
check until all errors that make the estimates unreliable are resolved.

After the final audit round, \nOmFindN{} serious findings remain unresolved across \nOmRuns{} papers. Of these findings,
\nOmEstFindN{} in \nOmEstFindRunsN{} papers can change estimates, \nOmInfFindN{} concern only inference, and
\nOmDocFindN{} concern diagnostics or documentation (Appendix~\ref{app:additional-openmajor}).

The other reviewers also send papers back for fixes (Appendix~\ref{app:reliab-details}). The
sample reviewer finds fixable gaps between the rebuilt sample and the paper's documented sample rules in
\nValSampleFixN{} papers. Some gaps remain unresolved at the end for \nValSampleOpenN{} of these papers. The agent that
checks the choice of the main estimate sends \nConfigBackN{} papers back at least once. The reproduction auditor
examines the \nValReproAuditN{} papers whose reproduced estimate is neither within 10\% of the published estimate nor
within half a unit of its last printed decimal (Appendix~\ref{app:pipeline}). The auditor finds errors beyond minor ones in \nReproBackN{} of these papers and sends them back for fixes, after
which \nReproBackErrFixedN{} of them pass the reproduction check.

We analyze only files that pass the script checks and match the hashes recorded at each paper's final check. For every
paper, a script reran the report code and obtained an identical report, \nReportFailWord{} after an initial failure.
Appendix~\ref{app:pipeline} lists the decisions we made outside the agents' rules
(Table~\ref{tab:tab_interventions}).

\subsection{Coders and Outcomes}\label{sec:reliab-coders}

Two agent coders classify the data source of each main estimate without seeing any results. They distinguish among the
sources named in the paper, the authors' data and other data (Section~\ref{sec:sample-data}). Their Cohen's kappa is
\nKappaData, and a third agent resolves disagreements for \nKappaDataAdjN{} papers. The two coders of the claim type
(Section~\ref{sec:sample-claim}) agree on every paper. Appendices~\ref{app:predict-details} and~\ref{app:grids-details}
report the agreement among the coders of the \emph{order of the outcome} and of the classes describing what papers say
about their riskiest choices.

The split-half reliabilities of the \emph{share supported} and the \emph{median signed $z$} within a grid are
\nRelShare{} and \nRelMedz{}, respectively, across the \nDir{} directional claims. Table~\ref{tab:tab_validation} reports the same
reliabilities for the \nRelAllN{} papers with a numeric published estimate (\nRelAllShare{} and \nRelAllMedz). To compute them, we randomly select
one varied choice for each paper, divide its options into two halves and compute both outcomes separately for the
specifications in each half (Appendix~\ref{app:reliab-details}).

When we restrict the grids to specifications whose options in the varied choices are all rated \emph{valid}, the mean
\emph{share supported} is \nValidMeanPct, compared with \nValidMeanAllPct{} on all specifications of the same papers.
Across papers, the two versions correlate at \nValidCorr.

\begin{figure}[!htbp]
\centering
\includegraphics[width=\textwidth]{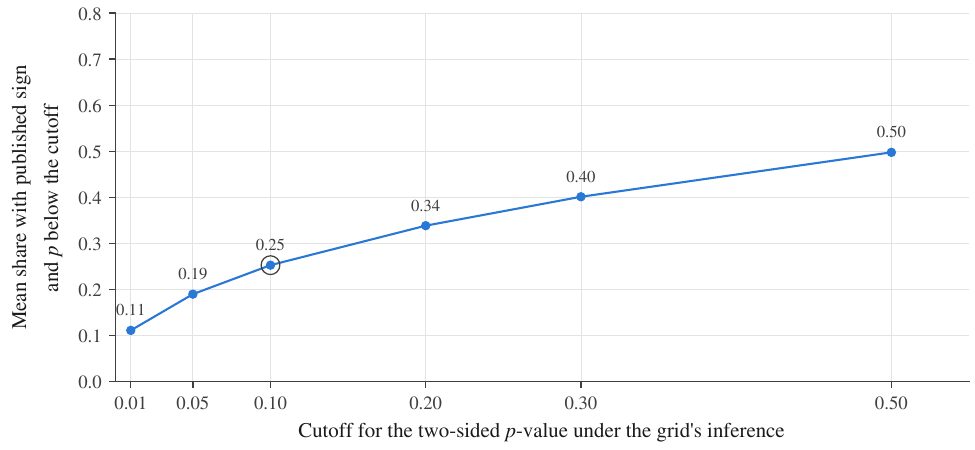}
\caption{Mean share supported under other $p$-value cutoffs}
\label{fig:fig_thresholds}
\par\vspace{0.4em}
\begin{minipage}{\textwidth}\footnotesize\singlespacing
Directional claims (\nDir{} papers). The circled point marks the cutoff of 0.10 that defines the share supported.
\end{minipage}
\end{figure}

The level of the \emph{share supported} depends on the $p$-value cutoff (Figure~\ref{fig:fig_thresholds}). Across the \nDir{} directional claims, the mean share of specifications with the published sign and
$p < 0.01$ is \nThrOnePct. It is \nThrTenPct{} at our cutoff of 0.10 and \nThrFiftyPct{} at 0.50.

\subsection{Limits}\label{sec:reliab-limits}

We run each paper once, and a single choice auditor rates its options. We therefore do not know how much the list of
choices, the ratings and the outcomes would change if different agents repeated the analysis. The split-half reliability
measures stability within a grid only. Repeated analyses of the same task by a large language model can produce
different results \citep{cui2026same}. Research teams given the same data and hypotheses produce nonstandard errors, i.e., dispersion in results
across analysts \citep{menkveld2024nonstandard}. AI agents testing the same market-quality hypotheses on the same New
York Stock Exchange Trade and Quote data also produce them \citep{gao2026nonstandard}.

The models' training data likely include most sample papers and published critiques of some of them, so the models
may already know these papers. We have no test to distinguish whether an agent recalls a paper or independently analyzes it.

\subsection{Additional Details on the Reliability Checks}\label{app:reliab-details}

\paragraph{Audit counts.} Table~\ref{tab:tab_validation} reports the audit counts for all \nPapers{} papers by data source.

\paragraph{Other reviewers.} When gaps between the rebuilt sample and the paper's documented sample rules
remain after an agent fixes the data, the paper's report records them. Before reproduction, the agent that reviews the
choice of main estimate identifies errors such as an incorrect main estimate or misstated controls, sample, fixed
effects or clustering. The reproduction auditor finds at least a minor error, or cannot resolve the miss, in
\nValReproErrN{} of the \nValReproAuditN{} papers it examines. Errors beyond minor ones lead to fixes and a new reproduction, and we granted an extra reproduction-audit round
for \nIntExtraAuditN{} papers. Minor errors are fixed, and a new
agent reruns the reproduction once more without another audit (Appendix~\ref{app:pipeline}).

\paragraph{Files.} The hashes recorded at each paper's final check cover the record of the main estimate, the file containing
the published and reproduced estimates, the list of choices and its audit, the grid script, the grid results, the audit
reports and the report. Appendix~\ref{app:pipeline} lists our decisions outside the agents' rules, the restarts and the process
failures we identified and whose outputs we discarded.

\paragraph{Coders.} The data-source coders and the third agent that settles their differences run on Sonnet for the RNE
papers and on Opus for the added papers.

\paragraph{Split-half reliability.} Averaged over 200 random splits, the correlation between the two halves across the \nDir{} directional claims is
\nRelShareR{} for the \emph{share supported} and \nRelMedzR{} for the \emph{median signed $z$}. The Spearman--Brown
correction $2r/(1+r)$ converts the half-grid correlation $r$ to a full-grid reliability, giving \nRelShare{} and
\nRelMedz{}, respectively.

\paragraph{Valid-only specifications.} Specifications whose options in the varied choices are all rated \emph{valid} account for \nValidCellsPct{} of the
\nCellsAll{} specifications across the \nCellsAllPapers{} papers. Of the \nDir{} directional claims, \nValidPapers{}
have at least one such specification. The \emph{share supported} is identical when calculated using only these
specifications or all specifications in \nValidSame{} papers.

\section{The Measures}\label{app:measure}

The specifications of a paper are the rows of its grid for the main estimate that have a numeric estimate and $p$-value.
We exclude specifications in which estimation fails or that use an option the choice auditor marks as changing what the
estimate measures. Estimation fails in \nCellsFail{} specifications across \nCellsFailPapers{} papers. After these
exclusions, \nCellsAll{} specifications remain across the \nCellsAllPapers{} papers with a numeric published estimate.
The remaining paper reports no numeric main estimate and has no outcome. A specification keeps the directional claim
when its estimate has the published sign and its two-sided $p$-value under the grid's statistical inference is below 0.10. A paper's
\emph{share supported} is the share of its specifications that keep the directional claim, as in
equation~(\ref{eq:share}). The \emph{median signed $z$} of equation~(\ref{eq:medz}) uses the same specifications and
$p$-values.

In equation~(\ref{eq:share}), $b_{ic}$ is the estimate in specification $c$ of paper $i$'s grid, $p_{ic}$ is its
two-sided $p$-value under the grid's statistical inference, $\hat{\beta}_i$ is the published main estimate, $\mathcal{C}_i$ is the set of
specifications that enter the share, and $N_i$ is their number. The indicator $\mathbf{1}\{\cdot\}$ equals one when its
condition holds and zero otherwise, and $\operatorname{sign}(x)$ equals 1, $-1$ or 0 when $x$ is positive, negative or
zero, respectively. The \emph{share with the published sign} drops the factor $\mathbf{1}\{p_{ic} < 0.10\}$, so it is
never below the \emph{share supported}. The difference between the two measures is the share of specifications with the
published sign and a $p$-value of 0.10 or more.

In equation~(\ref{eq:medz}), with conventional standard errors in a large sample, $z_{ic}$ equals the $t$-statistic up
to its sign. The floor on the $p$-value ensures that $|z_{ic}| \leq 3.89$ and gives a finite $z$ when a reported
$p$-value is zero.

Each grid contains a subset of the defensible specifications. We compute the outcomes from every specification in the
grid except the excluded ones, so no error comes from sampling specifications. The uncertainty we report for averages
reflects variation across papers and the experiments they use. We give each specification equal weight. Before the
exclusions, every option of a varied choice appears in the same number of specifications and therefore receives equal
weight.

Applying the sign and $p$-value rules as written would misclassify a few grids. The grid's hypothesis test can be one-sided or
require a second test, and one claim compares the estimate with a nonzero benchmark. We flag these grids using the files
from each paper's pipeline. Two agents independently examine the flagged papers. They read the files and the paper but
no analysis output. A third agent resolves their disagreements. We make three fixes based on the agents' verdicts.

\begin{enumerate}
\item In the \nFixOneSidedN{} grids whose grid $p$-value is one-sided in the published direction, we make the $p$-value
two-sided. It becomes twice the $p$-value when the estimate has the published sign and twice its complement
otherwise, capped at one.
\item In \nFixJointN{} grids the paper's claim requires two tests to hold jointly, and we take the larger of the two $p$-values as a
specification's $p$-value.
\item One paper's claim compares the estimate with a nonzero benchmark, so we measure the sign of each estimate
relative to that benchmark.
\end{enumerate}

When a grid also contains rows for a second estimate from the paper's main results, the script checks do not cover those
rows. We therefore score only the specifications of the main estimate that the checks cover.

\section{Additional Results}\label{app:additional}
\subsection{Choices Changed: Table Details}\label{app:additional-paths}

In Table~\ref{tab:tab_paths}, $k$ is the number of varied choices for which a specification uses an option different
from the paper's own. Choices held fixed do not count. The table covers the \nPathN{} directional claims whose own
specification we can identify. The rows for \emph{reproduced} and not \emph{reproduced} claims exclude the
\nPathRepMissWord{} claim for which \emph{reproduction} is not defined. Panel A uses the companion grid where available
and the main grid otherwise, so every paper has specifications at every $k$. A paper's own specification is its one
specification with $k = 0$, except in \nHeldOffWord{} papers where a choice held fixed in the grid uses a different
option. \emph{Reproduction} evaluates the paper's own specification using the paper's own statistical inference and requires $p \le 0.10$. We
therefore classify claims as \emph{reproduced} or not \emph{reproduced} based on their specification with $k = 0$. The
\emph{share supported} evaluates the same specification using the grid's statistical inference and requires $p < 0.10$. Thus, a
\emph{reproduced} claim need not keep the directional claim at $k = 0$. Because of this selection, comparing the two
groups is more informative from $k = 1$ onward. Panel B reports the mean \emph{share supported} by $k$ in the main grid.
In Panel B, \nPathAudNoKZeroN{} papers have no specification at $k = 0$ and \nPathAudNoKOneN{} none at $k = 1$, and a specification
with $k = 0$ can differ from the paper's own specification in a choice held fixed.

\begin{table}[!htbp]
\centering
\begin{threeparttable}
\caption{Share supported by choices changed from the paper's own specification}
\label{tab:tab_paths}
\small
\setlength{\tabcolsep}{5pt}
\begin{tabular*}{\textwidth}{@{\extracolsep{\fill}}lcccccc}
\toprule
 &  & \multicolumn{4}{c}{Choices changed, $k$} &  \\
\cmidrule(lr){3-6}
 & Papers & 0 & 1 & 2 & 3 or more & \makecell{All\\specifications} \\
\midrule
\multicolumn{7}{l}{\textit{Panel A. Companion grid (main grid if none)}} \\
All directional claims & 267 & 0.43 & 0.36 & 0.34 & 0.28 & 0.30 \\
Reproduced & 114 & 0.81 & 0.64 & 0.57 & 0.47 & 0.50 \\
Not reproduced & 152 & 0.14 & 0.15 & 0.15 & 0.14 & 0.14 \\
\midrule
\multicolumn{7}{l}{\textit{Panel B. Main grid; brackets: papers with specifications at that $k$}} \\
All directional claims & 267 & 0.40 & 0.33 & 0.30 & 0.26 & 0.27 \\
 &  & [126] & [214] & [256] & [267] &  \\
\addlinespace
Reproduced & 114 & 0.70 & 0.58 & 0.50 & 0.42 & 0.43 \\
 &  & [57] & [90] & [110] & [114] &  \\
\addlinespace
Not reproduced & 152 & 0.15 & 0.15 & 0.15 & 0.14 & 0.14 \\
 &  & [68] & [123] & [145] & [152] &  \\
\bottomrule
\end{tabular*}
\begin{tablenotes}[flushleft]
\footnotesize
\item[] Mean share supported across the directional claims whose own specification we can identify (Appendix~\ref{app:additional-paths}).
\end{tablenotes}
\end{threeparttable}
\end{table}

\subsection{Classification of Unresolved Audit Findings}\label{app:additional-openmajor}

Two coders, each running on a different language model, classify the \nOmFindN{} serious findings that remain unresolved
after each paper's final implementation audit. They classify the findings into three groups: those that can change
estimates, those that concern inference only, and those about diagnostics and documentation. The coders agree on all but
\nOmDisagreeN{} findings, with a Cohen's kappa of \nOmKappa. A third coder resolves the disagreements.

\subsection{Inside the Grids: Additional Details}\label{app:grids-details}

\paragraph{Choices with no range.} In \nDecZeroN{} of the \nDir{} directional claims, every choice has a range of zero because the \emph{share supported}
is either 0\% under every option or 100\% under every option.

\paragraph{Permutations.} In each of \nDecNullDraws{} draws, we permute the option labels of one choice at a time among specifications that share
the same options for all other varied choices. The permutation preserves the number of specifications for each option
and its combinations with the options of other choices. The resulting ranges therefore show what arises when the result
is independent of the options of a choice. We permute only the option labels and keep each specification's estimate and
$p$-value unchanged. Every family in Table~\ref{tab:tab_families} has a mean range above its mean range under the
permutations. After accounting for the permutations, the regression-discontinuity and instrumental-variable design
family and the estimator family still have the largest mean ranges, \nFamRdivNetPp{} and \nFamEstNetPp{} percentage
points, respectively.

\paragraph{Families by data source.} The differences between papers with \emph{alternative data} and exact data vary by choice family
(Table~\ref{tab:tab_families}). Estimator choices change the \emph{share supported} more in papers with
\emph{alternative data}, whereas regression-discontinuity and instrumental-variable design choices change it more in
papers with exact data.

\paragraph{Choice of Paper A.} We first apply a fixed rule to select \emph{reproduced} directional claims with exact data, at least 150 specifications
and a most influential choice in either the estimator family or the regression-discontinuity and instrumental-variable
design family (Table~\ref{tab:tab_families}). Among these claims, we pick Paper A by hand.

\paragraph{Companion grids.} The median companion grid restores the paper's own option in \nCompRestoredMedian{} choices. In a full companion grid,
the paper's own option receives only part of the weight because it is one of several options for each restored varied
choice. We therefore also score the companion specifications that use the paper's own option in every restored choice.
On these specifications, the median \emph{share supported} rises from \nCompSubMedianMainPct{} on the main grid to
\nCompSubMedianSubPct. Papers whose companion grid restores both an option rated \emph{invalid} and an option rated
\emph{valid} or \emph{questionable} but held fixed at another option gain \nVerMixedDiffPp{} percentage points
(\nVerMixedP).

\paragraph{Riskiest choices.} The prompt for the agent that codes the riskiest choices names the five classes without defining them. We wrote the definitions in Table~\ref{tab:tab_commentary} for the table, after the coding. Some options
became available after the papers were written. Papers from 2018 or earlier predate the journal versions of the
estimators for staggered adoption
\citep{dechaisemartin2020two,callaway2021difference,sun2021estimating,borusyak2024revisiting}. Using a text search, we
exclude choices from papers published in 2018 or earlier whose descriptions mention staggered adoption or its
estimators. The silent share is then \nStaggerSilentPct, compared with \nComSilentPct{} in the full coding. The search
excludes more choices than necessary. We manually check \nStaggerCheckN{} choices from RNE papers flagged by the search
and find that \nStaggerCheckOffN{} concern neither the estimator nor the comparison group under staggered adoption.

\paragraph{Second coder.} Because the second coder does not rerun any claim, it uses four classes instead of five. It works from its own written definitions and maps
both \emph{claim not supported in our rerun} and \emph{anticipated and defended} of Table~\ref{tab:tab_commentary} to
one class, defended. The two codings agree
on whether the paper is silent for \nComTwoAgreePct{} of the choices. Cohen's kappa for agreement across the four
classes is \nComTwoKappaFour.

\subsection{Additional Details on Paper Characteristics}\label{app:predict-details}

This section reports details of the characteristics and regressions in Section~\ref{sec:predict}.

\paragraph{Coding of the order.} Three agent coders assign the \emph{order of the outcome} of each paper's main estimate
(Section~\ref{sec:predict-chars}). One coder uses a different language model (Sonnet) from the other two. The ordinal
Krippendorff's alpha is \nAlphaOrder{}, where 1 indicates perfect agreement and 0 indicates agreement at the rate
expected by chance. The coders assign the same code to \nOrderSameN{} of the \nPapers{} papers.

\paragraph{Published $t$.} When the paper reports no $t$-statistic, we compute it from the reported standard error or
$p$-value, or from the significance level implied by its stars.

\paragraph{List of papers.} We construct the \emph{repeat-author count} and the \emph{number of earlier papers} from a list of papers on the
experiments drawn from RNE (Section~\ref{sec:predict-chars}). The list includes all \nListedN{} RNE entries and the
\nCrowdScreenN{} papers that pass our citation-network screens and are not versions of RNE papers
(Section~\ref{sec:sample-added}), regardless of whether we could obtain or analyze them. These \nCrowdEntries{} entries
cover \nCrowdPapers{} distinct papers. Each entry represents one paper and experiment, so a paper studying two
experiments has two entries (Appendix~\ref{app:sample}).

\paragraph{Standard errors.} We use heteroskedasticity-robust standard errors with the HC3 small-sample adjustment. We
report coefficients on the two shares and on \emph{reproduction} in percentage points in the text and as fractions in
Table~\ref{tab:tab_joint} and Figure~\ref{fig:fig_coefs}.

\paragraph{Design.} Of the \nIvN{} instrumental-variable claims, \nIvMfsRusN{} study the mutual fund scandal or the Russell reconstitution.
The \nOthN{} other-method claims come from \nOthExpN{} experiments. A regression of the \emph{other-method dummy} on
experiment indicators has an $R^2$ of \nRsqExpOth, so experiment fixed effects absorb this share of the variation in the
indicator. Of the \nExpRegfdN{} Regulation FD claims, \nOthRegfdN{} use other methods. The mean \emph{share supported}
for Regulation FD is \nExpRegfdSharePct, compared with the sample mean of \nShareMeanPct{}
(Figure~\ref{fig:fig_experiments}).

\paragraph{Journals.} With journal group indicators and experiment fixed effects (Section~\ref{sec:predict-jyd}), claims in FT50 journals
outside the UTD24 list have a \emph{share supported} lower by \cShareFeAloneFtfiftyPp{} percentage points than claims in
top journals (\cShareFeAloneFtfiftyP). Their coefficient on the \emph{median signed $z$} is \cMedzFeAloneFtfiftyCoef{}
(\cMedzFeAloneFtfiftyP). For claims in UTD24 journals outside the top journals, the coefficient on the \emph{median
signed $z$} is \cMedzFeAloneUtdCoef{} (\cMedzFeAloneUtdP).

\paragraph{Institution cutoffs.} When we use the top 10 instead of the top 20, the coefficient of the top-10 indicator on the \emph{median signed $z$}, with experiment fixed effects, is \kMedzFeAloneToptenCoef{}
(\kMedzFeAloneToptenP) with the indicator alone and \kMedzFeJointToptenCoef{} (\kMedzFeJointToptenP) with all
characteristics. When we use the top 50 instead of the top 20, the corresponding coefficients are
\kMedzFeAloneTopfiftyCoef{} (\kMedzFeAloneTopfiftyP) and \kMedzFeJointTopfiftyCoef{} (\kMedzFeJointTopfiftyP).

\paragraph{Inference for the paper count.} For the count of all papers on each claim's experiment
(Section~\ref{sec:predict-crowd}), we enter the count in tens of papers, cluster standard errors by experiment and
obtain $p$-values from a wild cluster bootstrap.

\paragraph{Repeat-author variants.} We also examine two alternative measures of the \emph{repeat-author count}, entering
each alone. When we use the largest number of other papers among the claim's authors
instead of their mean, the \emph{share supported} is lower by \kShareFeAloneAuthmaxPp{} percentage points per
additional paper (\kShareFeAloneAuthmaxP). When we count only the authors' papers on the claim's own experiment, the
\emph{share supported} is lower by \kShareFeAloneSamemPp{} percentage points per additional paper, a difference that is
not statistically significant (\kShareFeAloneSamemP).

\subsection{Additional Details on Reproduction and Robustness}\label{app:results45-details}

This subsection gives details on the reproduction results of Section~\ref{sec:repro} and the outcomes of
Section~\ref{sec:robust}.

\paragraph{A looser reproduction cutoff.} When we relax the $p$-value cutoff to $p < 0.20$ while keeping the sign condition and the closeness bound, i.e., the
limit on how far our estimate may be from the published one (Section~\ref{sec:method-reproduce}), we reproduce
\nRepOldAllPct{} of the main estimates of the \nRepAllN{} directional claims with a published test (\nRepOldOnlyN{} more
estimates).

\paragraph{Failed reproduction conditions.} Panel~B of Table~\ref{tab:tab_reproduction} shows that \nRepFailP{} (\nRepFailPPct) of the \nRepMiss{} claims not
reproduced have a $p$-value above 0.10. For \nRepFailOnlyP{} of these claims, the $p$-value condition is the only one
that fails. Our estimate falls outside the closeness bound for \nRepFailMag{} claims and has the opposite sign for
\nRepFailSign. Among the claims with the opposite sign, \nRepSignAlsoP{} also fail the $p$-value condition, and the
others fail the closeness bound. In all, \nRepFailMulti{} claims not reproduced fail more than one condition, including
\nRepFailAllThree{} that fail all three.

\paragraph{Reproduction by experiment.} Among the \nExpK{} experiments with at least three directional claims, the
\emph{reproduction} rates range from \nRepExpMin{} for \nRepExpMinName{} to \nRepExpMax{} for \nRepExpMaxName{} (third
panel of Figure~\ref{fig:fig_experiments}).

\paragraph{Correlations among the outcomes.} Across claims, the \emph{share supported} and the \emph{median signed $z$} have a correlation of \nCorrShareMedz{}
(Table~\ref{tab:tab_outcomes}). \emph{Reproduction} has a lower correlation with each outcome: \nCorrShareRep{} with the
\emph{share supported} and \nCorrRepMedz{} with the \emph{median signed $z$}. The \emph{share with the published sign}
has a correlation of \nCorrShareSign{} with the \emph{share supported} and \nCorrSignMedz{} with the \emph{median signed
$z$}.

\paragraph{Outcomes by data source.} The mean \emph{share supported} is \nShareAltMeanPct{} for claims with
\emph{alternative data} and \nShareExactMeanPct{} for claims with exact data.

\paragraph{Differences across experiments.} Regressions of the outcomes on experiment indicators have an $R^2$ of \nExpRsqShare{} for the \emph{share supported},
\nExpRsqSign{} for the \emph{share with the published sign}, \nExpRsqRep{} for \emph{reproduction} and \nExpRsqMedz{}
for the \emph{median signed $z$}. An $F$-test rejects equal experiment means for each outcome (\nExpFShareP,
\nExpFSignP, \nExpFRepP{} and \nExpFMedzP, respectively).

\paragraph{Null claims not reproduced by the null standard.} The reproduction standard for directional claims (Section~\ref{sec:method-reproduce}) requires a significant estimate
and therefore does not apply to null claims. Our estimates for \nNullRepN{} null claims meet the directional-claim
standard, meaning that our estimates are significant although the papers report no effect. For the remaining
null claim, our estimate meets neither standard.

\section{Results for the RNE Papers}\label{app:rne}

In this section, we repeat the main analyses for the \nIPapers{} papers listed in RNE, excluding the \nAddedPapers{}
papers that we add from their citation network (Section~\ref{sec:sample-added}). Of the RNE papers, \nIDir{} make a
directional claim, and \nIReg{} of these enter the regressions. Tables~\ref{tab:tab_outcomes_201},
\ref{tab:tab_reproduction_201} and~\ref{tab:tab_joint_201} correspond to Tables~\ref{tab:tab_outcomes},
\ref{tab:tab_reproduction} and~\ref{tab:tab_joint}.

\begin{table}[!htbp]
\centering
\begin{threeparttable}
\caption{The outcomes, RNE papers only}
\label{tab:tab_outcomes_201}
\small
\begin{tabular*}{\textwidth}{@{\extracolsep{\fill}}lcccc}
\toprule
 & Share supported & \makecell{Share with the\\published sign} & Reproduced & Median signed $z$ \\
 & (1) & (2) & (3) & (4) \\
\midrule
\textit{Panel A. Pearson correlation} & & & & \\
Share supported & 1 &  &  &  \\
Share with the published sign & 0.588 & 1 &  &  \\
Reproduced & 0.481 & 0.474 & 1 &  \\
Median signed $z$ & 0.881 & 0.821 & 0.516 & 1 \\
\addlinespace
\textit{Panel B. Spearman correlation} & & & & \\
Share supported & 1 &  &  &  \\
Share with the published sign & 0.703 & 1 &  &  \\
Reproduced & 0.491 & 0.503 & 1 &  \\
Median signed $z$ & 0.859 & 0.905 & 0.547 & 1 \\
\addlinespace
\textit{Panel C. Distribution} & & & & \\
Mean & 0.234 & 0.697 & 0.439 & 0.782 \\
Standard deviation & 0.288 & 0.287 & 0.498 & 1.119 \\
Median & 0.104 & 0.787 & 0.000 & 0.727 \\
Share of claims at zero & 0.262 & 0.016 & 0.561 &  \\
Papers & 187 & 187 & 187 & 187 \\
\bottomrule
\end{tabular*}
\begin{tablenotes}[flushleft]
\footnotesize
\item[] Directional claims in the regression sample (Section~\ref{sec:robust}). Outcomes as defined in Section~\ref{sec:method-measure}.
\end{tablenotes}
\end{threeparttable}
\end{table}

\begin{table}[!htbp]
\centering
\begin{threeparttable}
\caption{Reproduction of the main estimates, RNE papers only}
\label{tab:tab_reproduction_201}
\small
\begin{tabular*}{\textwidth}{@{\extracolsep{\fill}}lcccc}
\toprule
 & \multicolumn{2}{c}{Reproduced} & \multicolumn{2}{c}{$p < 0.20$ variant} \\
\cmidrule(lr){2-3}\cmidrule(lr){4-5}
 & Papers & Share & Papers & Share \\
\midrule
\multicolumn{5}{l}{\textit{Panel A. Reproduction rate}} \\
All directional claims & 82/188 & 0.436 & 95/188 & 0.505 \\
Exact data & 66/140 & 0.471 & 76/140 & 0.543 \\
Alternative data & 16/48 & 0.333 & 19/48 & 0.396 \\
Authors' materials & 10/12 & 0.833 & 10/12 & 0.833 \\
No authors' materials & 72/176 & 0.409 & 85/176 & 0.483 \\
\addlinespace
Alternative minus exact data, FE & 187 & $-$0.072 & 187 & $-$0.034 \\
 &  & (0.092) &  & (0.093) \\
Authors' materials minus none, FE & 187 & 0.481$^{***}$ & 187 & 0.426$^{***}$ \\
 &  & (0.155) &  & (0.157) \\
\addlinespace
\multicolumn{5}{l}{\textit{Panel B. Conditions not met (share of papers not reproduced)}} \\
Papers not reproduced & 106 &  & 93 &  \\
Fail closeness & 61 & 0.575 & 61 & 0.656 \\
Fail sign & 35 & 0.330 & 35 & 0.376 \\
Fail $p$-value & 90 & 0.849 & 73 & 0.785 \\
Closeness only & 13 & 0.123 & 14 & 0.151 \\
Sign only & 0 & 0.000 & 0 & 0.000 \\
$p$-value only & 35 & 0.330 & 22 & 0.237 \\
Closeness and sign only & 3 & 0.028 & 6 & 0.065 \\
Closeness and $p$-value only & 23 & 0.217 & 22 & 0.237 \\
Sign and $p$-value only & 10 & 0.094 & 10 & 0.108 \\
All three & 22 & 0.208 & 19 & 0.204 \\
\addlinespace
\multicolumn{5}{l}{\textit{Panel C. Sample size}} \\
Not reproduced, sample size off by more than 1\% & 88/101 & 0.871 & 78/90 & 0.867 \\
Spearman correlation, sample-size gap and estimate gap & 175 & 0.320 & 175 & 0.320 \\
\bottomrule
\end{tabular*}
\begin{tablenotes}[flushleft]
\footnotesize
\item[] Rows marked FE estimate the difference by OLS with experiment fixed effects. Heteroskedasticity-robust (HC3) standard errors are in parentheses, and *, ** and *** denote significance at the 10\%, 5\% and 1\% levels.
\end{tablenotes}
\end{threeparttable}
\end{table}

\begin{table}[!htbp]
\centering
\begin{threeparttable}
\caption{Paper characteristics and the outcomes, RNE papers only}
\label{tab:tab_joint_201}
\footnotesize
\setlength{\tabcolsep}{4pt}
\begin{tabular*}{\textwidth}{@{\extracolsep{\fill}}lcccccccc}
\toprule
 & \multicolumn{2}{c}{\makecell{Share\\supported}} & \multicolumn{2}{c}{\makecell{Share with the\\published sign}} & \multicolumn{2}{c}{Reproduced} & \multicolumn{2}{c}{\makecell{Median\\signed $z$}} \\
\cmidrule(lr){2-3}\cmidrule(lr){4-5}\cmidrule(lr){6-7}\cmidrule(lr){8-9}
 & (1) & (2) & (3) & (4) & (5) & (6) & (7) & (8) \\
\midrule
First-order outcome & 0.670$^{***}$ & 0.662$^{***}$ & 0.204$^{**}$ & 0.139 & 0.435$^{**}$ & 0.470$^{***}$ & 2.478$^{***}$ & 2.414$^{***}$ \\
 & (0.065) & (0.089) & (0.099) & (0.091) & (0.209) & (0.170) & (0.295) & (0.381) \\
Third or higher order & 0.031 & 0.008 & 0.052 & 0.054 & $-$0.001 & $-$0.022 & 0.296 & 0.189 \\
 & (0.044) & (0.044) & (0.048) & (0.054) & (0.079) & (0.088) & (0.184) & (0.179) \\
UTD24 (excl. top) & 0.027 & 0.065 & 0.155 & 0.202 & 0.246 & 0.363 & 0.353 & 0.580 \\
 & (0.114) & (0.126) & (0.108) & (0.127) & (0.214) & (0.251) & (0.411) & (0.469) \\
FT50 (excl. UTD24) & $-$0.071 & $-$0.081 & $-$0.086 & $-$0.055 & $-$0.094 & $-$0.076 & $-$0.424 & $-$0.404 \\
 & (0.073) & (0.078) & (0.088) & (0.090) & (0.130) & (0.132) & (0.301) & (0.332) \\
Outside UTD24 and FT50 & $-$0.058 & $-$0.051 & $-$0.050 & $-$0.014 & $-$0.017 & 0.020 & $-$0.312 & $-$0.234 \\
 & (0.050) & (0.051) & (0.062) & (0.068) & (0.097) & (0.100) & (0.214) & (0.223) \\
Working paper & $-$0.066 & $-$0.036 & $-$0.024 & 0.038 & 0.056 & 0.118 & $-$0.314 & $-$0.108 \\
 & (0.048) & (0.050) & (0.063) & (0.070) & (0.102) & (0.114) & (0.204) & (0.221) \\
IV design & $-$0.086 & $-$0.106 & $-$0.142$^{*}$ & $-$0.144 & $-$0.206$^{**}$ & $-$0.022 & $-$0.371$^{*}$ & $-$0.223 \\
 & (0.053) & (0.101) & (0.074) & (0.133) & (0.095) & (0.159) & (0.224) & (0.474) \\
RD design & 0.349$^{*}$ & 0.323 & 0.071 & $-$0.030 & 0.296 & 0.455 & 0.689 & 0.662 \\
 & (0.178) & (0.199) & (0.165) & (0.193) & (0.247) & (0.310) & (0.709) & (0.785) \\
Other method & 0.155$^{**}$ & 0.025 & 0.079 & 0.011 & 0.308$^{***}$ & 0.227 & 0.614$^{**}$ & 0.180 \\
 & (0.064) & (0.085) & (0.053) & (0.086) & (0.112) & (0.151) & (0.250) & (0.391) \\
Alternative data & 0.003 & 0.000 & $-$0.066 & $-$0.011 & $-$0.130 & $-$0.108 & $-$0.114 & $-$0.037 \\
 & (0.043) & (0.048) & (0.053) & (0.058) & (0.081) & (0.092) & (0.183) & (0.202) \\
Authors' materials & 0.081 & 0.073 & 0.218$^{***}$ & 0.246$^{***}$ & 0.451$^{***}$ & 0.449$^{***}$ & 0.501$^{***}$ & 0.523$^{**}$ \\
 & (0.065) & (0.078) & (0.054) & (0.065) & (0.147) & (0.166) & (0.191) & (0.255) \\
Log published $|t|$ & 0.171$^{***}$ & 0.159$^{**}$ & 0.058 & 0.072 & 0.179$^{*}$ & 0.192$^{**}$ & 0.433 & 0.410 \\
 & (0.059) & (0.062) & (0.068) & (0.068) & (0.097) & (0.096) & (0.271) & (0.278) \\
Top-20 institution & 0.046 & 0.055 & 0.032 & 0.054 & 0.022 & 0.001 & 0.231 & 0.280 \\
 & (0.050) & (0.053) & (0.049) & (0.055) & (0.093) & (0.091) & (0.190) & (0.202) \\
Repeat-author count & $-$0.028$^{*}$ & $-$0.018 & $-$0.041$^{*}$ & $-$0.034 & $-$0.017 & $-$0.021 & $-$0.142$^{**}$ & $-$0.114 \\
 & (0.015) & (0.018) & (0.021) & (0.023) & (0.034) & (0.033) & (0.064) & (0.074) \\
Earlier papers (tens) & 0.012 & 0.007 & $-$0.005 & $-$0.016 & 0.020 & $-$0.007 & 0.022 & $-$0.013 \\
 & (0.013) & (0.015) & (0.014) & (0.016) & (0.024) & (0.027) & (0.044) & (0.052) \\
\addlinespace
\midrule
Experiment fixed effects & No & Yes & No & Yes & No & Yes & No & Yes \\
Papers & 187 & 187 & 187 & 187 & 187 & 187 & 187 & 187 \\
$R^2$ & 0.34 & 0.38 & 0.16 & 0.23 & 0.20 & 0.28 & 0.30 & 0.37 \\
\bottomrule
\end{tabular*}
\begin{tablenotes}[flushleft]
\footnotesize
\item[] Characteristics as defined in Section~\ref{sec:predict-chars}, with second-order outcomes, top journals and difference-in-differences as the base groups. Heteroskedasticity-robust (HC3) standard errors are in parentheses, and *, ** and *** denote significance at the 10\%, 5\% and 1\% levels.
\end{tablenotes}
\end{threeparttable}
\end{table}

The outcomes are close to those of the full sample. The mean \emph{share supported} is \nIShareMeanPct{} and the median
is \nIShareMedianPct. For \nIShareZeroPct{} of the claims, no specification keeps the directional claim. The median
claim has a \emph{median signed $z$} of \nIMedzMedian, and the mean \emph{share with the published sign} is
\nISignMeanPct. We reproduce \nIRepAllK{} of the \nIRepAllN{} main estimates of the directional claims with a published
test (\nIRepAllPct).

When we enter each characteristic alone with experiment fixed effects, the \emph{log published $|t|$} and the \emph{authors' materials} have the same relations with the \emph{share supported}, \emph{reproduction} and the \emph{median signed $z$} as in
the full sample; the relation of the \emph{log published $|t|$} with the \emph{share with the published sign} is not
statistically significant (\iSignFeAloneLogtP).
A one-unit increase in the \emph{log published $|t|$} is associated with a \emph{share
supported} higher by \iShareFeAloneLogtPp{} percentage points (\iShareFeAloneLogtP). Claims for which we find the
\emph{authors' materials} have a probability of \emph{reproduction} higher by \iRepFeAloneMatPp{} percentage points than
claims without these materials (\iRepFeAloneMatP). Claims in journals outside both the UTD24 and FT50 lists have a lower
\emph{median signed $z$} than claims in top journals (\iMedzFeAloneOthjCoef, \iMedzFeAloneOthjP).

The coefficients on the \emph{repeat-author count} (Section~\ref{sec:predict-crowd}) have the same sign as in the full
sample. With experiment fixed effects and the count alone, a one-unit increase in the \emph{repeat-author count} is associated with a
\emph{share supported} lower by \iShareFeAloneAuthmPpd{} percentage points (\iShareFeAloneAuthmP) and a coefficient of
\iMedzFeAloneAuthmCoef{} on the \emph{median signed $z$} (\iMedzFeAloneAuthmP). With all characteristics in the model,
the \emph{share supported} is lower by \iShareFeJointAuthmPpd{} percentage points (\iShareFeJointAuthmP), a difference that is not statistically significant. With the
\emph{number of earlier papers} alone in the model, ten more earlier papers are associated with a coefficient of
\iMedzFeAloneEarlierCoef{} on the \emph{median signed $z$}, compared with \cMedzFeAloneEarlierCoef{} in the
full sample. The RNE coefficient is not statistically significant (\iMedzFeAloneEarlierP). With all characteristics in the model, the coefficient is
\iMedzFeJointEarlierCoef{} (\iMedzFeJointEarlierP), which is not statistically significant, compared with \cMedzFeJointEarlierCoef{} in the full sample. Claims with an author at a \emph{top-20 institution} account for
\nITopInstPct{} of the RNE claims and are more robust when we enter the indicator alone. With experiment fixed effects,
their \emph{median signed $z$} is higher by \iMedzFeAloneTopinstCoef{} (\iMedzFeAloneTopinstP) and their \emph{share
supported} by \iShareFeAloneTopinstPp{} percentage points (\iShareFeAloneTopinstP). In the full sample, the difference in the \emph{share supported} is not statistically
significant (\cShareFeAloneTopinstP).

Two further comparisons differ from the full sample. First, the \nIOrdOneN{} RNE claims about first-order effects have a mean
\emph{share supported} of \nIOrdOneSharePct, and all of them reproduce. With experiment fixed effects and the indicator alone, their
\emph{share supported} is higher by \iShareFeAloneOrdOnePp{} percentage points than that of second-order claims
(\iShareFeAloneOrdOneP), compared with \cShareFeAloneOrdOnePp{} percentage points in the full sample
(\cShareFeAloneOrdOneP). Second, with the design indicators alone, instrumental-variable claims have a \emph{share supported} lower by
\iShareFeAloneIvPp{} percentage points than difference-in-differences claims within experiments (\iShareFeAloneIvP),
compared with \cShareFeAloneIvPp{} percentage points in the full sample (\cShareFeAloneIvP).

\section{Sample Papers}\label{app:papers}

Table~\ref{tab:tab_papers} lists the \nPapers{} sample papers by experiment. For the RNE papers, the authors, year, title
and journal are those RNE gives, after the corrections we describe in Appendix~\ref{app:sample}. For the added
papers, marked with a dagger, they are those of the version we analyze.

{\footnotesize\singlespacing
\begin{longtable}{@{}>{\raggedright\arraybackslash}p{1.6in}>{\raggedright\arraybackslash}p{3.05in}>{\raggedright\arraybackslash}p{1.45in}@{}}
\caption{Papers in the sample}\label{tab:tab_papers}\\
\toprule
Authors (year) & Title & Journal \\
\midrule
\endfirsthead
\multicolumn{3}{l}{\textit{Papers in the sample, continued}} \\
\toprule
Authors (year) & Title & Journal \\
\midrule
\endhead
\bottomrule
\endfoot
\bottomrule
\multicolumn{3}{@{}l@{}}{\parbox{\linewidth}{\vspace{4pt}Authors, year, title and journal as listed in RNE, with the corrections of Appendix~\ref{app:sample}. $^{\dagger}$Paper added from the citation network (Section~\ref{sec:sample-added}), shown as the version we analyze.}} \\
\endlastfoot
\multicolumn{3}{@{}l}{\textit{Reg SHO pilot (45)}} \\
Babenko, Choi, and Sen (2018) & Management (of) Proposals & Working paper \\
Bai, Lee, and Zhang (2020) & Capital Market Frictions and Human Capital Investment: Evidence from Workplace Safety Around Regulation SHO & Financial Review \\
Billett, Liu, and Tian (2019) & Information Spillovers and Cross Monitoring between the Stock Market and Loan Market & Working paper \\
Blau and Griffith (2016) & Price clustering and the stability of stock prices & Journal of Business Research \\
Brockman, Luo, and Xu (2020) & The impact of short-selling pressure on corporate employee relations & Journal of Corporate Finance \\
Chang, Huang, Su, and Tseng (2018) & Short-Termist CEO Compensation in Speculative Markets: A Controlled Experiment & Working paper \\
Chang, Lin, and Ma (2020) & Does Short-selling Threat Discipline Managers in Mergers and Acquisitions Decisions? & Journal of Accounting and Economics \\
Chen, Chen, and Chou (2020) & Short-Sale Constraints and Options Trading: Evidence from Reg SHO & Journal of Financial and Quantitative Analysis \\
Chen, Cheng, Luo, and Yue (2019) & Short Sellers and Long-Run Management Forecasts & Contemporary Accounting Research \\
Chen, Zhu, and Chang (2019) & Short-selling constraints and corporate payout policy & Accounting and Finance \\
Cheng and Zhang (2019) & The Disciplining Effect of Short Sellers on Credit Rating Properties: Evidence from a Natural Experiment & Working paper \\
Choi (2018) & Short selling and the rounding of analysts' forecasts & Finance Research Letters \\
Chu (2015) & Short Selling and the Product Market: Evidence from SHO & Working paper \\
Chu, Hirshleifer, and Ma (2020) & The Causal Effect of Limits to Arbitrage on Asset Pricing Anomalies & Journal of Finance \\
Clinch, Li, and Zhang (2019) & Short selling and firms' disclosure of bad news: Evidence from Regulation SHO & Journal of Financial Reporting \\
De Angelis, Grullon, and Michenaud (2017) & The Effects of Short-Selling Threats on Incentive Contracts: Evidence from an Experiment & Review of Financial Studies \\
Deng, Gao, and Kim (2020) & Short-sale Constraints and Stock Price Crash Risk: Causal Evidence from a Natural Experiment & Journal of Corporate Finance \\
Diether, Lee, and Werner (2009) & It's SHO Time! Short-Sale Price Tests and Market Quality & Journal of Finance \\
Fang, Huang, and Karpoff (2016) & Short Selling and Earnings Management & Journal of Finance \\
Francis, Samuel, and Wu (2023) & The Impact of Financial Markets on Payout Policy: Evidence from Short Selling & Working paper \\
Gao, He, and Wu (2022) & Standing Out from the Crowd via Corporate Social Responsibility: Evidence from Non-Fundamental-Driven Price Pressure & Working paper \\
Gong (2020) & Short Selling Threat and Corporate Financing Decisions & Journal of Banking and Finance \\
Grullon, Michenaud, and Weston (2015) & The Real Effects of Short-Selling Constraints & Review of Financial Studies \\
He and Tian (2016) & Do Short Sellers Exacerbate or Mitigate Managerial Myopia? Evidence from Patenting Activities & Working paper \\
Hope, Hu, and Zhao (2017) & Third-Party Consequences of Short-Selling Threats: The Case of Auditor Behavior & Journal of Accounting and Economics \\
Hvidkjaer, Massa, and Rzeznik (2018) & Informed Trading and Co-Illiquidity & Working paper \\
Jang and Lee (2018) & CEO Confidence Matters: The Real Effects of Short Sale Constraints Revisited & Working paper \\
Jia, Gao, and Julian (2019) & Do Firms Use Corporate Social Responsibility to Insure Against Stock Price Risk? Evidence from a Natural Experiment & Strategic Management Journal \\
Kan and Gong (2018) & Does High Stock return synchronicity Indicate High or Low Price Informativeness? Evidence from a regulatory experiment & International Review of Finance \\
Ke, Lo, Sheng, and Zhang (2018) & Does short selling improve analyst forecast quality? & Working paper \\
Kim, Lu, and Peng (2020) & Short-Selling Pressure and Aggressive Corporate Tax Avoidance & Working paper \\
Kubick, Omer, and Song (2019) & Short Selling and Tax Disclosure: Evidence from Regulation SHO & Working paper \\
Leung, Evans, and Mazouz (2020) & The R\&D anomaly: Risk or mispricing? & Journal of Banking and Finance \\
Li and Zhang (2015) & Short Selling Pressure, Stock Price Behavior, and Management Forecast Precision: Evidence from a Natural Experiment & Journal of Accounting Research \\
Li, Zhao, and Zhong (2015) & Migrate or not? The effects of regulation SHO on options trading activities & Review of Derivatives Research \\
Lin, Lin, and Ho (2022) & Equity Short Selling and the Bank Loan Market & Working paper \\
Lu, Yu, and Zhang (2018) & Short Selling Pressure and Corporate Social Responsibility Performance: Evidence from a Natural Experiment & Working paper \\
Maffett and Owens (2018) & Equity-Market Trading Restrictions and Credit Prices & Working paper \\
Massa, Qian, Xu, and Zhang (2015) & Competition of the informed: Does the presence of short sellers affect insider selling? & Journal of Financial Economics \\
Massa, Zhang, and Zhang (2015) & The Invisible Hand of Short Selling: Does Short Selling Discipline Earnings Management? & Review of Financial Studies \\
Su, Wong, Xue, and Zhao (2021) & Do Short-Sale Constraints Inhibit Information Acquisition? Evidence from Two Natural Experiments & Working paper \\
Sun and Xu (2018) & Short Selling and Readability in Financial Disclosures: A Controlled Experiment & Working paper \\
To, Treepongkaruna, and Wu (2018) & Are all insiders on the inside? Evidence from the initiation of CDS trading and short selling in the financial sector & Journal of International Financial Markets, Institutions and Money \\
Wang, Wang, Wei, Zhang, and Zhou (2022) & Insider Sales under the Threat of Short Sellers: New Hypothesis and New Tests & Working paper \\
Wu and Xu (2021) & Changes in Ownership Breadth and Capital Market Anomalies & Working paper \\
\addlinespace
\multicolumn{3}{@{}l}{\textit{Business combination laws (31)}} \\
Amore and Bennedsen (2016) & Corporate Governance and Green Innovation & Journal of Environmental Economics and Management \\
Atanassov (2013) & Do hostile takeovers stifle innovation? Evidence from antitakeover legislation and corporate patenting & Journal of Finance \\
Bertrand and Mullainathan (1999) & Is there discretion in wage setting? A test using takeover legislation & RAND Journal of Economics \\
Bharath and Hertzel (2019) & External Governance and Debt Structure & Review of Financial Studies \\
Bharath, Dahiya, and Hallak (2021)$^{\dagger}$ & Corporate Governance and Loan-Syndicate Structure & Journal of Financial and Quantitative Analysis \\
Bhargava, Faircloth, and Zeng (2017) & Takeover protection and stock price crash risk: Evidence from state antitakeover laws & Journal of Business Research \\
Cain, McKeon, and Solomon (2017) & Do takeover laws matter? Evidence from five decades of hostile takeovers & Journal of Financial Economics \\
Callen, Guan, and Qiu (2014) & The market for corporate control and accounting conservatism & Working paper \\
Cardella, Fairhurst, and Klasa (2015) & What determines the composition of a firm's total cash reserves? & Working paper \\
Cen, Dasgupta, and Sen (2016) & Discipline or disruption? Stakeholder relationships and the effect of takeover threat & Management Science \\
Fich, Harford, and Yore (2022) & The Effect of Takeover Protection in Quiet Life and Bonding Firms & Working paper \\
Francis, Hasan, John, and Song (2011) & Corporate governance and dividend payout policy: A test using antitakeover legislation & Financial Management \\
Gao, Jia, and Kim (2020) & The Effect of State Antitakeover Laws on Corporate Tax Avoidance & Working paper \\
Gao, Li, and Ma (2021) & Stakeholder orientation and the cost of debt: Evidence from state-level adoption of constituency statutes & Journal of Financial and Quantitative Analysis \\
Garvey and Hanka (1999)$^{\dagger}$ & Capital Structure and Corporate Control: The Effect of Antitakeover Statutes on Firm Leverage & Journal of Finance \\
Giroud and Mueller (2010) & Does corporate governance matter in competitive industries? & Journal of Financial Economics \\
Gormley and Matsa (2016) & Playing it safe? Managerial preferences, risk, and agency conflicts & Journal of Financial Economics \\
Jayaraman and Shivakumar (2013)$^{\dagger}$ & Agency-based demand for conservatism: evidence from state adoption of antitakeover laws & Review of Accounting Studies \\
Jiang and Lie (2016)$^{\dagger}$ & Cash holding adjustments and managerial entrenchment & Journal of Corporate Finance \\
John, Li, and Pang (2017)$^{\dagger}$ & Does Corporate Governance Matter More for High Financial Slack Firms? & Management Science \\
Juks (2008) & How does anti-takeover regulation affect workplace safety & Working paper \\
Karpoff and Wittry (2018) & Institutional and legal context in natural experiments: The case of state antitakeover laws & Journal of Finance \\
Kim and Purnanandam (2014) & Seasoned Equity Offerings, Corporate Governance, and Investments & Review of Finance \\
Li and Song (2018)$^{\dagger}$ & Corporate governance, accounting information environment and investment-cash flow sensitivity & International Journal of Accounting and Information Management \\
Meneghetti, Williams, and Xiao (2023)$^{\dagger}$ & The Market for Corporate Control as a Limit to Short Arbitrage & Journal of Financial and Quantitative Analysis \\
Pasquariello (2022) & Agency costs and strategic speculation in the US stock market & Working paper \\
Sauvagnat (2013) & Takeover Discipline and Asset Tangibility & Working paper \\
Tang (2018)$^{\dagger}$ & When does competition mitigate agency problems? & Journal of Corporate Finance \\
Wang and Yi (2025)$^{\dagger}$ & Takeover threats, institutional ownership, and investment efficiency & International Review of Economics and Finance \\
Zeng (2014) & Financial Constraints, Antitakeover Protection, and Corporate Innovation: An Empirical Analysis using Antitakeover Legislation & Review of Economics and Finance \\
Zhao and Chen (2009) & Earnings quality effect of state antitakeover statutes & Journal of Accounting and Public Policy \\
\addlinespace
\multicolumn{3}{@{}l}{\textit{Mutual fund scandal (20)}} \\
Anton and Polk (2014) & Connected Stocks & Journal of Finance \\
Anton, Ederer, Gine, and Schmalz (2022) & Common Ownership and Relative Performance Evaluation & Working paper \\
Banerjee, Huang, Nanda, and Xiao (2022) & Managerial Overconfidence and Market Feedback Effects & Management Science \\
Bindal (2019) & Essays in Product Market Competition & Working paper \\
Brown and Schwarz (2020) & Do market participants care about portfolio disclosure? Evidence from funds' 13F filings & Working paper \\
Cline, Fu, and Tang (2020) & Shareholder investment horizons and bank debt financing & Journal of Banking and Finance \\
Fang and Fu (2018) & The Bright Side of Earnings Management & Working paper \\
Francis, Hasan, Shen, and Ye (2021) & Stock price fragility and the cost of bank loans & Journal of Empirical Finance \\
Freeman (2021) & Overlapping ownership along the supply chain & Journal of Financial and Quantitative Analysis \\
Fu (2016) & Earnings Management in Response to Stock Price Pressure: Evidence from Fire Sales (and Purchases) by Mutual Funds & Working paper \\
Goncalves-Pinto and Schmidt (2022) & Co-insurance in mutual fund families & Working paper \\
Huang, Li, and Zhang (2020) & The Demand-driven Information Market & Working paper \\
Jakab (2018) & Competitor Scale and Mutual Fund Behavior & Working paper \\
Jylh\"a, Suominen, and Tomunen (2017) & Beta Bubbles & Review of Asset Pricing Studies \\
Kadach (2016) & Management Earnings Forecasts during Price Pressure: Evidence from Mutual Funds Trades & Working paper \\
Kang and Si (2021) & Shareholder Financial Difficulties and Firms' Risk-Shifting Behavior: Evidence from the 2003 Mutual Fund Scandal & Working paper \\
Kisin (2011) & The impact of shareholders on corporate investment: Evidence from mutual fund holdings & Working paper \\
Koch, Ruenzi, and Starks (2016) & Commonality in liquidity: a demand-side explanation & Review of Financial Studies \\
Qian and Shi (2018) & Style Investing, Mutual Fund Flows, and Return Comovement & Journal of Portfolio Management \\
Semov (2017) & Common ownership, competition and firm financial policy & Working paper \\
\addlinespace
\multicolumn{3}{@{}l}{\textit{Brokerage closures and mergers (36)}} \\
Adhikari (2016) & Causal effect of analyst following on corporate social responsibility & Journal of Corporate Finance \\
Akyol, Qian, and Yu (2023)$^{\dagger}$ & How do experienced analysts improve price efficiency? & Journal of Banking and Finance \\
Ali, Azam, Baghdadi, Hasan, and Puwanenthiren (2026)$^{\dagger}$ & Analyst career concerns and stock price crash risk & International Review of Financial Analysis \\
Allen, Francis, Wu, and Zhao (2016)$^{\dagger}$ & Analyst coverage and corporate tax aggressiveness & Journal of Banking and Finance \\
Anantharaman and Zhang (2011) & Cover me: Managers' responses to changes in analyst coverage in the post-regulation FD period & The Accounting Review \\
Billett, Garfinkel, and Yu (2017) & The effect of asymmetric information on product market outcomes & Journal of Financial Economics \\
Chan, Guo, Wang, and Yang (2022) & Organization capital and analyst coverage & Journal of Empirical Finance \\
Chen and Lin (2017) & Does information asymmetry affect corporate tax aggressiveness? & Journal of Financial and Quantitative Analysis \\
Chen, Chiu, and Shevlin (2018) & Do analysts matter for corporate tax planning? Evidence from a natural experiment & Contemporary Accounting Research \\
Chen, Kelly, and Wu (2020) & Sophisticated investors and market efficiency: Evidence from a natural experiment & Journal of Financial Economics \\
Derrien and Kecsk\'es (2013) & The real effects of financial shocks: Evidence from exogenous changes in analyst coverage & Journal of Finance \\
Dong, Lin, and Zhan (2017) & Stock Analysts and Corporate Social Responsibility & Working paper \\
Gao, Shen, and Chan (2021)$^{\dagger}$ & Does analyst following restrain tunneling? Evidence from brokerage closures and mergers & Finance Research Letters \\
Guo, Pérez-Castrillo, and Toldrà-Simats (2019)$^{\dagger}$ & Firms’ innovation strategy under the shadow of analyst coverage & Journal of Financial Economics \\
Hameed, Morck, Shen, and Yeung (2015)$^{\dagger}$ & Information, Analysts, and Stock Return Comovement & Review of Financial Studies \\
He and Tian (2013) & The dark side of analyst coverage: The case of innovation & Journal of Financial Economics \\
Hong and Kacperczyk (2010) & Competition and Bias & Quarterly Journal of Economics \\
Huang, Pereira, and Wang (2017) & Analyst Coverage and the Likelihood of Meeting or Beating Analyst Earnings Forecasts & Contemporary Accounting Research \\
Imbierowicz and Streitz (2024)$^{\dagger}$ & Financial debt contracting and managerial agency problems & Financial Management \\
Irani and Oesch (2013) & Monitoring and corporate disclosure: Evidence from a natural experiment & Journal of Financial Economics \\
Irani and Oesch (2016) & Analyst Coverage and Real Earnings Management: Quasi-Experimental Evidence & Journal of Financial and Quantitative Analysis \\
Israelsen (2016)$^{\dagger}$ & Does Common Analyst Coverage Explain Excess Comovement? & Journal of Financial and Quantitative Analysis \\
Kelly and Ljungqvist (2012) & Testing asymmetric-information asset pricing models & Review of Financial Studies \\
Kim, Kim, and Shim (2023)$^{\dagger}$ & Do Financial Analysts Herd? & Global Economic Review \\
Li (2020) & Does Information Asymmetry Impede Market Efficiency? Evidence from Analyst Coverage & Journal of Banking and Finance \\
Li, Lin, and Zhan (2019)$^{\dagger}$ & Does Change in the Information Environment Affect Financing Choices? & Management Science \\
Luong, Qiu, and Wu (2021)$^{\dagger}$ & Does it pay to be socially connected with wall street brokerages? Evidence from cost of equity & Journal of Corporate Finance \\
Nguyen, Cheong, and Zurbruegg (2021)$^{\dagger}$ & Brokerage M\&As and the peer effect on analyst forecast accuracy & International Review of Financial Analysis \\
Nguyen, Cheong, Truong, and Zurbruegg (2023)$^{\dagger}$ & Experience of a Job Loss and Analysts’ Subsequent Performance & European Accounting Review \\
Phua, Tham, and Wei (2023)$^{\dagger}$ & Peer Effects in Equity Research & Journal of Financial and Quantitative Analysis \\
Qian, Lu, and Yu (2019) & Financial analyst coverage and corporate social performance: Evidence from natural experiments & Strategic Management Journal \\
Shahhosseini (2022)$^{\dagger}$ & The Effect of Asymmetric Information on the Pricing of Equity Issuance & International Journal of Finance \\
To, Navone, and Wu (2018) & Analyst coverage and the quality of corporate investment decisions & Journal of Corporate Finance \\
Wang, Sun, and John Wei (2020)$^{\dagger}$ & Does competition induce analyst effort? evidence from a natural experiment of broker mergers & Journal of Banking and Finance \\
Young (2018)$^{\dagger}$ & Do analysts affect bad news timeliness? & Accounting and Business Research \\
Zhang and Toffanin (2018)$^{\dagger}$ & The information environment of the firm and the market valuation of R\&D & Journal of Business Finance and Accounting \\
\addlinespace
\multicolumn{3}{@{}l}{\textit{Universal demand laws (25)}} \\
Adhikari, Agrawal, and Sharma (2022) & Does litigation risk deter insider trading? Evidence from universal demand laws & Working paper \\
Appel (2019) & Governance by litigation & Working paper \\
Bilokha, Kong, and Micale (2019) & Shareholder Litigation Risk and Stakeholders: Evidence from the Auditor Perspective & Working paper \\
Bourveau, Lou, and Wang (2018) & Shareholder litigation and corporate disclosure: Evidence from derivative lawsuits & Journal of Accounting Research \\
Chen, Li, and Xu (2021) & Universal demand laws and the monitoring demand for accounting conservatism & Journal of Business Finance and Accounting \\
Chu and Zhao (2021) & The dark side of shareholder litigation: Evidence from corporate takeovers & Financial Management \\
Do (2021)$^{\dagger}$ & Shareholder litigation rights and corporate payout policy: Evidence from universal demand laws & Research in International Business and Finance \\
Foroughi, Marcus, Nguyen, and Tehranian (2022) & Peer effects in corporate governance practices: Evidence from universal demand laws & Review of Financial Studies \\
Gao, Ma, Ng, and Wu (2022)$^{\dagger}$ & The Sound of Silence: What Do We Know When Insiders Do Not Trade? & Management Science \\
Hayes, Peng, and Wang (2020) & Shareholder Lawsuits and CEO Turnover Decisions & Working paper \\
Houston, Lin, and Xie (2018) & Shareholder protection and the cost of capital & Journal of Law and Economics \\
Huang, Li, Yu, and Zhou (2020) & The effect of managerial litigation risk on earnings warnings: Evidence from a natural experiment & Journal of Accounting Research \\
Huang, Ozkan, and Xu (2023)$^{\dagger}$ & Shareholder Litigation Risk and Firms’ Choice of External Growth & Journal of Financial and Quantitative Analysis \\
Le, Nguyen, and Sila (2019) & Does Shareholder Litigation Risk Cause Public Firms to Delist? Evidence from Universal Demand Laws & Working paper \\
Le, Nguyen, and Sila (2021)$^{\dagger}$ & Does shareholder litigation affect the corporate information environment? & Journal of Financial Markets \\
Li, Monroe, and Coulton (2018) & Litigation risk and cost behavior: Evidence from derivative lawsuits & Working paper \\
Li, Monroe, and Coulton (2018) & Managerial litigation risk and corporate investment efficiency: Evidence from derivative lawsuits & Working paper \\
Lin, Liu, and Manso (2021) & Shareholder litigation and corporate innovation & Management Science \\
Luo (2024)$^{\dagger}$ & Shareholder litigation risk and stock returns & Financial Review \\
Luu, Nguyen, Nguyen, Nguyen, and Johari (2024)$^{\dagger}$ & Shareholder Litigation Rights and Bank Dividends & Journal of Financial Services Research \\
Manchiraju, Pandey, and Subramanyam (2021) & Shareholder litigation and conservative accounting: Evidence from universal demand laws & The Accounting Review \\
Nguyen, Phan, and Sun (2018) & Shareholder litigation rights and corporate cash holdings: Evidence from universal demand laws & Journal of Corporate Finance \\
Ni and Yin (2018) & Shareholder litigation rights and the cost of debt: Evidence from derivative lawsuits & Journal of Corporate Finance \\
Treepongkaruna and Padungsaksawasdi (2024)$^{\dagger}$ & Green innovation and shareholder litigation rights & Finance Research Letters \\
Treepongkaruna, Kyaw, and Jiraporn (2022)$^{\dagger}$ & Shareholder litigation rights and ESG controversies: A quasi-natural experiment & International Review of Financial Analysis \\
\addlinespace
\multicolumn{3}{@{}l}{\textit{Russell index reconstitution (38)}} \\
Agarwal, Hanouna, Moussawi, and Stahel (2018) & Do ETFs increase the commonality in liquidity of underlying stocks? & Working paper \\
Ahn and Patatoukas (2021) & Identifying the Effect of Stock Indexing: Impetus or Impediment to Arbitrage and Price Discovery? & Journal of Financial and Quantitative Analysis \\
Appel, Gormley, and Keim (2016)$^{\dagger}$ & Passive investors, not passive owners & Journal of Financial Economics \\
Baghdadi, Bhatti, Nguyen, and Podolski (2018) & Skill or effort? Institutional ownership and managerial efficiency & Journal of Banking and Finance \\
Ben-David, Franzoni, and Moussawi (2018) & Do ETFs increase volatility? & Journal of Finance \\
Billett, Garfinkel, and Jiang (2023)$^{\dagger}$ & The capital supply channel in peer effects: The case of SEOs & Journal of Banking and Finance \\
Bird and Karolyi (2016) & Do institutional investors demand public disclosure? & Review of Financial Studies \\
Bird and Karolyi (2017)$^{\dagger}$ & Governance and Taxes: Evidence from Regression Discontinuity & The Accounting Review \\
Boone and White (2015) & The effect of institutional ownership on firm transparency and information production & Journal of Financial Economics \\
Brooks, Chen, and Zeng (2018)$^{\dagger}$ & Institutional cross-ownership and corporate strategy: The case of mergers and acquisitions & Journal of Corporate Finance \\
Chang, Hong, and Liskovich (2015) & Regression discontinuity and the price effects of stock market indexing & Review of Financial Studies \\
Chen, Dong, and Lin (2020) & Institutional shareholders and corporate social responsibility & Journal of Financial Economics \\
Chen, Huang, Li, and Shevlin (2019)$^{\dagger}$ & How does quasi-indexer ownership affect corporate tax planning? & Journal of Accounting and Economics \\
Chen, Ma, Wu, and Zhang (2023)$^{\dagger}$ & Does common ownership constrain managerial rent extraction? Evidence from insider trading profitability & Journal of Corporate Finance \\
Coles, Heath, and Ringgenberg (2022) & On index investing & Journal of Financial Economics \\
Crane, Michenaud, and Weston (2016) & The effect of institutional ownership on payout policy: Evidence from index thresholds & Review of Financial Studies \\
Cremers, Pareek, and Sautner (2020)$^{\dagger}$ & Short-Term Investors, Long-Term Investments, and Firm Value: Evidence from Russell 2000 Index Inclusions & Management Science \\
Dai, Parwada, Winchester, and Zhang (2022)$^{\dagger}$ & The more we know, the less we agree: A test of the trading horizon heterogeneity theory & Financial Review \\
Dong, Eugster, and Vazquez (2024)$^{\dagger}$ & Passive Investors and Audit Quality: Evidence from the U.S. & European Accounting Review \\
El Kalak and Tosun (2022)$^{\dagger}$ & ETF ownership and corporate cash holdings & European Financial Management \\
Fich, Harford, and Tran (2015) & Motivated monitors: The importance of institutional investors' portfolio weights & Journal of Financial Economics \\
Fu, Tang, and Yan (2019) & Why do institutions like corporate social responsibility investments? Evidence from horizon heterogeneity & Journal of Empirical Finance \\
Ghaly, Dang, and Stathopoulos (2020)$^{\dagger}$ & Institutional investors' horizons and corporate employment decisions & Journal of Corporate Finance \\
Glosten, Nallareddy, and Zou (2021)$^{\dagger}$ & ETF Activity and Informational Efficiency of Underlying Securities & Management Science \\
Heath, Macciocchi, Michaely, and Ringgenberg (2021)$^{\dagger}$ & Do Index Funds Monitor? & Review of Financial Studies \\
Hillegeist and Weng (2021)$^{\dagger}$ & Quasi‐Indexer Ownership and Insider Trading: Evidence from Russell Index Reconstitutions* & Contemporary Accounting Research \\
Hou and Zhang (2025)$^{\dagger}$ & Do passive investors influence corporate social responsibility? A risk‐management perspective & International Review of Finance \\
Hwang, Wong, and Yi (2022) & What explains the dispersion effect? Evidence from institutional ownership & Pacific-Basin Finance Journal \\
Im, Selvam, and Tan (2024)$^{\dagger}$ & Effect of stock liquidity on the economic value of patents: Evidence from U.S. patent data & International Review of Financial Analysis \\
Jang and Kang (2024)$^{\dagger}$ & ETF and corporate reporting & Financial Review \\
Jang and Lee (2022)$^{\dagger}$ & Taking a long view: Investor trading horizon and earnings management strategy & International Review of Finance \\
Khan, Srinivasan, and Tan (2017) & Institutional ownership and corporate tax avoidance: New evidence & The Accounting Review \\
Lin, Mao, and Wang (2018)$^{\dagger}$ & Institutional Ownership, Peer Pressure, and Voluntary Disclosures & The Accounting Review \\
Lin, Mei, Tan, and Zhang (2024)$^{\dagger}$ & Do exchange traded funds affect corporate cash holdings? & Journal of Business Finance and Accounting \\
Schmidt and Fahlenbrach (2017) & Do exogenous changes in passive institutional ownership affect corporate governance and firm value? & Journal of Financial Economics \\
von Beschwitz, Honkanen, and Schmidt (2025)$^{\dagger}$ & Passive ownership and short selling & Review of Finance \\
Ward, Yin, and Zeng (2020) & Motivated monitoring by institutional investors and firm investment efficiency & European Financial Management \\
Yin, Ward, and Tsolacos (2018)$^{\dagger}$ & Motivated monitoring: The importance of the institutional investment horizon & International Review of Financial Analysis \\
\addlinespace
\multicolumn{3}{@{}l}{\textit{Decimalization (27)}} \\
Amihud and Levi (2022) & The effect of stock liquidity on the firm's investment and production & Review of Financial Studies \\
Bessembinder (2003)$^{\dagger}$ & Trade Execution Costs and Market Quality after Decimalization & Journal of Financial and Quantitative Analysis \\
Brogaard, Li, and Xia (2017) & Stock liquidity and default risk & Journal of Financial Economics \\
Chang, Chen, and Zolotoy (2017) & Stock liquidity and stock price crash risk & Journal of Financial and Quantitative Analysis \\
Chang, Lin, and Ma (2020)$^{\dagger}$ & Governance through trading on acquisitions of public firms & Journal of Corporate Finance \\
Chang, Tan, Yang, and Zhang (2018) & Stock Liquidity and Corporate Social Responsibility & Working paper \\
Chen, Ge, Louis, and Zolotoy (2019) & Stock liquidity and corporate tax avoidance & Review of Accounting Studies \\
Chen, Gong, and Muckley (2020)$^{\dagger}$ & Stock market illiquidity, bargaining power and the cost of borrowing & Journal of Empirical Finance \\
Chen, Rhee, Veeraraghavan, and Zolotoy (2015) & Stock liquidity and managerial short-termism & Journal of Banking and Finance \\
Cheng and Fang (2023)$^{\dagger}$ & Stock liquidity and the accrual anomaly & China Accounting and Finance Review \\
Cheung, Chung, and Fung (2015) & The effects of stock liquidity on firm value and corporate governance: Endogeneity and the REIT experiment & Journal of Corporate Finance \\
Chou, Lee, and Chen (2005)$^{\dagger}$ & The Market Reaction Around Ex-Dates of Stock Splits Before and After Decimalization & Review of Pacific Basin Financial Markets and Policies \\
Dai, Goncalves-Pinto, and Xu (2019) & How does illiquidity affect delegated portfolio choice? & Journal of Financial and Quantitative Analysis \\
Ee, Hasan, and Huang (2022) & Stock liquidity and corporate labor investment & Journal of Corporate Finance \\
Fang, Myers, and Raymundo (2020) & Managers' Use of External Information in Their Accrual Estimates: Evidence from the Impact of Stock Liquidity on Accruals Quality & Working paper \\
Fang, Noe, and Tice (2009) & Stock market liquidity and firm value & Journal of Financial Economics \\
Gong (2024)$^{\dagger}$ & Stock liquidity and tone of press releases & Accounting and Finance \\
Graham, Michaely, and Roberts (2003) & Do price discreteness and transactions costs affect stock returns? Comparing ex-dividend pricing before and after decimalization & Journal of Finance \\
Hardin, Liano, Huang, and Nagel (2007)$^{\dagger}$ & REITs, Decimalization, and Ex-dividend Stock Prices & Journal of Real Estate Finance and Economics \\
Hu, Li, and Zeng (2019) & Stock liquidity and corporate cash holdings & Finance Research Letters \\
Huang, Maharjan, and Nanda (2024)$^{\dagger}$ & Liquid stock as an acquisition currency & Journal of Corporate Finance \\
Im, Noe, and Cheung (2026)$^{\dagger}$ & Liquidity, Information Production, and Corporate Financing Decisions & Journal of Business Finance and Accounting \\
Marshall, Nguyen, Nguyen, and Visaltanachoti (2021) & Does a change in the information environment affect labor adjustment costs? & International Review of Financial Analysis \\
Michaely and Qian (2022)$^{\dagger}$ & Does stock market liquidity affect dividends? & Pacific-Basin Finance Journal \\
R\"osch, Subrahmanyam, and Van Dijk (2022) & Investor short-termism and real investment & Journal of Financial Markets \\
Spiropoulos and Zhao (2023)$^{\dagger}$ & Stock liquidity, cash flow sensitivity and the value of cash & International Review of Economics and Finance \\
Zhao and Chung (2006)$^{\dagger}$ & Decimal Pricing and Information‐Based Trading: Tick Size and Informational Efficiency of Asset Price & Journal of Business Finance and Accounting \\
\addlinespace
\multicolumn{3}{@{}l}{\textit{Tick Size Pilot (20)}} \\
Ahmed, Li, and Xu (2020) & Tick size and financial reporting quality in small-cap firms: Evidence from a natural experiment & Journal of Accounting Research \\
Ahmed, Li, McMartin, and Xu (2026)$^{\dagger}$ & The Rise of Machines: Algorithmic Trading and Stock Price Crash Risk & Journal of Accounting, Auditing and Finance \\
Albuquerque, Song, and Yao (2020) & The price effects of liquidity shocks: A study of the SEC's tick size experiment & Journal of Financial Economics \\
Ashraf, Saleem, Qi, and Shakill (2026)$^{\dagger}$ & Tick size increase and default risk of small-cap U.S. firms: Evidence from a natural experiment & Journal of Financial Markets \\
Barbopoulos, Putniņš, and Rzayev (2026)$^{\dagger}$ & The investment effects of dark trading & Journal of Financial Markets \\
Brogaard and Pan (2022) & Dark pool trading and information acquisition & Review of Financial Studies \\
Chen, Huffman, Narayanamoorthy, and Zhang (2021) & Minimum tick size and analyst coverage: Evidence from the Tick Size Pilot Program & Journal of Business Finance and Accounting \\
Chen, Ng, and Yang (2021) & Informed Investors and Earnings Guidance: Evidence from SEC Tick Size Pilot Program & Working paper \\
Deng, Lin, and Zhou (2022) & Does better liquidity for large orders attract institutional investors and analysts? & Working paper \\
Griffith, Roseman, and Shang (2020) & The effects of an increase in equity tick size on stock and option transaction costs & Journal of Banking and Finance \\
Hope and Liu (2022) & Does stock liquidity shape voluntary disclosure? Evidence from the SEC tick size pilot program & Review of Accounting Studies \\
Kim, Oh, Seo, and Zuo (2026)$^{\dagger}$ & Algorithmic trading and forced CEO turnover: a learning hypothesis & Review of Accounting Studies \\
Lee and Watts (2021) & Tick size tolls: Can a trading slowdown improve earnings news discovery? & The Accounting Review \\
Li and Xia (2021) & Gauging the effects of stock liquidity on earnings management: Evidence from the SEC tick size pilot test & Journal of Corporate Finance \\
Lin, Yao, and Zou (2026)$^{\dagger}$ & The real effects of tick-size adjustments: Evidence from the 2016 tick-size pilot & Journal of Corporate Finance \\
THOMAS, WANG, and ZHANG (2024)$^{\dagger}$ & Algorithmic Trading and Forward‐Looking MD\&A Disclosures & Journal of Accounting Research \\
Thomas, Zhang, and Zhu (2021) & Dark trading and post-earnings-announcement drift & Management Science \\
Thomas, Zhang, and Zhu (2022) & Measuring the information content of disclosures: The role of return noise & The Accounting Review \\
Ye, Zheng, and Zhu (2022) & The Effect of Tick Size on Managerial Learning from Stock Prices & Journal of Accounting and Economics \\
Zheng and Zhu (2023)$^{\dagger}$ & Algorithmic trading and block ownership initiation: An information perspective & British Accounting Review \\
\addlinespace
\multicolumn{3}{@{}l}{\textit{Regulation FD (48)}} \\
Agrawal, Chadha, and Chen (2006) & Who Is Afraid of Reg FD? The Behavior and Performance of Sell-Side Analysts Following the SEC's Fair Disclosure Rules & Journal of Business \\
Ahmed and Schneible (2007) & The impact of regulation fair disclosure on investors' prior information quality: evidence from an analysis of changes in trading volume and stock price reactions to earnings announcements & Journal of Corporate Finance \\
Akono, Karim, and Nwaeze (2019)$^{\dagger}$ & Analyst rounding of EPS forecasts and stock recommendations & Advances in Accounting \\
Albring, Banyi, Dhaliwal, and Pereira (2016)$^{\dagger}$ & Does the Firm Information Environment Influence Financing Decisions? A Test Using Disclosure Regulation & Management Science \\
Albring, Huang, Pereira, and Xu (2020)$^{\dagger}$ & Disclosure and liquidity management: Evidence from regulation fair disclosure & Journal of Contemporary Accounting and Economics \\
Bailey, Li, Mao, and Zhong (2003) & Regulation fair disclosure and earnings information: Market, analyst, and corporate responses & Journal of Finance \\
Barniv, Hope, Myring, and Thomas (2009)$^{\dagger}$ & Do Analysts Practice What They Preach and Should Investors Listen? Effects of Recent Regulations & The Accounting Review \\
BHOJRAJ, JUN CHO, and YEHUDA (2012)$^{\dagger}$ & Mutual Fund Family Size and Mutual Fund Performance: The Role of Regulatory Changes & Journal of Accounting Research \\
Cai, Fu, Lau, and Yin (2026)$^{\dagger}$ & Earnings Announcement Informativeness and Expected Stock Returns & Journal of Accounting, Auditing and Finance \\
Canace, Caylor, Johnson, and Lopez (2010) & The effect of Regulation Fair Disclosure on expectations management: International evidence & Journal of Accounting and Public Policy \\
Chen, Dhaliwal, and Xie (2010) & Regulation fair disclosure and the cost of equity capital & Review of Accounting Studies \\
Chen, Huang, and Lin (2016)$^{\dagger}$ & Changes in analyst following for less covered firms accompanying Regulation Fair Disclosure: the roles of ability and industry experience & Review of Quantitative Finance and Accounting \\
Cheng, Huang, and Luo (2020)$^{\dagger}$ & Corporate disclosure quality and institutional investors' holdings during market downturns & Journal of Corporate Finance \\
Chichernea and Franz (2016)$^{\dagger}$ & The Effect of Regulation on the Relationship between Earnings and Stock Returns & Accounting and Finance Research \\
Cook and Tang (2010) & The impact of regulation FD on institutional investor informativeness & Financial Management \\
Dong, Hu, Li, and Liu (2017)$^{\dagger}$ & Analyst Firm Coverage and Forecast Accuracy: The Effect of Regulation Fair Disclosure & Abacus \\
Ferreira and Smith (2006)$^{\dagger}$ & Effect of Reg FD on Information in Analysts’ Rating        Changes & Financial Analysts Journal \\
Findlay and Mathew (2006)$^{\dagger}$ & An Examination of the Differential Impact of Regulation FD on Analysts' Forecast Accuracy & Financial Review \\
Gintschel and Markov (2004) & The effectiveness of Regulation FD & Journal of Accounting and Economics \\
Gomes, Gorton, and Madureira (2007)$^{\dagger}$ & SEC Regulation Fair Disclosure, information, and the cost of capital & Journal of Corporate Finance \\
Hahn and Song (2013)$^{\dagger}$ & Regulation Fair Disclosure and analysts’ reliance on earnings announcements & Journal of Accounting and Public Policy \\
He and Marginson (2020)$^{\dagger}$ & The impact of insider trading on analyst coverage and forecasts & Accounting Research Journal \\
Heflin, Subramanyam, and Zhang (2003) & Regulation FD and the Financial Information Environment: Early Evidence & The Accounting Review \\
Herrmann, Hope, and Thomas (2008)$^{\dagger}$ & International Diversification and Forecast Optimism: The Effects of Reg FD & Accounting Horizons \\
Irani and Karamanou (2003)$^{\dagger}$ & Regulation Fair Disclosure, Analyst Following, and Analyst Forecast Dispersion & Accounting Horizons \\
Irani and Karamanou (2004)$^{\dagger}$ & A STUDY OF THE ECONOMIC CONSEQUENCES OF REGULATION FD (FAIR DISCLOSURE) & Research in Accounting Regulation \\
Jackson and Madura (2007)$^{\dagger}$ & Impact of regulation fair disclosure on the information flow associated with profit warnings & Journal of Economics and Finance \\
Jiang, Nanda, and Xiao (2021)$^{\dagger}$ & Stock-market disruptions and corporate disclosure policies & Journal of Corporate Finance \\
Jorion, Liu, and Shi (2005) & Informational effects of regulation FD: evidence from rating agencies & Journal of Financial Economics \\
Ke, Petroni, and Yu (2008) & The Effect of Regulation FD on Transient Institutional Investors' Trading Behavior & Journal of Accounting Research \\
Kross and Suk (2012) & Does Regulation FD work? Evidence from analysts' reliance on public disclosure & Journal of Accounting and Economics \\
Kwag and Small (2007)$^{\dagger}$ & The impact of regulation fair disclosure on earnings management and analyst forecast bias & Journal of Economics and Finance \\
Kwag (2014)$^{\dagger}$ & A Behavioral Shift in Earnings Response After Regulation FD & Journal of Behavioral Finance \\
Lawrence, Karels, Prakash, and Shankar (2011)$^{\dagger}$ & Effect of regulation FD on disclosures of information by firms & Applied Financial Economics \\
Li and Zhang (2021)$^{\dagger}$ & Firms’ Disclosure Policies and Capital Investment: Evidence From Regulation Fair Disclosure & Journal of Accounting, Auditing and Finance \\
Li (2015)$^{\dagger}$ & Regulation Fair Disclosure and the Determinants of Analyst Coverage & Journal of Finance and Accounting \\
Li, Saunders, and Shao (2015)$^{\dagger}$ & Did Regulation Fair Disclosure affect credit markets? & Journal of Banking and Finance \\
Lo and Xu (2013)$^{\dagger}$ & Regulation FD and analysts’ vs. investors’ weightings of the cash components of earnings & Research in Accounting Regulation \\
Mathew, Michayluk, and Kofman (2007)$^{\dagger}$ & Are foreign issuers complying with Regulation Fair Disclosure? & Journal of International Financial Markets, Institutions and Money \\
Mo and Kim (2016)$^{\dagger}$ & Public Debt Issuance Impact on Management Forecasting Frequency: Evidence from Regulation Fair Disclosure & Asia-Pacific Journal of Financial Studies \\
Mohanram and Sunder (2006) & How Has Regulation FD Affected the Operations of Financial Analysts? & Contemporary Accounting Research \\
Palmon and Yezegel (2011)$^{\dagger}$ & Analysts’ Recommendation Revisions and Subsequent Earnings Surprises & Journal of Accounting, Auditing and Finance \\
Petruska (2011)$^{\dagger}$ & The Informativeness Of Cash Flow Forecasts And The Regulation FD Environment & Journal of Business and Economics Research (JBER) \\
Sinha and Gadarowski (2010)$^{\dagger}$ & The Efficacy of Regulation Fair Disclosure & Financial Review \\
Srinidhi, Leung, and Jaggi (2009)$^{\dagger}$ & Differential effects of regulation FD on short- and long-term analyst forecasts & Journal of Accounting and Public Policy \\
Wang (2007)$^{\dagger}$ & Private Earnings Guidance and Its Implications for Disclosure Regulation & The Accounting Review \\
Yu and Webb (2017)$^{\dagger}$ & Market adaptation to Regulation Fair Disclosure: The use of industry information to enhance the informational environment & Journal of Economics and Business \\
Ziebart, Chi, and Ross (2026)$^{\dagger}$ & Insights into the effects of regulation fair disclosure on stock price efficiency using Spatio-Temporal entropy & Review of Quantitative Finance and Accounting \\
\addlinespace
\multicolumn{3}{@{}l}{\textit{EDGAR phase-in (12)}} \\
Abhyankar et al. (2020) & The Impact of Enhanced Information Flows Across Firms: Cross-Sectional Evidence From the EDGAR Roll-Out & Working paper \\
Asthana and Balsam (2001) & The effect of EDGAR on the market reaction to 10-K filings & Journal of Accounting and Public Policy \\
Bowles, Reed, Ringgenberg, and Thornock (2022) & Anomaly Time & Working paper \\
Gao and Huang (2020)$^{\dagger}$ & Informing the Market: The Effect of Modern Information Technologies on Information Production & Review of Financial Studies \\
GOLDSTEIN, YANG, and ZUO (2023)$^{\dagger}$ & The Real Effects of Modern Information Technologies: Evidence from the EDGAR Implementation & Journal of Accounting Research \\
Gomez (2020) & The effect of mandatory disclosure dissemination on information asymmetry: Evidence from the implementation of the EDGAR System & Working paper \\
Gu, Li, and Ni (2022) & Too Much to Learn: The Unintended Consequences of RegTech Development on Mergers and Acquisitions & Working paper \\
Kim (2022) & Do Information Acquisition Costs Matter?: The Effect of SEC EDGAR on Stock Anomalies & Working paper \\
Kim and Valentine (2022) & Corporate Financial Disclosures and the Market for Innovation & Working paper \\
Lai, Lin, and Ma (2022) & Regtech Adoption and the Cost of Capital & Working paper \\
Ni, Wang, and Yin (2021) & Does Modern Information Technology Attenuate Managerial Information Hoarding? Evidence from the EDGAR Implementation & Journal of Corporate Finance \\
Shi (2021) & How Does Better Access to Public Firm Disclosures Affect Private Firm Financing? & Working paper \\
\addlinespace
\multicolumn{3}{@{}l}{\textit{BlackRock-BGI acquisition (7)}} \\
Azar, Qiu, and Sojourner (2021) & Common Ownership Reduces Wages and Employment & Working paper \\
Azar, Schmalz, and Tecu (2018) & Anticompetitive effects of common ownership & Journal of Finance \\
Dai and Qiu (2021) & Common ownership and corporate social responsibility & Review of Corporate Finance Studies \\
Friberg, Goldstein, and Hankins (2024)$^{\dagger}$ & Corporate responses to stock price fragility & Journal of Financial Economics \\
Gilje, Gormley, and Levit (2021) & Who's paying attention? Measuring common ownership and its impact on managerial incentives & Journal of Financial Economics \\
Liang (2016) & Common ownership and executive compensation & Working paper \\
Pawliczek and Skinner (2018) & Common ownership and voluntary disclosure & Working paper \\
\addlinespace
\multicolumn{3}{@{}l}{\textit{Dodd-Frank Act (14)}} \\
Akono (2024)$^{\dagger}$ & The impact Dodd Frank's Act on audit firm tenure & Journal of Corporate Accounting and Finance \\
Allen, Cyree, Whitledge, and Winters (2018)$^{\dagger}$ & An event study analysis of too-big-to-fail after the Dodd-Frank act: Who is too big to fail? & Journal of Economics and Business \\
Andriosopoulos, Chan, Dontis-Charitos, and Staikouras (2017)$^{\dagger}$ & Wealth and risk implications of the Dodd-Frank Act on the U.S. financial intermediaries & Journal of Financial Stability \\
Balasubramnian and Cyree (2014) & Has market discipline on banks improved after the Dodd-Frank Act? & Journal of Banking and Finance \\
Basnet, Blomkvist, Felixson, Liljeblom, and Vyas (2026)$^{\dagger}$ & The Impact of the Dodd-Frank Act on Acquisition Activity & Journal of Law and Economics \\
Bedendo, Cathcart, and El-Jahel (2018)$^{\dagger}$ & Reputational shocks and the information content of credit ratings & Journal of Financial Stability \\
Bordo and Duca (2018) & The impact of the Dodd-Frank Act on small business & Working paper \\
Cohn, Gillan, and Hartzell (2016) & On enhancing shareholder control: A (Dodd-) Frank assessment of proxy access & Journal of Finance \\
Cumming, Dai, and Johan (2020) & Dodd-Franking the hedge Funds & Journal of Banking and Finance \\
Dimitrov, Palia, and Tang (2015) & Impact of the Dodd-Frank act on credit ratings & Journal of Financial Economics \\
Duanmu and McBrayer (2024)$^{\dagger}$ & Structural Drivers of Credit Rating Uncertainty: An Examination of the Changes Imposed by Dodd-Frank & Journal of Financial Services Research \\
Gao, Liao, and Wang (2018)$^{\dagger}$ & Capital markets’ assessment of the economic impact of the Dodd–Frank Act on systemically important financial firms & Journal of Banking and Finance \\
Leledakis and Pyrgiotakis (2022)$^{\dagger}$ & U.S. bank M\&As in the post-Dodd–Frank Act era: Do they create value? & Journal of Banking and Finance \\
Sharma, Adhikari, Agrawal, Arthur, and Rabarison (2022) & Unintended Consequences of the Dodd-Frank Act on Credit Rating Risk and Corporate Finance & Journal of Financial and Quantitative Analysis \\
\addlinespace
\multicolumn{3}{@{}l}{\textit{EBA capital exercise (1)}} \\
Schmidt (2019) & Bank leverage, capital requirements and the implied cost of (equity) capital & Working paper \\
\end{longtable}
}

\end{document}